\documentclass[11pt]{article}

\usepackage[margin=1in]{geometry}

\usepackage[round]{natbib}

\usepackage[utf8]{inputenc}
\usepackage[T1]{fontenc}
\usepackage{amsmath,amssymb}        % For math
\usepackage{mathtools}
\usepackage{amsthm}
\usepackage{thmtools}
\usepackage{graphicx}               % For figures
\usepackage{algorithm}
\usepackage{algorithmic}
\usepackage{booktabs}

\usepackage{tabularx}
\usepackage{makecell}
\usepackage[dvipsnames]{xcolor}
\usepackage{xspace}
\usepackage{enumitem}
\usepackage{todonotes}
\setuptodonotes{inline}
\usepackage{changepage}
\usepackage{subcaption}
\usepackage{wrapfig}
\usepackage{authblk}
\usepackage{microtype}
\usepackage{url}
\usepackage{hyperref}               % For hyperlinks (load last)
\hypersetup{colorlinks=true, linkcolor=blue!50!black, citecolor=blue!50!black, urlcolor=blue!50!black}

\usepackage[section]{placeins}

\newcommand{\rebuttal}[1]{#1}
\newcommand{\revision}[1]{#1}

\newcommand{\tabref}[1]{Table~\ref{#1}}
\newcommand{\figref}[1]{Fig.~\ref{#1}}
\newcommand{\eqnref}[1]{\text{Eq.}~(\ref{#1})}
\newcommand{\secref}[1]{\S\ref{#1}}
\newcommand{\appref}[1]{Appendix~\ref{#1}}

\title{Direct Regret Optimization in Bayesian Optimization}

\author[1]{Fengxue Zhang}
\author[1]{Yuxin Chen}
\affil[1]{Department of Computer Science, University of Chicago}
\affil[ ]{\texttt{\{zhangfx, chenyuxin\}@uchicago.edu}}

\begin{document}

\maketitle

\begin{abstract}
Bayesian optimization (BO) is a powerful paradigm for optimizing expensive black-box functions. Traditional BO methods typically rely on separate hand-crafted acquisition functions and surrogate models for the underlying function, and often operate in a myopic manner. In this paper, we propose a novel \emph{direct regret optimization} approach that jointly learns the optimal model and non-myopic acquisition by distilling from a set of candidate models and acquisitions, and explicitly targets minimizing the multi-step regret. Our framework leverages an \emph{ensemble of Gaussian Processes (GPs)} with varying hyperparameters to generate simulated BO trajectories, each guided by an acquisition function drawn from a pool of conventional choices and terminated by a \emph{Bayesian early stop criterion}. These trajectories train an end-to-end Decision Transformer that selects the next query so as to improve the ultimate objective, following a \emph{dense training--sparse learning} paradigm: the transformer is trained on abundant simulated data, while a limited number of real evaluations refine the GPs online. On synthetic and real-world benchmarks, our method attains the best or near-best final simple regret against standard, lookahead, trust-region and amortized BO baselines, with the largest gains in high-dimensional settings. Ablations attribute the gains jointly to region-of-interest filtering and the learned policy, and matched-budget comparisons against explicit two-step lookahead acquisitions show that the advantage is not shared by lookahead alone.
\end{abstract}
\section{Introduction}
\label{sec:introduction}

Bayesian optimization (BO) has emerged as a powerful framework for optimizing expensive black-box functions under limited evaluation budgets. In many applications, such as hyperparameter tuning in machine learning \citep{Bergstra2012}, engineering design \citep{Jones1998}, and robotic control \citep{Calandra2016}, the objective function is expensive or time-consuming to evaluate. Therefore, classical optimization techniques that rely on dense sampling or gradient information become impractical. Instead, BO constructs a probabilistic surrogate model---commonly a Gaussian Process (GP)---over the objective and uses an \emph{acquisition function} (e.g., Expected Improvement (EI) or Upper Confidence Bound (UCB)) to decide where to sample next. This strategy has proven effective in balancing exploration and exploitation, often achieving promising results with relatively few function evaluations \citep{Shahriari2016}.

Despite BO's success, traditional methods typically rely on separately hand-crafted acquisition functions and surrogate models (e.g., GPs, deep models) and often operate in a myopic manner. This separation and the heuristic nature of acquisition functions mean they only \emph{indirectly} target the primary goal of minimizing the \emph{simple regret}---the difference between the global maximum and the best solution found. Such an approach can be suboptimal for achieving optimal multi-step performance, especially in high-dimensional settings. Furthermore, tuning these components and their hyperparameters for complex problems can be non-trivial and may fail to capture intricate objective structures, thereby hindering overall optimization performance.

\begin{figure}[t]
    \centering
    \includegraphics[width=.78\linewidth]{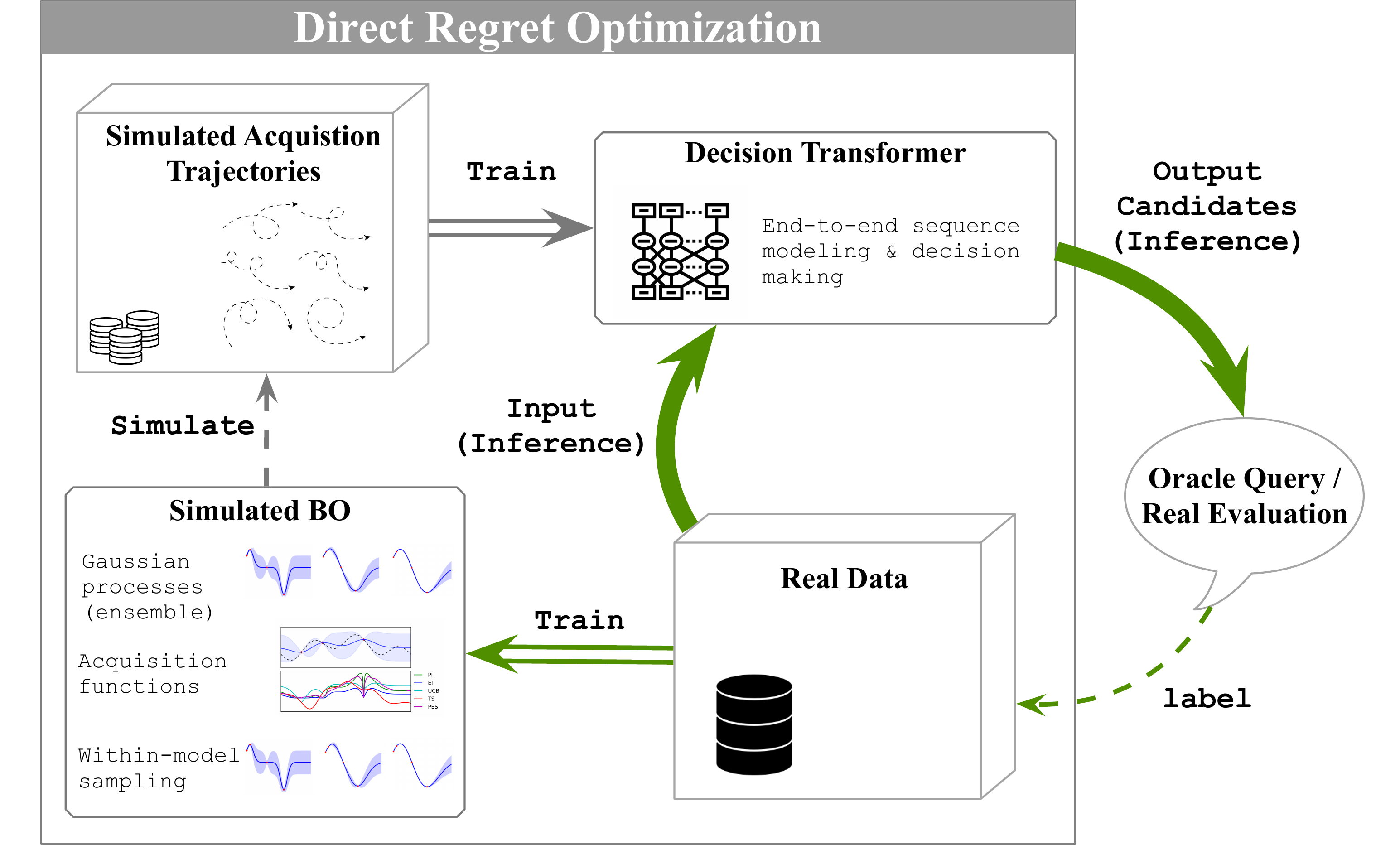} % ensure it fits exactly in one column
    \caption{The DRO framework. \textcolor{gray}{Gray arrows} indicate processes involving simulated data, while \textcolor{ForestGreen}{green arrows} correspond to real data. The \textcolor{gray}{dense training} (\textcolor{gray}{gray}) and \textcolor{ForestGreen}{sparse learning} (\textcolor{ForestGreen}{green}) steps are depicted in hollow arrows.}
    \label{fig:dro}
    \vspace{-3mm}
\end{figure}
\paragraph{Direct regret optimization.}

Motivated by these challenges, we propose \emph{Direct Regret Optimization (DRO)}\footnote{Not to be confused with distributionally robust optimization.}, a novel approach that \textbf{jointly learns an optimal decision-making model and a non-myopic acquisition policy}, explicitly targeting the minimization of multi-step regret (\figref{fig:dro}). Our framework achieves this by \textbf{distilling knowledge from a diverse set of candidate models and acquisition strategies}. Specifically, DRO leverages an \emph{ensemble of GPs} with varying hyperparameters to generate simulated BO trajectories. Each trajectory, guided by an acquisition function from a pool of conventional choices, is strategically constrained within an identified \emph{region of interest (ROI)}---a subset of the domain likely to contain the optimum---and is managed by an early-stopping criterion. These simulated trajectories, capturing rich multi-step exploration strategies, are then used to train a \emph{Decision Transformer} end-to-end. This transformer directly learns to select the next queries to improve the objective, effectively embodying the learned non-myopic acquisition policy. Throughout, \emph{non-myopic} refers to the training signal rather than to explicit planning: every action is labeled with the improvement still to be realized by the end of the complete trajectory it belongs to, not with a one-step utility (\secref{subsec:dt}). We claim neither formal lookahead planning nor an end-to-end guarantee for the learned policy. % and decision model. 
This paradigm shifts BO from relying on separately designed, often myopic heuristics to an integrated sequence modeling approach that directly optimizes for long-term outcomes.

\paragraph{Dense training and sparse learning.}
A second ingredient is a \emph{dense training--sparse learning} paradigm: the Decision Transformer is trained extensively on abundant simulated trajectories from the GP ensemble, learning to anticipate multi-step outcomes without real evaluations, while each costly real evaluation provides sparse feedback that re-anchors the ensemble, and hence the policy, in the true objective.

\paragraph{Contributions.}
Our contributions can be summarized as follows:
\begin{itemize}[leftmargin=*,topsep=0pt]
\item We introduce a \textbf{direct regret optimization} framework that jointly learns a decision-making model and a non-myopic acquisition policy by distilling from candidate strategies, explicitly minimizing multi-step regret (\secref{sec:problem-formulation} and \secref{sec:method}).
\item We propose a novel use of \textbf{ensemble-based rollout simulations}, constrained within a \textbf{Region of Interest (ROI)} and guided by early stopping, to generate rich trajectory data for training a \emph{Decision Transformer} end-to-end (\secref{subsec:ensemble_gps}--\ref{subsec:early_stop_method}).
\item We present a novel \textbf{dense training--sparse learning} paradigm that enables robust multi-step decision-making using offline simulated data, while efficiently incorporating limited real evaluations for online refinement (\secref{subsec:dense-training-sparse-learning}).
\item We conduct experiments on a range of synthetic and real-world benchmarks, showing that DRO attains the best or near-best final simple regret against standard, one- and two-step lookahead, trust-region and amortized baselines, with the largest margins in high-dimensional settings (\secref{sec:results-discussion}). Ablations isolate the roles of ROI filtering, ensemble diversity, the composition of the rollout pool, and the learned policy itself (\secref{sec:ablations}). 
\end{itemize}

% \paragraph{Organization.}
% The remainder of this paper is structured as follows. In \secref{sec:problem-formulation}, we formalize the Bayesian optimization problem. \secref{sec:method} details our proposed direct regret optimization approach, including the ensemble GP simulator, Bayesian early stopping, region of interest identification, and Decision Transformer training. \secref{sec:experiments} describes our experimental setup, benchmarks, and baselines, while \secref{sec:results-discussion} reports empirical results and provides analysis. We conclude in \secref{sec:conclusion} with a discussion of limitations and possible future directions.

\section{Related Work}
\label{sec:related-work}
%Bayesian optimization has seen significant advancements through integration with machine learning, particularly reinforcement learning (RL) and deep learning, to address limitations of traditional approaches.
% \footnote{We include additional related work in the appendix.} 

\subsection{Reinforcement and Deep Learning in Bayesian Optimization}
\label{subsec:rl_and_deep_learning_bo_summary}

Advanced machine learning, including reinforcement learning (RL) and deep learning, enhances BO with adaptive and scalable strategies.
RL offers adaptive sampling in BO: FSAF \citep{Hsieh2021} and RLABO \citep{Liu2022} learn query selection, while frameworks like EARL-BO by \citet{Cheon2022, Cheon2024} enable non-myopic, multi-step lookahead. RL also aids surrogate modeling (RLBOF; \citealp{Huang2024}) and multi-objective BO (e.g., in TinyML NAS; \citealp{Deutel2025}). These methods rely on specific modeling assumptions or on meta-data from related tasks, whereas DRO learns its policy from simulated rollouts of the task at hand.

Broader deep learning techniques improve BO's flexibility, exemplified by latent space optimization \citep{Tripp2020} and deep surrogates for complex problems \citep{Kim2021, Zhang2023ROI}. Transformer architectures are particularly prominent: Prior-Data Fitted Networks (PFNs4BO; \citealp{muller2023pfns4bo}) provide amortized Bayesian inference and implicitly marginalize over surrogate uncertainty via meta-training. Meta-learning with Neural Processes (NPs) \citep{garnelo2018neural, garnelo2018conditional, kim2019attentive}, meta-learned acquisition functions \citep{volpp2020meta}, or amortized query policies \citep{swersky2020amortized} aim to learn transferable models. Furthermore, end-to-end meta-BO frameworks like OptFormer \citep{chen2022towards} and Neural Acquisition Process (NAP) \citep{maraval2023end} leverage transformers to learn universal optimizers. \rebuttal{Our approach is also related to amortized experimental design and optimization, where models are trained to output queries based on simulated trajectories \citep{foster2021deep, ivanova2021implicit, blau2022optimizing, huang2024amortized, chang2025amortized, zhang2025pabbo}. For instance, \citet{chen2017learning} train an RNN to generate queries from simulated acquisition function trajectories. However, their method relies on batch pre-training before any real optimization, whereas DRO interleaves simulation and real evaluations for online refinement. Additionally, unlike meta-BO methods that require multi-task meta-data for training \citep[e.g.,][]{maraval2023end}, DRO operates in a zero-shot setting, training exclusively on GP-sampled synthetic functions.}

\paragraph{Non-myopic acquisition functions.}
Multi-step lookahead has been pursued through approximate dynamic programming and rollout \citep{Lam2016}, by simulating the future steps of the optimizer \citep{gonzalez2016glasses}, by two-step lookahead \citep{wu2019practical}, and by jointly optimizing one-shot multi-step trees \citep{jiang2020efficient}; the knowledge gradient \citep{frazier2008knowledge} prices the one-step gain in the maximum of the posterior mean. These methods plan explicitly under a single surrogate at a cost that grows quickly with the horizon; DRO instead learns a policy from ensemble-simulated trajectories labeled with their terminal return, and \secref{sec:lookahead} compares it with one- and two-step lookahead acquisitions under a matched budget.

\subsection{Handling Unknown Hyperparameters in Bayesian Optimization}
\label{subsec:unknown-hyperparams-bo_summary}

BO performance is notably sensitive to surrogate model hyperparameters (e.g., for Gaussian Processes), which are often fixed or unreliably estimated from sparse data via Maximum Likelihood (MLE) or Maximum A Posteriori (MAP) estimation. Fully Bayesian approaches attempt to mitigate this by marginalizing hyperparameters using priors and techniques like Markov Chain Monte Carlo (MCMC) \citep{snoek2012practical} to compute integrated acquisition functions (e.g., Integrated Expected Improvement; \citealp{brochu2010tutorial}). Ensemble methods offer a practical alternative by averaging predictions or acquisition functions from models with different hyperparameter settings \citep{wistuba2018scalable}. \rebuttal{Similarly, portfolio allocation methods \citep{hoffman2011portfolio} and modern frameworks like HEBO \citep{cowen2022hebo} optimize the Pareto frontier of multiple acquisition functions, while packages like scikit-optimize \citep{head2021scikit} implement GP-Hedge strategies. Furthermore, focusing computational resources on areas of interest using mixtures of GPs has been explored in funneled BO \citep{martinez2018funneled}.} As noted earlier, transformer-based approaches like PFNs \citep{muller2023pfns4bo} can also implicitly address hyperparameter uncertainty through their meta-training process. Other research focuses on acquisition functions that explicitly account for hyperparameter uncertainty in addition to uncertainty about the function's maximum value \citep{hvarfner2023self}, though these may still depend on conventional GP surrogates and computationally intensive information-theoretic criteria. Our work, in contrast, utilizes an ensemble of GPs with varying hyperparameters primarily to generate diverse simulation data, thereby addressing hyperparameter sensitivity within the data generation phase that fuels the learning of our decision policy.

\section{Problem Statement}
\label{sec:problem-formulation}

Let $\mathcal{X}\subseteq \mathbb{R}^d$ be a $d$-dimensional design (or input) space, and let $f:\mathcal{X}\rightarrow\mathbb{R}$ be an unknown objective function that is typically expensive to evaluate. The goal of \textit{Bayesian optimization} %(BO) 
%\yuxin{Removing acronyms (BO, GP, ML, ROI, DRO, etc). I tried to clean it up but may miss a few -- make sure acronyms are not defined multiple times} 
is to find a design $\mathbf{x}^*\in \mathcal{X}$ that \textit{maximizes} this objective:
\begin{equation}
    \mathbf{x}^* = \underset{\mathbf{x}\in\mathcal{X}}{\mathrm{argmax}}\; f(\mathbf{x}).
\end{equation}

The conventional BO process is iterative. At each iteration $t$, a probabilistic surrogate model, commonly a Gaussian process, is updated based on the accumulated data $\mathcal{D}_{t-1}=\{(\mathbf{x}_i,y_i)\}_{i=1}^{t-1}$, where $y_i = f(\mathbf{x}_i) + \epsilon_i$ are (potentially noisy) observations with $\epsilon_i \sim \mathcal{N}(0, \sigma_n^2)$ representing i.i.d. Gaussian noise. This GP yields a posterior distribution $p(f(\mathbf{x}) \mid \mathcal{D}_{t-1})$ for any point $\mathbf{x} \in \mathcal{X}$, characterized by its mean $\mu_{t-1}(\mathbf{x})$ and variance $\sigma^2_{t-1}(\mathbf{x})$ (with standard deviation $\sigma_{t-1}(\mathbf{x})$). An acquisition function $\alpha(\mathbf{x}; \mathcal{D}_{t-1})$, leveraging $\mu_{t-1}(\mathbf{x})$ and $\sigma_{t-1}(\mathbf{x})$, is then maximized to select the next point $\mathbf{x}_t = \underset{\mathbf{x}\in\mathcal{X}}{\mathrm{argmax}}\; \alpha(\mathbf{x}; \mathcal{D}_{t-1})$. The objective function is subsequently evaluated at $\mathbf{x}_t$ to obtain $y_t$, and the dataset is augmented, $\mathcal{D}_t = \mathcal{D}_{t-1} \cup \{(\mathbf{x}_t, y_t)\}$, with this cycle repeating for a total of $T$ iterations within a given budget.

\paragraph{Objective and regret.}
Let $f^* = \max_{\mathbf{x}\in\mathcal{X}} f(\mathbf{x})$ be the true maximum value of the objective function. After $T$ evaluations, let $\mathbf{x}_T^+$ be the point among $\{\mathbf{x}_1, \dots, \mathbf{x}_T\}$ that has yielded the highest true function value, i.e., $f(\mathbf{x}_T^+) = \max_{1 \le t \le T} f(\mathbf{x}_t)$.
The \textit{simple regret} at iteration $T$ is then defined as 
    $\mathrm{Regret}_{\text{simple}}(T) = f^* - f(\mathbf{x}_T^+)$. 
% \begin{equation}
%     \mathrm{Regret}_{\text{simple}}(T) = f^* - f(\mathbf{x}_T^+).
% \end{equation}
Alternatively, the \textit{cumulative regret} considers the sum of regrets at each step:  $\mathrm{Regret}_{\text{cumulative}}(T) = \sum_{t=1}^T \bigl(f^* - f(\mathbf{x}_t)\bigr).$
% \begin{equation}
%     \mathrm{Regret}_{\text{cumulative}}(T) = \sum_{t=1}^T \bigl(f^* - f(\mathbf{x}_t)\bigr).
% \end{equation}

BO minimizes these regrets by balancing \textit{exploration} and \textit{exploitation}; greedily maximizing a myopic, one-step acquisition function, however, need not minimize the regret at the end of the horizon, which is the quantity we target.
\section{Method}
\label{sec:method}

We now introduce our Direct Regret Optimization (DRO) framework. DRO departs from conventional BO methods by directly targeting minimization of the final simple regret. The core strategy involves simulating diverse potential BO trajectories using an ensemble of GP models, each with different characteristics. These simulations, constrained within an adaptively identified Region of Interest (ROI), are then used to train a decision-making model---specifically, a Decision Transformer---to propose candidate points predicted to yield the lowest simple regret at the end of the horizon.

\subsection{Ensemble of Gaussian Processes}
\label{subsec:ensemble_gps}
At each time step $t$ of the optimization process, we maintain an \emph{ensemble} of $M$ GP surrogate models:
\begin{equation}
    \{ \mathrm{GP}_1, \mathrm{GP}_2, \ldots, \mathrm{GP}_M \}.
\end{equation}
Each $\mathrm{GP}_m$ in the ensemble is constructed with a distinct, single, isotropic lengthscale, drawn from a predefined set that includes both small and large values. \revision{This is motivated by theoretical findings concerning BO with unknown hyperparameters \citep{berkenkamp2019no}, which suggest that achieving no-regret bounds necessitates the lengthscale becoming sufficiently small in later stages of optimization to prevent the model from overly restricting the function space. By incorporating a varied set of lengthscales, our ensemble aims to ensure adequate function modeling capability and capture diverse beliefs about the objective. This ensemble-based approach serves as a robust and computationally tractable alternative to perfectly tuning a more complex model, providing an effective sparse discretization of the multi-dimensional hyperparameter space.} All GPs are conditioned on the same accumulated dataset $\mathcal{D}_{t-1}$. Rollouts are allocated uniformly across members by default; allocating them in proportion to the softmax of each member's marginal log-likelihood on $\mathcal{D}_{t-1}$, which gradually down-weights hypotheses that the real data contradict, is a drop-in variant evaluated in \appref{subsec:kernel_diversity}.

\subsection{Region of Interest (ROI) Filtering}
\label{subsec:roi_filtering}
To focus computational resources on the most promising areas of the search space, we incorporate an adaptive ROI filtering mechanism, similar to filtering in trust-region BO \citep{eriksson2019scalable}, constrained BO \citep{eriksson2021scalable, pmlr-v202-xu23h, zhang2024finding}, and adaptive batched BO \citep{li2022gaussian}. At each time step $t$, after updating the GP models, each $\mathrm{GP}_m$ in the ensemble identifies its own ROI $\hat{\mathcal{X}}_{m,t} \subseteq \mathcal{X}$.

For each $\mathrm{GP}_m$, which provides a posterior distribution over $f$ conditioned on $\mathcal{D}_{t-1}$, we compute its Upper Confidence Bound (UCB) and Lower Confidence Bound (LCB) across $\mathcal{X}$:
\begin{align}
\mathrm{UCB}_{m,t}(\mathbf{x}) &= \mu_{m,t-1}(\mathbf{x}) + \beta_t^{1/2}\sigma_{m,t-1}(\mathbf{x}), \label{eq:ucb_m}\\
\mathrm{LCB}_{m,t}(\mathbf{x}) &= \mu_{m,t-1}(\mathbf{x}) - \beta_t^{1/2}\sigma_{m,t-1}(\mathbf{x}), \label{eq:lcb_m}
\end{align}
where $\mu_{m,t-1}(\mathbf{x})$ and $\sigma_{m,t-1}(\mathbf{x})$ are the posterior mean and standard deviation from $\mathrm{GP}_m$ based on $\mathcal{D}_{t-1}$, and $\beta_t$ is an exploration-exploitation trade-off parameter.
The ROI for $\mathrm{GP}_m$, denoted $\hat{\mathcal{X}}_{m,t}$, is then defined as the superlevel set of points where its UCB is no less than the maximum LCB it observes anywhere in the space:
\begin{equation}
\hat{\mathcal{X}}_{m,t} = \{ \mathbf{x} \in \mathcal{X} \mid \mathrm{UCB}_{m,t}(\mathbf{x}) \ge \max_{\mathbf{x}' \in \mathcal{X}} \mathrm{LCB}_{m,t}(\mathbf{x}') \}.
\label{eq:roi_definition_m} % Changed label
\end{equation}
This dynamically identified ROI, $\hat{\mathcal{X}}_{m,t}$, serves as the constrained search space for the rollout simulations generated using $\mathrm{GP}_m$, ensuring that its simulated efforts are concentrated in regions it has high confidence of containing the global optimum. \revision{Appendix~\ref{sec:theory} shows that GP-UCB maximized inside such an adaptively identified region retains the standard GP-UCB regret bound on the same high-probability event, because the region contains the optimizer whenever the confidence bounds hold (Theorem~\ref{thm:roi_regret}). This concerns the trajectory-generating subroutine of \eqnref{eq:rollout}, not the learned policy, for which we claim no end-to-end guarantee.}

\subsection{Within-Model Sampling %(Rollout Generation) 
within ROI}
\label{subsec:within_model_sampling}
For each $\mathrm{GP}_m$ in the ensemble (from \secref{subsec:ensemble_gps}), we generate a set of simulated future BO trajectories, often referred to as \emph{rollouts}. These rollouts are produced by iteratively selecting candidate points using a conventional acquisition function $\alpha(\cdot; \mathrm{GP}_m)$, such as EI or UCB, evaluated with respect to $\mathrm{GP}_m$'s posterior distribution $p_m(f \mid \mathcal{D}_{t-1})$. Crucially, the search for the next point in the simulation is constrained to the identified Region of Interest $\hat{\mathcal{X}}_{m,t}$. Formally, at each simulation step $\tau$ within a rollout for $\mathrm{GP}_m$:
\begin{equation}
    \mathbf{x}_\tau^{(m)}
    = \underset{\mathbf{x}\in\hat{\mathcal{X}}_{m,t}}{\operatorname{argmax}} % Constrained to ROI
        \; \alpha\bigl(\mathbf{x}; \mathrm{GP}_m, \mathcal{D}_{t-1}^{(\tau, m)}\bigr),
\label{eq:rollout}
\end{equation}
where $\mathcal{D}_{t-1}^{(\tau, m)}$ represents the data accumulated up to simulation step $\tau-1$ for that specific rollout under $\mathrm{GP}_m$. The function $f(\mathbf{x}_\tau^{(m)})$ is then ``observed'' by sampling from $\mathrm{GP}_m$'s posterior predictive distribution at $\mathbf{x}_\tau^{(m)}$. This iterative sampling continues until a \textit{Bayesian early stop} criterion (detailed in \secref{subsec:early_stop_method}) is met. This yields simulated trajectories of varying lengths, each a plausible future under one member's beliefs and focused within its ROI. To further diversify the simulated experiences, we rotate the acquisition function used for each maximization uniformly at random over a pool (EI, UCB, PI and MES by default); \appref{subsec:pool} studies how the composition of this pool affects the learned policy.

\subsection{Bayesian Early Stopping (BES)}
\label{subsec:early_stop_method}
To ensure computational tractability and focus rollouts on informative sequences, we implement a \textit{Bayesian Early Stopping} (BES) mechanism. After each simulated query step $\tau$ within a rollout for a given $\mathrm{GP}_m$, we evaluate a stopping criterion. One approach is EI-based: the rollout (within ROI $\hat{\mathcal{X}}_{m,t}$) terminates if the maximum Expected Improvement attainable at the next step falls below a threshold $\delta$. Let $f_{m,\text{best}}^{(\tau)}$ be the incumbent value after simulated step $\tau$. The stopping condition is:
\begin{equation}
\max_{\mathbf{x} \in \hat{\mathcal{X}}_{m,t}} \mathbb{E}_{p_m(f(\mathbf{x}) \mid \mathcal{D}_{t-1}^{(\tau+1, m)})} \bigl[\max(0, f(\mathbf{x}) - f_{m,\text{best}}^{(\tau)})\bigr] < \delta. \label{eq:bes_ei}
\end{equation}
This threshold $\delta$ can be a small constant or scaled dynamically. In the released implementation the test is applied to the realized improvement of the sampled point, $\max(0, y_\tau - f_{m,\text{best}}^{(\tau-1)})$, a one-sample proxy for \eqnref{eq:bes_ei} that avoids an additional acquisition maximization per step; rollouts are also capped at a maximum length $L$ (\appref{sec:implementation_details}). This criterion prevents excessively long rollouts into regions where $\mathrm{GP}_m$ predicts negligible further improvement, ensuring simulations capture plausible scenarios within the ROI. \revision{Appendix~\ref{sec:theory} bounds the expected improvement at any point by the surrogate's mean error and posterior width there plus the current simple regret (Theorem~\ref{thm:ei_convergence}); on an ROI where the simulator is consistent, vanishing simple regret therefore drives the maximal expected improvement to zero, which is what makes EI a principled indicator for terminating uninformative rollouts.}

\subsection{Training and Inference of Decision Transformer}
\label{subsec:dt}
Simulated trajectories form an offline dataset for training a \emph{Decision Transformer} \citep{chen2021decision}. Each trajectory is converted into sequences of (state, action, return-to-go) tuples: states include BO progress features (the fitted hyperparameters of every ensemble member, the incumbent value and location, and the normalized iteration index $t/T$), actions are ROI-constrained query points, and the return-to-go at step $\tau$ is the improvement still to be realized by the end of the trajectory, $R_\tau = f_{m,\text{best}}^{(L)} - f_{m,\text{best}}^{(\tau)}$ (with the incumbent held at its final value once the rollout stops), i.e., the sum of the future simulated improvements. The transformer is thus \emph{not} trained to imitate the acquisition functions that generated the data: the same action receives different labels depending on what followed it, which is the sense in which the learned policy is non-myopic. At decision time we condition on a high target return, selecting the high-return slice of a diverse behavior distribution ($M$ lengthscale hypotheses $\times$ rotating acquisitions $\times$ stochastic rollouts), the stitching mechanism of return-conditioned offline RL \citep{chen2021decision}. \rebuttal{To ensure generalization across different function evaluation ranges, observations are standardized within each trajectory to have zero mean and unit variance, and the return-to-go is scaled to $[0, 1]$ based on the observed range in simulated data.} The transformer is trained to output actions achieving a high target return (low final regret). In real BO, the trained model, conditioned on the current state and a target return, proposes $\mathbf{x}_t$. When the task's optimal reward is known (e.g., bounded accuracy), returns are exact; otherwise we derive an optimistic estimate from simulated trajectories and calculate the return-to-go accordingly.

\subsection{Overall DRO Procedure}
Algorithm~\ref{alg:direct-regret-optimization} summarizes one real iteration: update the ensemble on $\mathcal{D}_{t-1}$, identify each member's ROI (\eqnref{eq:roi_definition_m}), simulate ROI-constrained rollouts with BES (\secref{subsec:within_model_sampling}, \ref{subsec:early_stop_method}), fine-tune the Decision Transformer (DT) on the resulting buffer, and evaluate the point it proposes. A single transformer and its optimizer persist across real iterations and are warm-started, so the policy accumulates experience as the ensemble is refined.

\begin{algorithm}[t]
\caption{Direct Regret Optimization with ROI Filtering}
\label{alg:direct-regret-optimization}
\begin{algorithmic}[1]
    \REQUIRE Initial dataset $\mathcal{D}_0$, ensemble size $M$, evaluation budget $T$, early stop threshold $\delta$, rollouts per GP $K$, ROI params (e.g., $\beta_t$)
    \FOR{$t = 1$ to $T$}
        \STATE \textbf{Fit/Update GPs}: Update $\mathrm{GP}_m$ for $m=1 \dots M$ using $\mathcal{D}_{t-1}$.
        \STATE \textbf{Identify ROI}: Determine $\hat{\mathcal{X}}_{m,t} \leftarrow \{ \mathbf{x} \in \mathcal{X} \mid \mathrm{UCB}_{m,t}(\mathbf{x}) \ge \max_{\mathbf{x}' \in \mathcal{X}} \mathrm{LCB}_{m,t}(\mathbf{x}') \}$ using $\mathrm{GP}_{m}$.
        \STATE $\mathcal{B}_{sim} \gets \{\}$.
        \FOR{$m = 1$ to $M$, $k = 1$ to $K$}
            \STATE Simulate $\mathcal{T}_{m,k} = \{(\mathbf{s}_\tau, \mathbf{a}_\tau, R_\tau)\}_{\tau=0}^{L_{m,k}-1}$ with $\mathrm{GP}_m$, acquisition $\alpha$ maximized over $\hat{\mathcal{X}}_{m,t}$, and Bayesian early stopping; add it to $\mathcal{B}_{sim}$.
        \ENDFOR
        \STATE \textbf{Train/Update DT} using trajectories in $\mathcal{B}_{sim}$ to predict actions that minimize final regret.
        \STATE \textbf{Infer Next Candidate} with  transformer $\mathbf{x}_t \gets \mathrm{DT}(\text{current real state } \mathbf{s}_t, \text{target return } R_{\text{target}})$.
        \STATE Evaluate $y_t = f(\mathbf{x}_t)$ and augment real data $\mathcal{D}_t \gets \mathcal{D}_{t-1} \cup \{(\mathbf{x}_t, y_t)\}$.
    \ENDFOR
\end{algorithmic}
\end{algorithm}

\paragraph{Training objective for DRO.}
The implicit training objective of the Decision Transformer is the final simple regret $\mathrm{Regret}_{\text{simple}}(T) = f^* - f(\mathbf{x}_{T}^+)$: instead of a one-step acquisition used as a proxy, the policy conditions its decisions on achieving low regret at the end of the horizon, using multi-step trajectories simulated inside the ROI.

\paragraph{Dense training vs. sparse learning.}
\label{subsec:dense-training-sparse-learning}
DRO separates two learning regimes. In the \textit{dense training} phase the Decision Transformer learns from a large corpus of simulated trajectories, generated at every real iteration by the GP ensemble inside the ROI under varied model assumptions; this is where multi-step behavior is learned, from inexpensive ``imagined'' experience focused on promising regions. In the \textit{sparse learning} phase each costly real evaluation $f(\mathbf{x}_t)$ updates the ensemble, grounds the simulator in the true objective and refines the ROI for the next iteration. Training on rollouts from an ensemble rather than a single surrogate also makes the policy robust to the inaccuracies of any individual model, a benefit the ROI amplifies by discounting regions where model error matters least.

\rebuttal{\paragraph{Computational complexity.}
The overhead of DRO comes from the simulation phase: fitting the ensemble costs $\mathcal{O}(M n^3)$ for $n$ observations, identifying each ROI costs one UCB and one LCB maximization, and each real iteration runs $M \times K$ rollouts of at most $L$ steps (one ROI-constrained acquisition maximization and one posterior draw per step) followed by a fixed number of fine-tuning epochs of the transformer; $K$ and $L$ are listed per task in \appref{sec:implementation_details}. This is cheap relative to expensive real evaluations, though not negligible for objectives that take seconds to evaluate (wall times in \appref{subsec:time}). By avoiding the need to compute and differentiate through multi-step GP posteriors---a bottleneck in methods that directly optimize joint acquisition scores over long horizons---DRO maintains computational feasibility even in high-dimensional tasks.}

\section{Experiments}
\label{sec:experiments}

\revision{We evaluate the empirical performance of DRO against established baselines, followed by ablation studies to isolate the contributions of its core components.}

% \subsection{Benchmarks}
% \label{sec:benchmarks}
% Our evaluation suite for this study focuses on:
% \begin{itemize}
%     \item \textbf{Synthetic Functions:} The \textit{Ackley Function}~\citep{ernesto2005mvf}, evaluated across multiple dimensions (2D, 5D, 10D, and 20D) to assess scalability and performance (\figref{fig:ackley_dim_study_main}). The Ackley 10D version also serves as the testbed for our ablation studies (\figref{fig:ablation_and_lunar_results}). Note we add universal shift of 10 in the $[-32.768, 32.768]^D$ domain in all dimensions so that global optimum no longer lies in the center. We also introduce Gaussian noise of $0.1$ standard deviation to the observations.
%     \item \textbf{Hyperparameter Optimization (HPO) Tasks:} We test on four HPO scenarios: tuning hyperparameters for XGBoost, and for Neural Networks with the ADAM optimizer on UCI datasets (Wine, Iris, Breast Cancer). The XGBoost benchmark is extracted from YAHPO Gym \citep{pfisterer2022yahpo}, while the remaining are extracted from Bayesmark \citep{muller2023pfns4bo}. These are presented collectively in \figref{fig:hpo_regrets_main}. 
%     \item \textbf{Complex Simulation:} The \textit{LunarLander-v3}~\citep{towers2024gymnasium} control problem from OpenAI Gym, where policy parameters are optimized (\figref{fig:lunarlander_main}).
%     % \item \textbf{Synthetic Function for Ablations:} The \textit{Ackley 10D function}~\citep{ernesto2005mvf} serves as the testbed for our ablation studies (Figures~\ref{fig:ackley_10d_roi_ablation_main} and \ref{fig:ackley_10d_ensemble_study_main}).
% \end{itemize}

\subsection{Experimental Setup}
\label{sec:setup}

\paragraph{Benchmarks.} Our evaluation suite includes synthetic functions, Hyperparameter Optimization (HPO) tasks, and a reinforcement learning (RL) control task. Specifically, we use the \textit{Ackley Function}~\citep{ernesto2005mvf} across multiple dimensions (2D, 5D, 10D, and 20D, \figref{fig:ackley_dim_hpo_study}) to assess scalability. \rebuttal{We intentionally design our benchmark environment to be more challenging than the simplified protocols often used in the literature. For these Ackley tasks, we shift the \emph{input} by $+10$ in every coordinate within the standard domain $[-32.768, 32.768]^D$, so that the optimizer sits at $\mathbf{x}^* = 10\cdot\mathbf{1}$ rather than at the center of the domain where space-filling initial designs land close to it, and we add Gaussian observation noise with standard deviation $\sigma = 0.01$. The LunarLander reward is observed with the same noise level; the HPO tasks are evaluated without added noise. This distinction explains why baseline performance may appear lower compared to studies using noiseless or unshifted objectives. For completeness, Appendix~\ref{sec:additional_results} includes \figref{fig:ackley_logei_reproduce}, a reproduction under the simplified protocol studied in \citet{ament2023unexpected}: noiseless, no output shift, asymmetric reduced domain $[(-32.768/4, 32.768/2)]^D$, and larger Sobol initialization $N_{init}=2(d+1)$, where DRO likewise excels.} The Ackley 10D version also serves as the testbed for our ablation studies (\figref{fig:ablation_and_lunar_results}). HPO tasks include tuning XGBoost (extracted from YAHPO Gym~\citep{pfisterer2022yahpo}) and ADAM optimizer hyperparameters for Neural Networks on UCI datasets (Wine, Iris, Breast Cancer, based on Bayesmark~\citep{muller2023pfns4bo} setups), presented collectively in \figref{fig:ackley_dim_hpo_study}. For the RL control task, we optimize the 12-dimensional parameters of a heuristic controller in the \textit{LunarLander-v3}~\citep{towers2024gymnasium} environment (\figref{fig:ablation_and_lunar_results}).

\paragraph{General setup.} Inputs are normalized to $[0,1]^d$. Runs typically initialize with $N_{init}=5$ points using Sobol sequences. The total evaluation budget $T$ is 30-40 for HPO tasks, up to 100 for LunarLander, and 500 for Ackley studies. Performance is measured by simple regret (for HPO) or best observed objective value (for LunarLander, Ackley), averaged over at least 10 independent trials. Shaded regions in plots denote $\pm$1 standard error. We rely on BoTorch \citep{balandat2020botorch} for all implementation to guarantee fair performance comparison.
%\paragraph{DRO configuration.} 
Unless otherwise specified (e.g., in ablations), DRO uses an ensemble of $M=10$ GPs with \textbf{RBF kernels}, ROI filtering, and Bayesian early stopping. The Decision Transformer has an embedding dimension of 128, 4 attention heads, and 4 layers, trained with Adam; the per-task simulation budget (rollouts per member $K$ and maximum rollout length $L$) is listed in \appref{sec:implementation_details}. For the ROI ablation, `DRO ROI' is this default configuration, while `DRO GLOBAL' disables ROI filtering for rollouts.

\paragraph{Baselines.} On the Ackley, HPO and LunarLander tasks, DRO is compared against logEI acquisition(labeled as ``EI'')~\citep{ament2023unexpected}, knowledge gradient (labeled as ``KG'')~\citep{frazier2008knowledge} which is a look-ahead acquisition function, PFNs4BO~\citep{muller2023pfns4bo} which relies on the transformer as surrogate model, random search (labeled as ``Random''), which calibrates the difficulty of each task, SCoreBO~\citep{hvarfner2023self} which samples hyperparameters and incorporates hyperparameter uncertainty in the acquisition to actively address model misspecification, and TuRBO~\citep{eriksson2019scalable} which is the state-of-the-art trust-region local BO method. On Ackley 5D and 10D we additionally compare, under a matched budget of $T=100$, against explicit two-step lookahead acquisitions implemented with one-shot multi-step trees \citep{jiang2020efficient}: two-step EI \citep{wu2019practical} and two-step KG (\secref{sec:lookahead}).

\subsection{Comparative Performance}
\label{sec:results-discussion}

\begin{figure*}[t]
    \centering
    \includegraphics[width=.8\linewidth]{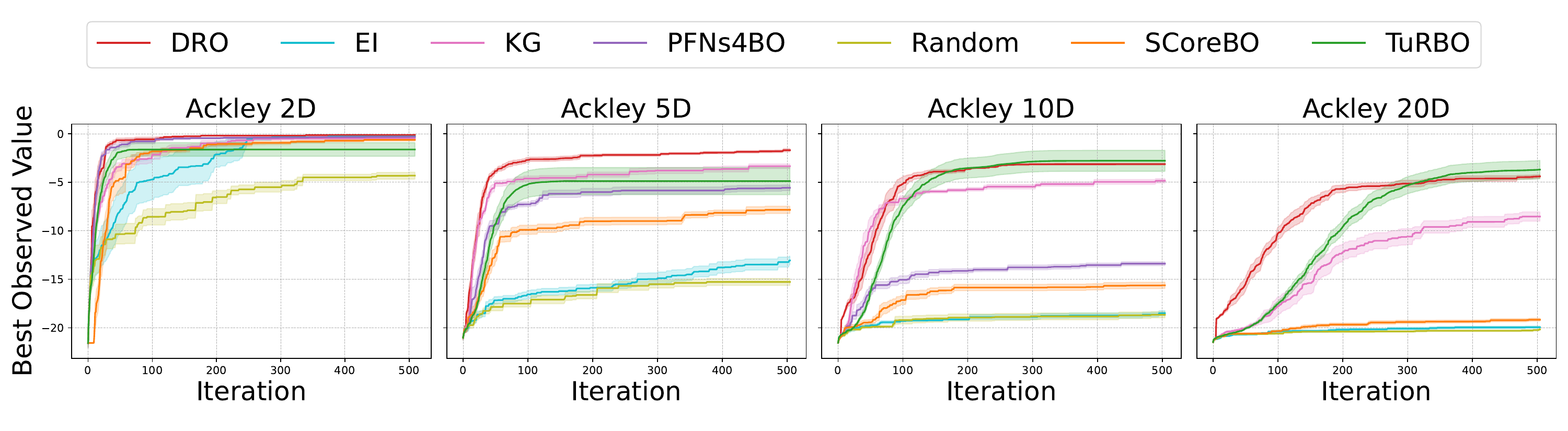}
    % \\ \vspace{-3mm}
    \includegraphics[width=.8\linewidth,trim={0 0 0 2.5cm}, clip]{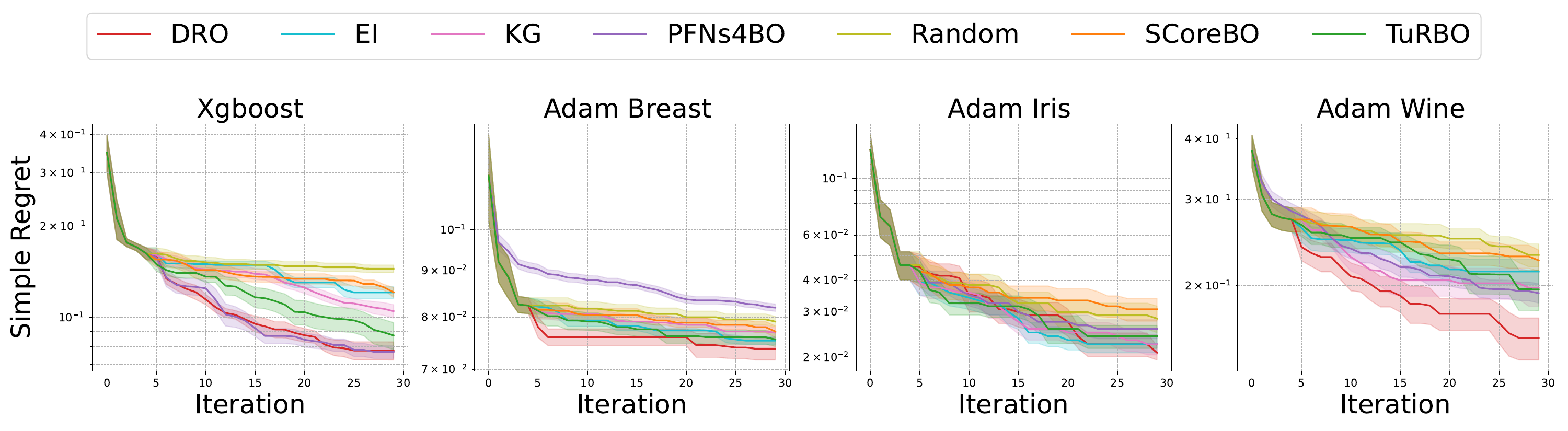} 
    \caption{Upper: Performance on Ackley Function across dimensions (2D, 5D, 10D, 20D - Best Objective Value Found). Higher values are better. PFNs4BO is omitted from Ackley 20D due to scalability. Lower: Performance on HPO Tasks (Simple Regret). Lower values are better.}
    \label{fig:ackley_dim_hpo_study}
    % \vspace{-3mm}
\end{figure*}

\revision{We begin by assessing DRO's scalability on the synthetic Ackley function across dimensions 2D, 5D, 10D, and 20D (upper panel of \figref{fig:ackley_dim_hpo_study}). DRO consistently achieves the best or near-best objective values across all tested dimensions compared to standard BO, PFNs4BO (up to 10D), SCoreBO, and TuRBO. Notably, in the 20D task---where PFNs4BO is omitted due to reported scaling limitations beyond 18 dimensions---DRO maintains robust performance and clearly outperforms the remaining baselines, demonstrating favorable scalability to higher-dimensional problems.} DRO does not lead at every budget, however. Comparing the incumbents of DRO and KG at every iteration of the same runs, DRO trails KG on Ackley 10D for the first $\approx 80$ iterations and leads thereafter (by $0.69$ at iteration 100, with the margin widening through $T=500$), whereas on 5D it leads at every iteration ($+2.05$ at iteration 25 and $+2.96$ at iteration 100). Giving up early performance and overtaking later is the behavior one expects from a policy trained against terminal rather than immediate improvement.

\paragraph{Lookahead baselines.}
\label{sec:lookahead}
KG is a one-step lookahead acquisition. To compare against genuinely multi-step planning, we run two-step EI \citep{wu2019practical} and two-step KG, implemented as one-shot multi-step trees \citep{jiang2020efficient}, under a matched budget of $T=100$ on Ackley 5D and 10D with the seeds and initial designs of the main experiments (\tabref{tab:lookahead} in \appref{subsec:lookahead_appendix}). Explicit two-step trees do not add over their one-step counterparts on this noisy, shifted protocol, and DRO is ahead of all of them at this budget ($-2.63 \pm 0.22$ / $-5.05 \pm 0.71$ on 5D / 10D; within one standard error of KG on 10D) after trailing KG and two-step KG at iteration 50 on 10D.

\begin{figure*}[t]
    \centering
    \begin{subfigure}[b]{0.28\textwidth}
        \centering
    \includegraphics[width=\textwidth]{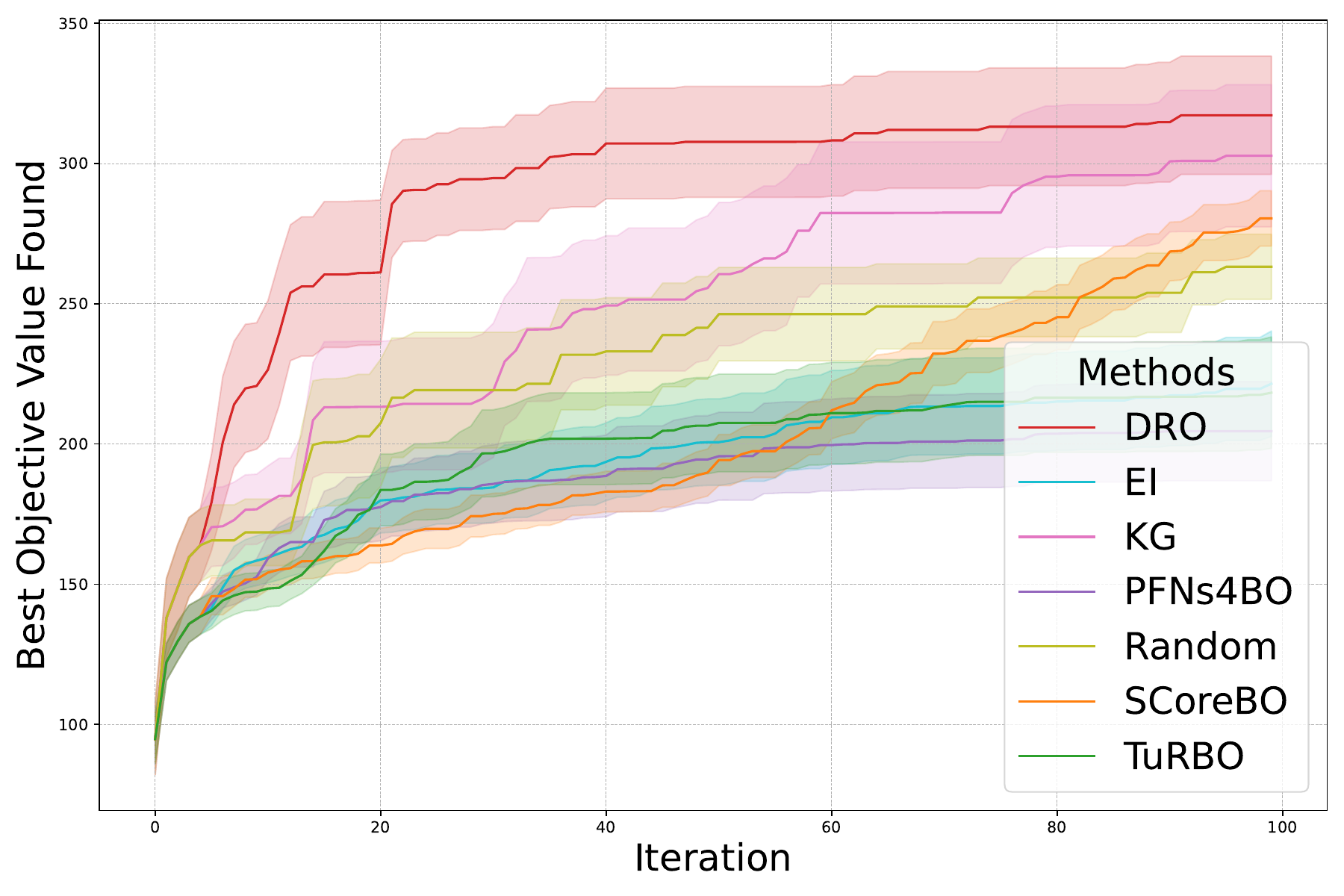}
        \caption{LunarLander Controller}
        \label{fig:lunarlander_main}
    \end{subfigure}
    \hfill
    \begin{subfigure}[b]{0.28\textwidth}
        \centering
        \includegraphics[width=\textwidth]{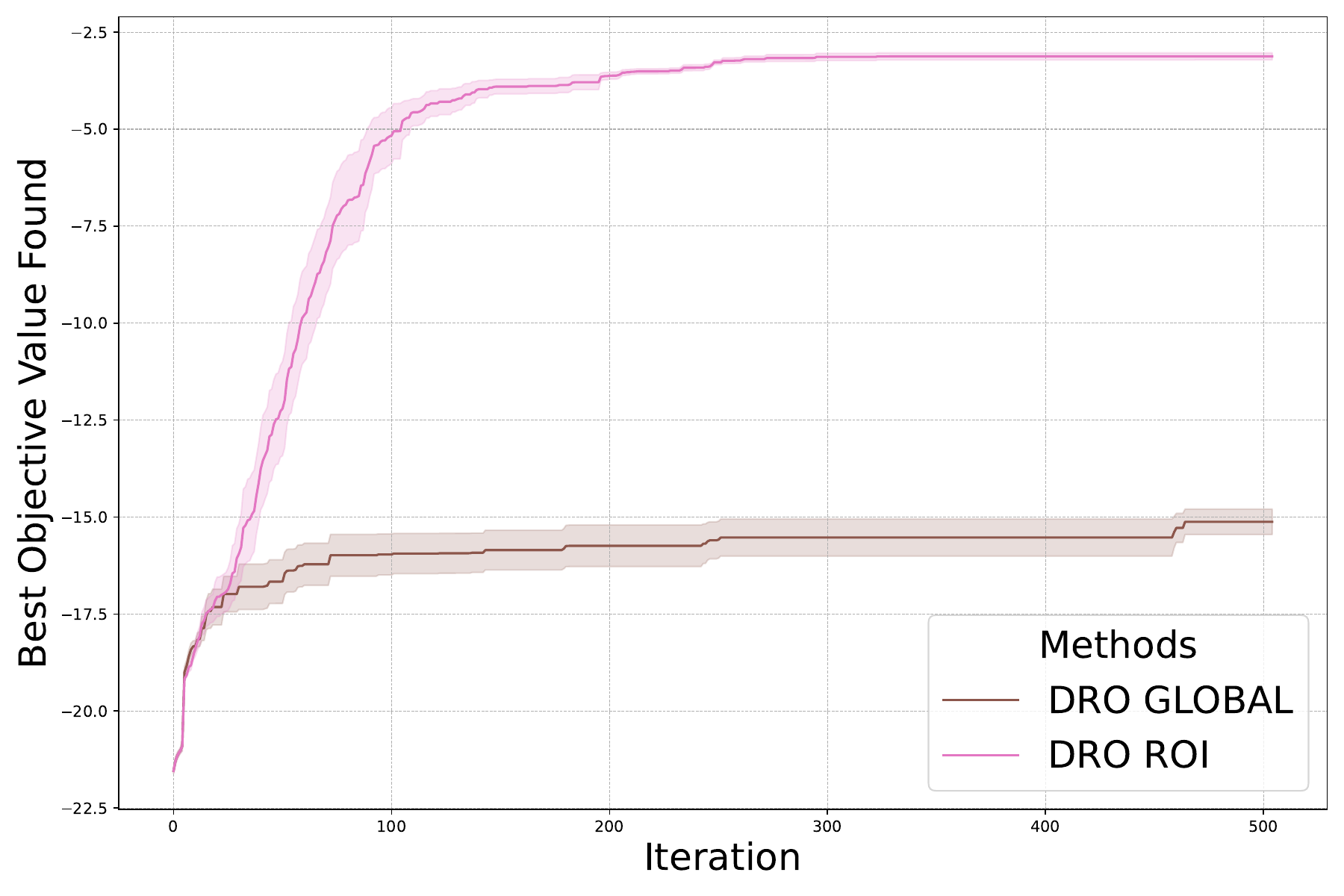}
        \caption{Effect of ROI Filtering}
        \label{fig:ackley_10d_roi_ablation_main}
    \end{subfigure}
    \hfill
    \begin{subfigure}[b]{0.28\textwidth}
        \centering
        \includegraphics[width=\textwidth]{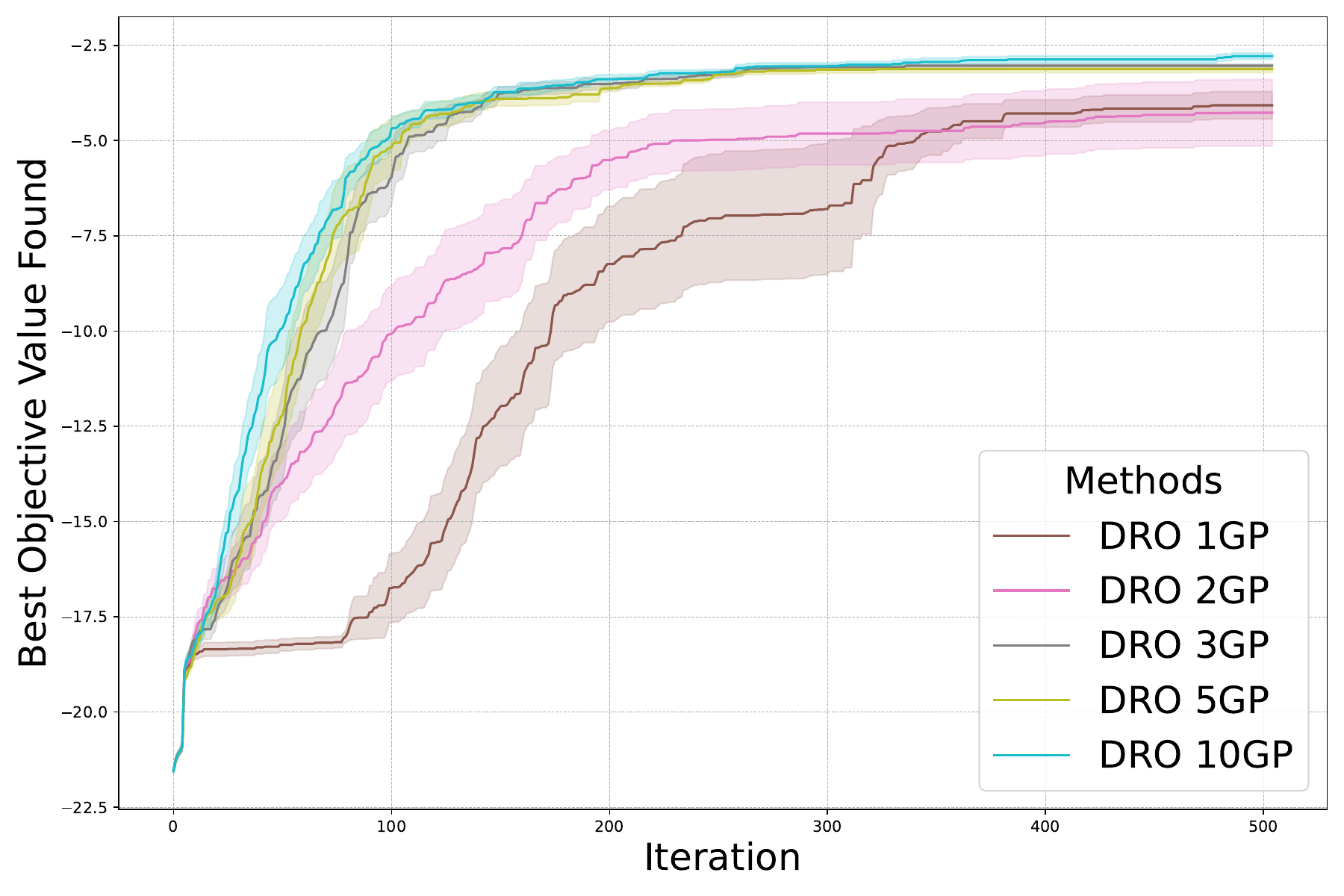}
        \caption{Effect of Ensemble Size ($M$)}
        \label{fig:ackley_10d_ensemble_study_main}
    \end{subfigure}
    \caption{Performance on LunarLander and ablation studies for DRO on Ackley 10D. Higher values are better (Best Objective Value Found). All DRO variants in ablations run for 10 trials.}
    \label{fig:ablation_and_lunar_results}
    % \vspace{-3mm}
\end{figure*}

\revision{This strong performance extends to practical applications. On the suite of HPO tasks (lower panel of \figref{fig:ackley_dim_hpo_study}), DRO consistently converges to a lower simple regret faster and more reliably than other methods for XGBoost tuning. A similar leading trend is observed for the Adam Wine HPO task, while on Adam Iris and Breast Cancer, DRO remains highly competitive, consistently ranking among the top-performing methods. Furthermore, in the LunarLander RL control task (\figref{fig:lunarlander_main}), DRO is numerically ahead of standard BO, PFNs4BO, SCoreBO and TuRBO at every checkpoint, although with ten trials the error bands of the strongest methods overlap, so we read this as a consistent rather than a statistically separated advantage.}

\subsection{Ablation Studies}
\label{sec:ablations}

\revision{To understand the drivers of DRO's performance, we conduct ablation studies on the Ackley 10D task (\figref{fig:ablation_and_lunar_results}). First, we examine the effect of ROI filtering (\figref{fig:ackley_10d_roi_ablation_main}) by comparing the standard DRO configuration (``DRO ROI'') against a variant that performs rollouts over the entire global search space (``DRO GLOBAL''). ``DRO ROI'' converges faster and to a much higher objective value (closer to 0, the optimum of Ackley), confirming that constraining simulations to promising regions is crucial.

Next, we investigate the influence of the number of GP models ($M$) in the ensemble (\figref{fig:ackley_10d_ensemble_study_main}). Performance improves with the ensemble size: ``DRO 10GP'' ($M=10$) is best, closely followed by $M=5$, and even $M=2$ or $3$ clearly outperform a single GP, which shows the value of model diversity for generating rich simulation data.}

\paragraph{Is the ROI or the learned policy responsible?} Removing each component in turn on Ackley 10D (\appref{subsec:attribution}) shows that neither suffices alone: at iteration 500, DRO-NoDT (ensemble, ROI and rotating pool, but no transformer) reaches $-10.01 \pm 2.56$, DRO-GLOBAL (transformer without ROI) $-15.12 \pm 0.33$, and full DRO $-3.12 \pm 0.10$. ROI-restricted logEI on the most likely member reaches $-12.11 \pm 2.49$ and is bimodal across seeds, with about half of the runs never leaving the initial plateau, whereas DRO succeeds on all ten. The policy is not an imitator of its pool either: rollouts driven by EI alone train a policy reaching $-2.72 \pm 0.11$ (\figref{fig:ackley_10d_acq_study}), while logEI itself plateaus at $-18.49 \pm 0.27$.

\paragraph{Composition of the rollout pool.} The non-myopic signal comes from the terminal-return labels, not from the behavior policy: adding an explicitly non-myopic two-step EI to the pool leaves DRO unchanged (paired $p=0.61$), while a purely exploratory posterior-variance acquisition hurts at an equal share ($-8.70 \pm 1.32$ versus $-3.12 \pm 0.10$) and is neutral at a $10\%$ share ($p=0.55$). A mixed-kernel ensemble with evidence-weighted rollout allocation matches the RBF-only control ($p=0.27$; \appref{subsec:pool}, \appref{subsec:kernel_diversity}).

\vspace{-.8mm}

\vspace{-.8mm}
\section{Conclusion and Future Work}
\label{sec:conclusion}

\revision{This paper introduces Direct Regret Optimization, a novel Bayesian Optimization framework that directly targets the minimization of final simple regret. By leveraging a Decision Transformer trained on simulated trajectories from an ensemble of GPs, and focusing exploration via an adaptive Region of Interest, DRO learns a non-myopic policy that is both sample-efficient and robust.

% \paragraph{Opportunities for future research and broader impacts.} 
Future work includes meta-learning from related tasks to complement the GP-sampled training data, richer hyperparameter marginalization (e.g., full MCMC) in place of the sparse lengthscale ensemble, auxiliary training signals such as ensemble disagreement or ROI membership and hindsight relabeling of stored trajectories, and the constrained and multi-fidelity extensions sketched in \appref{sec:dro_extensions}. This foundational algorithmic work has potential utility for scientific experimental design, and we foresee no direct negative societal impact.}

\paragraph{Limitations.} \appref{sec:theory} covers the simulation components, not the learned policy, for which no guarantee is given; non-myopia is a property of the training signal, not of explicit planning, and DRO can trail one-step lookahead baselines early; the simulation overhead is not negligible for cheap objectives; and the target return at inference is an optimistic estimate when the optimal value is unknown.

% AI Use Statement (carried over from the AISTATS 2027 version)
% Required by the AISTATS 2027 call for papers (updated 30 Sep 2026: the
% statement goes in the paper only). AUTHORS: confirm this wording before
% submission; it must match the AISTATS 2027 AI Policy for Authors.
\section*{AI Use Statement}
Generative AI tools were used to edit and tighten the prose of this manuscript, to check and format the bibliography entries against publisher records, and to draft and check the written form of the proofs in \appref{sec:theory} under the authors' direction. The research questions, the method, the experimental design, the code, and all reported numbers are the authors' own. All AI-assisted text and derivations were reviewed and verified by the authors, who take full responsibility for the final content of this work, including any text, claims, or artifacts produced with the aid of generative AI.

%%%%%%%%%%%%%%%%%%%%%%%%%%%%%%%%%%%%%%%%%%%%%%%%%%%%%%%%%%%%

% Bibliography
\bibliographystyle{plainnat}
\bibliography{references}

@article{Jones1998,
  title   = {Efficient Global Optimization of Expensive Black-Box Functions},
  author  = {Jones, Donald R. and Schonlau, Matthias and Welch, William J.},
  journal = {Journal of Global Optimization},
  volume  = {13},
  number  = {4},
  pages   = {455--492},
  year    = {1998},
  doi     = {10.1023/A:1008306431147}
}

@inproceedings{Srinivas2010,
  title     = {{G}aussian Process Optimization in the Bandit Setting: No Regret and Experimental Design},
  author    = {Srinivas, Niranjan and Krause, Andreas and Kakade, Sham M. and Seeger, Matthias W.},
  booktitle = {Proceedings of the 27th International Conference on Machine Learning (ICML)},
  pages     = {1015--1022},
  year      = {2010}
}

@article{Bergstra2012,
  title   = {Random Search for Hyper-Parameter Optimization},
  author  = {Bergstra, James and Bengio, Yoshua},
  journal = {Journal of Machine Learning Research},
  volume  = {13},
  pages   = {281--305},
  year    = {2012}
}

@article{Calandra2016,
  title   = {{B}ayesian Optimization for Learning Gaits under Uncertainty: An Experimental Comparison on a Dynamic Bipedal Walker},
  author  = {Calandra, Roberto and Seyfarth, Andr{\'e} and Peters, Jan and Deisenroth, Marc Peter},
  journal = {Annals of Mathematics and Artificial Intelligence},
  volume  = {76},
  number  = {1--2},
  pages   = {5--23},
  year    = {2016},
  doi     = {10.1007/s10472-015-9463-9}
}

@article{Shahriari2016,
  title     = {Taking the Human Out of the Loop: A Review of {B}ayesian Optimization},
  author    = {Shahriari, Bobak and Swersky, Kevin and Wang, Ziyu and Adams, Ryan P. and de Freitas, Nando},
  journal   = {Proceedings of the IEEE},
  volume    = {104},
  number    = {1},
  pages     = {148--175},
  year      = {2016},
  publisher = {IEEE}
}

@book{rasmussen2006gaussian,
  title     = {{G}aussian Processes for Machine Learning},
  author    = {Rasmussen, Carl Edward and Williams, Christopher K. I.},
  year      = {2006},
  publisher = {MIT Press},
  address   = {Cambridge, MA}
}

@inproceedings{snoek2012practical,
  title     = {Practical {B}ayesian Optimization of Machine Learning Algorithms},
  author    = {Snoek, Jasper and Larochelle, Hugo and Adams, Ryan P.},
  booktitle = {Advances in Neural Information Processing Systems 25 (NIPS)},
  pages     = {2951--2959},
  year      = {2012}
}

@article{brochu2010tutorial,
  title   = {A Tutorial on {B}ayesian Optimization of Expensive Cost Functions, with Application to Active User Modeling and Hierarchical Reinforcement Learning},
  author  = {Brochu, Eric and Cora, Vlad M. and de Freitas, Nando},
  journal = {arXiv preprint arXiv:1012.2599},
  year    = {2010}
}

@article{frazier2008knowledge,
  title     = {A Knowledge-Gradient Policy for Sequential Information Collection},
  author    = {Frazier, Peter I. and Powell, Warren B. and Dayanik, Savas},
  journal   = {SIAM Journal on Control and Optimization},
  volume    = {47},
  number    = {5},
  pages     = {2410--2439},
  year      = {2008},
  publisher = {SIAM}
}

@article{berkenkamp2019no,
  title   = {No-Regret {B}ayesian Optimization with Unknown Hyperparameters},
  author  = {Berkenkamp, Felix and Schoellig, Angela P. and Krause, Andreas},
  journal = {Journal of Machine Learning Research},
  volume  = {20},
  number  = {50},
  pages   = {1--24},
  year    = {2019}
}

@inproceedings{Chowdhury2017,
  title     = {On Kernelized Multi-armed Bandits},
  author    = {Chowdhury, Sayak Ray and Gopalan, Aditya},
  booktitle = {Proceedings of the 34th International Conference on Machine Learning (ICML)},
  series    = {Proceedings of Machine Learning Research},
  volume    = {70},
  pages     = {844--853},
  year      = {2017},
  publisher = {PMLR}
}

@article{Bull2011,
  title   = {Convergence Rates of Efficient Global Optimization Algorithms},
  author  = {Bull, Adam D.},
  journal = {Journal of Machine Learning Research},
  volume  = {12},
  number  = {88},
  pages   = {2879--2904},
  year    = {2011}
}

@inproceedings{nguyen2017regret,
  title     = {Regret for Expected Improvement over the Best-Observed Value and Stopping Condition},
  author    = {Nguyen, Vu and Gupta, Sunil and Rana, Santu and Li, Cheng and Venkatesh, Svetha},
  booktitle = {Proceedings of the 9th Asian Conference on Machine Learning (ACML)},
  series    = {Proceedings of Machine Learning Research},
  volume    = {77},
  pages     = {279--294},
  year      = {2017},
  publisher = {PMLR}
}

@techreport{ernesto2005mvf,
  title       = {{MVF} -- Multivariate Test Functions Library in {C} for Unconstrained Global Optimization},
  author      = {Adorio, Ernesto P.},
  institution = {Department of Mathematics, University of the Philippines Diliman},
  address     = {Quezon City, Philippines},
  year        = {2005}
}

@inproceedings{Lam2016,
  title     = {{B}ayesian Optimization with a Finite Budget: An Approximate Dynamic Programming Approach},
  author    = {Lam, R{\'e}mi R. and Willcox, Karen E. and Wolpert, David H.},
  booktitle = {Advances in Neural Information Processing Systems 29 (NIPS)},
  pages     = {883--891},
  year      = {2016}
}

@inproceedings{Hsieh2021,
  title     = {Reinforced Few-Shot Acquisition Function Learning for {B}ayesian Optimization},
  author    = {Hsieh, Bing-Jing and Hsieh, Ping-Chun and Liu, Xi},
  booktitle = {Advances in Neural Information Processing Systems 34 (NeurIPS)},
  year      = {2021}
}

@article{Liu2022,
  title   = {Robust {B}ayesian Optimization with Reinforcement Learned Acquisition Functions},
  author  = {Liu, Zijing and Qu, Xiyao and Liu, Xuejun and Lyu, Hongqiang},
  journal = {arXiv preprint arXiv:2210.00476},
  year    = {2022}
}

@article{Cheon2022,
  title   = {Reinforcement Learning Based Multi-Step Look-Ahead {B}ayesian Optimization},
  author  = {Cheon, Mujin and Byun, Haeun and Lee, Jay H.},
  journal = {IFAC-PapersOnLine},
  volume  = {55},
  number  = {7},
  pages   = {100--105},
  year    = {2022},
  doi     = {10.1016/j.ifacol.2022.07.428}
}

@inproceedings{Cheon2024,
  title     = {{EARL-BO}: Reinforcement Learning for Multi-Step Lookahead, High-Dimensional {B}ayesian Optimization},
  author    = {Cheon, Mujin and Lee, Jay H. and Koh, Dong-Yeun and Tsay, Calvin},
  booktitle = {Proceedings of the 42nd International Conference on Machine Learning (ICML)},
  series    = {Proceedings of Machine Learning Research},
  volume    = {267},
  pages     = {10182--10198},
  year      = {2025},
  publisher = {PMLR}
}

@inproceedings{Huang2024,
  title     = {{RLBOF}: Reinforcement Learning from {B}ayesian Optimization Feedback},
  author    = {Huang, Hailong and Liang, Xiubo and Zhang, Quanwei and Wang, Hongzhi and Li, Xiangdong},
  booktitle = {Proceedings of the International Joint Conference on Neural Networks (IJCNN)},
  year      = {2024},
  publisher = {IEEE},
  doi       = {10.1109/IJCNN60899.2024.10650719}
}

@article{Deutel2025,
  title   = {Combining Multi-Objective {B}ayesian Optimization with Reinforcement Learning for {TinyML}},
  author  = {Deutel, Mark and Kontes, Georgios and Mutschler, Christopher and Teich, J{\"u}rgen},
  journal = {ACM Transactions on Evolutionary Learning and Optimization},
  volume  = {5},
  number  = {3},
  articleno = {17},
  pages   = {1--21},
  year    = {2025},
  doi     = {10.1145/3715012}
}

@inproceedings{chen2021decision,
  title     = {Decision Transformer: Reinforcement Learning via Sequence Modeling},
  author    = {Chen, Lili and Lu, Kevin and Rajeswaran, Aravind and Lee, Kimin and Grover, Aditya and Laskin, Misha and Abbeel, Pieter and Srinivas, Aravind and Mordatch, Igor},
  booktitle = {Advances in Neural Information Processing Systems 34 (NeurIPS)},
  pages     = {15084--15097},
  year      = {2021}
}

@inproceedings{Tripp2020,
  title     = {Sample-Efficient Optimization in the Latent Space of Deep Generative Models via Weighted Retraining},
  author    = {Tripp, Austin and Daxberger, Erik and Hern{\'a}ndez-Lobato, Jos{\'e} Miguel},
  booktitle = {Advances in Neural Information Processing Systems 33 (NeurIPS)},
  year      = {2020}
}

@article{Kim2021,
  title   = {Deep Learning for {B}ayesian Optimization of Scientific Problems with High-Dimensional Structure},
  author  = {Kim, Samuel and Lu, Peter Y. and Loh, Charlotte and Smith, Jamie and Snoek, Jasper and Solja{\v c}i{\'c}, Marin},
  journal = {Transactions on Machine Learning Research},
  issn    = {2835-8856},
  year    = {2022}
}

@inproceedings{Zhang2023ROI,
  title     = {Learning Regions of Interest for {B}ayesian Optimization with Adaptive Level-Set Estimation},
  author    = {Zhang, Fengxue and Song, Jialin and Bowden, James C. and Ladd, Alexander and Yue, Yisong and Desautels, Thomas and Chen, Yuxin},
  booktitle = {Proceedings of the 40th International Conference on Machine Learning (ICML)},
  series    = {Proceedings of Machine Learning Research},
  volume    = {202},
  pages     = {41579--41595},
  year      = {2023},
  publisher = {PMLR}
}

@inproceedings{muller2023pfns4bo,
  title     = {{PFNs4BO}: In-Context Learning for {B}ayesian Optimization},
  author    = {M{\"u}ller, Samuel and Feurer, Matthias and Hollmann, Noah and Hutter, Frank},
  booktitle = {Proceedings of the 40th International Conference on Machine Learning (ICML)},
  series    = {Proceedings of Machine Learning Research},
  volume    = {202},
  pages     = {25444--25470},
  year      = {2023},
  publisher = {PMLR}
}

@inproceedings{nagler2023statistical,
  title     = {Statistical Foundations of Prior-Data Fitted Networks},
  author    = {Nagler, Thomas},
  booktitle = {Proceedings of the 40th International Conference on Machine Learning (ICML)},
  series    = {Proceedings of Machine Learning Research},
  volume    = {202},
  pages     = {25660--25676},
  year      = {2023},
  publisher = {PMLR}
}

@inproceedings{muller2021transformers,
  title     = {Transformers Can Do {B}ayesian Inference},
  author    = {M{\"u}ller, Samuel and Hollmann, Noah and Pineda Arango, Sebastian and Grabocka, Josif and Hutter, Frank},
  booktitle = {International Conference on Learning Representations (ICLR)},
  year      = {2022}
}

@inproceedings{maraval2023end,
  title     = {End-to-End Meta-{B}ayesian Optimisation with Transformer Neural Processes},
  author    = {Maraval, Alexandre and Zimmer, Matthieu and Grosnit, Antoine and Bou Ammar, Haitham},
  booktitle = {Advances in Neural Information Processing Systems 36 (NeurIPS)},
  pages     = {11246--11260},
  year      = {2023}
}

@inproceedings{chen2022towards,
  title     = {Towards Learning Universal Hyperparameter Optimizers with Transformers},
  author    = {Chen, Yutian and Song, Xingyou and Lee, Chansoo and Wang, Zi and Zhang, Qiuyi and Dohan, David and Kawakami, Kazuya and Kochanski, Greg and Doucet, Arnaud and Ranzato, Marc'Aurelio and Perel, Sagi and de Freitas, Nando},
  booktitle = {Advances in Neural Information Processing Systems 35 (NeurIPS)},
  pages     = {32053--32068},
  year      = {2022}
}

@inproceedings{garnelo2018conditional,
  title     = {Conditional Neural Processes},
  author    = {Garnelo, Marta and Rosenbaum, Dan and Maddison, Christopher and Ramalho, Tiago and Saxton, David and Shanahan, Murray and Teh, Yee Whye and Rezende, Danilo and Eslami, S. M. Ali},
  booktitle = {Proceedings of the 35th International Conference on Machine Learning (ICML)},
  series    = {Proceedings of Machine Learning Research},
  volume    = {80},
  pages     = {1704--1713},
  year      = {2018},
  publisher = {PMLR}
}

@inproceedings{garnelo2018neural,
  title     = {Neural Processes},
  author    = {Garnelo, Marta and Schwarz, Jonathan and Rosenbaum, Dan and Viola, Fabio and Rezende, Danilo J. and Eslami, S. M. Ali and Teh, Yee Whye},
  booktitle = {ICML 2018 Workshop on Theoretical Foundations and Applications of Deep Generative Models},
  year      = {2018},
  note      = {arXiv:1807.01622}
}

@inproceedings{kim2019attentive,
  title     = {Attentive Neural Processes},
  author    = {Kim, Hyunjik and Mnih, Andriy and Schwarz, Jonathan and Garnelo, Marta and Eslami, Ali and Rosenbaum, Dan and Vinyals, Oriol and Teh, Yee Whye},
  booktitle = {International Conference on Learning Representations (ICLR)},
  year      = {2019}
}

@inproceedings{swersky2020amortized,
  title     = {Amortized {B}ayesian Optimization over Discrete Spaces},
  author    = {Swersky, Kevin and Rubanova, Yulia and Dohan, David and Murphy, Kevin},
  booktitle = {Proceedings of the 36th Conference on Uncertainty in Artificial Intelligence (UAI)},
  series    = {Proceedings of Machine Learning Research},
  volume    = {124},
  pages     = {769--778},
  year      = {2020},
  publisher = {PMLR}
}

@inproceedings{volpp2020meta,
  title     = {Meta-Learning Acquisition Functions for Transfer Learning in {B}ayesian Optimization},
  author    = {Volpp, Michael and Fr{\"o}hlich, Lukas P. and Fischer, Kirsten and Doerr, Andreas and Falkner, Stefan and Hutter, Frank and Daniel, Christian},
  booktitle = {International Conference on Learning Representations (ICLR)},
  year      = {2020}
}

@inproceedings{chen2017learning,
  title     = {Learning to Learn without Gradient Descent by Gradient Descent},
  author    = {Chen, Yutian and Hoffman, Matthew W. and Colmenarejo, Sergio G{\'o}mez and Denil, Misha and Lillicrap, Timothy P. and Botvinick, Matt and de Freitas, Nando},
  booktitle = {Proceedings of the 34th International Conference on Machine Learning (ICML)},
  series    = {Proceedings of Machine Learning Research},
  volume    = {70},
  pages     = {748--756},
  year      = {2017},
  publisher = {PMLR}
}

@inproceedings{foster2021deep,
  title     = {Deep Adaptive Design: Amortizing Sequential {B}ayesian Experimental Design},
  author    = {Foster, Adam and Ivanova, Desi R. and Malik, Ilyas and Rainforth, Tom},
  booktitle = {Proceedings of the 38th International Conference on Machine Learning (ICML)},
  series    = {Proceedings of Machine Learning Research},
  volume    = {139},
  pages     = {3384--3395},
  year      = {2021},
  publisher = {PMLR}
}

@inproceedings{ivanova2021implicit,
  title     = {Implicit Deep Adaptive Design: Policy-Based Experimental Design without Likelihoods},
  author    = {Ivanova, Desi R. and Foster, Adam and Kleinegesse, Steven and Gutmann, Michael U. and Rainforth, Tom},
  booktitle = {Advances in Neural Information Processing Systems 34 (NeurIPS)},
  pages     = {25785--25798},
  year      = {2021}
}

@inproceedings{blau2022optimizing,
  title     = {Optimizing Sequential Experimental Design with Deep Reinforcement Learning},
  author    = {Blau, Tom and Bonilla, Edwin V. and Chades, Iadine and Dezfouli, Amir},
  booktitle = {Proceedings of the 39th International Conference on Machine Learning (ICML)},
  series    = {Proceedings of Machine Learning Research},
  volume    = {162},
  pages     = {2107--2128},
  year      = {2022},
  publisher = {PMLR}
}

@inproceedings{huang2024amortized,
  title     = {Amortized {B}ayesian Experimental Design for Decision-Making},
  author    = {Huang, Daolang and Guo, Yujia and Acerbi, Luigi and Kaski, Samuel},
  booktitle = {Advances in Neural Information Processing Systems 37 (NeurIPS)},
  year      = {2024}
}

@inproceedings{chang2025amortized,
  title     = {Amortized Probabilistic Conditioning for Optimization, Simulation and Inference},
  author    = {Chang, Paul E. and Loka, Nasrulloh and Huang, Daolang and Remes, Ulpu and Kaski, Samuel and Acerbi, Luigi},
  booktitle = {Proceedings of the 28th International Conference on Artificial Intelligence and Statistics (AISTATS)},
  series    = {Proceedings of Machine Learning Research},
  volume    = {258},
  pages     = {703--711},
  year      = {2025},
  publisher = {PMLR}
}

@inproceedings{zhang2025pabbo,
  title     = {{PABBO}: Preferential Amortized Black-Box Optimization},
  author    = {Zhang, Xinyu and Huang, Daolang and Kaski, Samuel and Martinelli, Julien},
  booktitle = {International Conference on Learning Representations (ICLR)},
  year      = {2025}
}

@article{kennedy2000predicting,
  title     = {Predicting the Output from a Complex Computer Code when Fast Approximations Are Available},
  author    = {Kennedy, Marc C. and O'Hagan, Anthony},
  journal   = {Biometrika},
  volume    = {87},
  number    = {1},
  pages     = {1--13},
  year      = {2000},
  publisher = {Oxford University Press}
}

@inproceedings{kandasamy2016gaussian,
  title     = {{G}aussian Process Bandit Optimisation with Multi-Fidelity Evaluations},
  author    = {Kandasamy, Kirthevasan and Dasarathy, Gautam and Oliva, Junier B. and Schneider, Jeff and P{\'o}czos, Barnab{\'a}s},
  booktitle = {Advances in Neural Information Processing Systems 29 (NIPS)},
  pages     = {992--1000},
  year      = {2016}
}

@inproceedings{kandasamy2017multi,
  title     = {Multi-Fidelity {B}ayesian Optimisation with Continuous Approximations},
  author    = {Kandasamy, Kirthevasan and Dasarathy, Gautam and Schneider, Jeff and P{\'o}czos, Barnab{\'a}s},
  booktitle = {Proceedings of the 34th International Conference on Machine Learning (ICML)},
  series    = {Proceedings of Machine Learning Research},
  volume    = {70},
  pages     = {1799--1808},
  year      = {2017},
  publisher = {PMLR}
}

@inproceedings{takeno2020multi,
  title     = {Multi-Fidelity {B}ayesian Optimization with Max-Value Entropy Search and Its Parallelization},
  author    = {Takeno, Shion and Fukuoka, Hitoshi and Tsukada, Yuhki and Koyama, Toshiyuki and Shiga, Motoki and Takeuchi, Ichiro and Karasuyama, Masayuki},
  booktitle = {Proceedings of the 37th International Conference on Machine Learning (ICML)},
  series    = {Proceedings of Machine Learning Research},
  volume    = {119},
  pages     = {9334--9345},
  year      = {2020},
  publisher = {PMLR}
}

@inproceedings{zhang2025robust_mfbo,
  title     = {Robust Multi-Fidelity {B}ayesian Optimization with Deep Kernel and Partition},
  author    = {Zhang, Fengxue and Desautels, Thomas and Chen, Yuxin},
  booktitle = {Proceedings of the 28th International Conference on Artificial Intelligence and Statistics (AISTATS)},
  series    = {Proceedings of Machine Learning Research},
  volume    = {258},
  pages     = {2683--2691},
  year      = {2025},
  publisher = {PMLR}
}

@article{wistuba2018scalable,
  title   = {Scalable {G}aussian Process-Based Transfer Surrogates for Hyperparameter Optimization},
  author  = {Wistuba, Martin and Schilling, Nicolas and Schmidt-Thieme, Lars},
  journal = {Machine Learning},
  volume  = {107},
  number  = {1},
  pages   = {43--78},
  year    = {2018},
  doi     = {10.1007/s10994-017-5684-y}
}

@inproceedings{hvarfner2023self,
  title     = {Self-Correcting {B}ayesian Optimization through {B}ayesian Active Learning},
  author    = {Hvarfner, Carl and Hellsten, Erik and Hutter, Frank and Nardi, Luigi},
  booktitle = {Advances in Neural Information Processing Systems 36 (NeurIPS)},
  pages     = {79173--79199},
  year      = {2023}
}

@inproceedings{hoffman2011portfolio,
  title     = {Portfolio Allocation for {B}ayesian Optimization},
  author    = {Hoffman, Matthew and Brochu, Eric and de Freitas, Nando},
  booktitle = {Proceedings of the 27th Conference on Uncertainty in Artificial Intelligence (UAI)},
  pages     = {327--336},
  year      = {2011}
}

@article{martinez2018funneled,
  title     = {Funneled {B}ayesian Optimization for Design, Tuning and Control of Autonomous Systems},
  author    = {Martinez-Cantin, Ruben},
  journal   = {IEEE Transactions on Cybernetics},
  volume    = {49},
  number    = {4},
  pages     = {1489--1500},
  year      = {2019},
  publisher = {IEEE}
}

@article{cowen2022hebo,
  title   = {{HEBO}: Pushing the Limits of Sample-Efficient Hyper-Parameter Optimisation},
  author  = {Cowen-Rivers, Alexander I. and Lyu, Wenlong and Tutunov, Rasul and Wang, Zhi and Grosnit, Antoine and Griffiths, Ryan Rhys and Maraval, Alexandre Max and Hao, Jianye and Wang, Jun and Peters, Jan and Bou-Ammar, Haitham},
  journal = {Journal of Artificial Intelligence Research},
  volume  = {74},
  pages   = {1269--1349},
  year    = {2022}
}

@misc{head2021scikit,
  title        = {scikit-optimize/scikit-optimize},
  author       = {Head, Tim and Kumar, Manoj and Nahrstaedt, Holger and Louppe, Gilles and Shcherbatyi, Iaroslav},
  year         = {2021},
  howpublished = {Zenodo},
  doi          = {10.5281/zenodo.5565057}
}

@inproceedings{zhang2024finding,
  title     = {Finding Interior Optimum of Black-Box Constrained Objective with {B}ayesian Optimization},
  author    = {Zhang, Fengxue and Chen, Yuxin},
  booktitle = {Proceedings of the 41st Conference on Uncertainty in Artificial Intelligence (UAI)},
  series    = {Proceedings of Machine Learning Research},
  volume    = {286},
  pages     = {5000--5029},
  year      = {2025},
  publisher = {PMLR}
}

@inproceedings{li2022gaussian,
  title     = {{G}aussian Process Bandit Optimization with Few Batches},
  author    = {Li, Zihan and Scarlett, Jonathan},
  booktitle = {Proceedings of the 25th International Conference on Artificial Intelligence and Statistics (AISTATS)},
  series    = {Proceedings of Machine Learning Research},
  volume    = {151},
  pages     = {92--107},
  year      = {2022},
  publisher = {PMLR}
}

@inproceedings{pmlr-v202-xu23h,
  title     = {Constrained Efficient Global Optimization of Expensive Black-Box Functions},
  author    = {Xu, Wenjie and Jiang, Yuning and Svetozarevic, Bratislav and Jones, Colin},
  booktitle = {Proceedings of the 40th International Conference on Machine Learning (ICML)},
  series    = {Proceedings of Machine Learning Research},
  volume    = {202},
  pages     = {38485--38498},
  year      = {2023},
  publisher = {PMLR}
}

@inproceedings{eriksson2019scalable,
  title     = {Scalable Global Optimization via Local {B}ayesian Optimization},
  author    = {Eriksson, David and Pearce, Michael and Gardner, Jacob and Turner, Ryan D. and Poloczek, Matthias},
  booktitle = {Advances in Neural Information Processing Systems 32 (NeurIPS)},
  year      = {2019}
}

@inproceedings{eriksson2021scalable,
  title     = {Scalable Constrained {B}ayesian Optimization},
  author    = {Eriksson, David and Poloczek, Matthias},
  booktitle = {Proceedings of the 24th International Conference on Artificial Intelligence and Statistics (AISTATS)},
  series    = {Proceedings of Machine Learning Research},
  volume    = {130},
  pages     = {730--738},
  year      = {2021},
  publisher = {PMLR}
}

@inproceedings{towers2024gymnasium,
  title     = {Gymnasium: A Standard Interface for Reinforcement Learning Environments},
  author    = {Towers, Mark and Kwiatkowski, Ariel and Balis, John U. and De Cola, Gianluca and Deleu, Tristan and Goul{\~a}o, Manuel and Kallinteris, Andreas and Krimmel, Markus and KG, Arjun and Perez-Vicente, Rodrigo and Terry, Jordan K. and Pierr{\'e}, Andrea and Schulhoff, Sander and Tai, Jun Jet and Tan, Hannah and Younis, Omar G.},
  booktitle = {Advances in Neural Information Processing Systems 38 (NeurIPS), Datasets and Benchmarks Track},
  year      = {2025}
}

@inproceedings{ament2023unexpected,
  title     = {Unexpected Improvements to Expected Improvement for {B}ayesian Optimization},
  author    = {Ament, Sebastian and Daulton, Samuel and Eriksson, David and Balandat, Maximilian and Bakshy, Eytan},
  booktitle = {Advances in Neural Information Processing Systems 36 (NeurIPS)},
  pages     = {20577--20612},
  year      = {2023}
}

@inproceedings{balandat2020botorch,
  title     = {{BoTorch}: A Framework for Efficient {M}onte-{C}arlo {B}ayesian Optimization},
  author    = {Balandat, Maximilian and Karrer, Brian and Jiang, Daniel and Daulton, Samuel and Letham, Ben and Wilson, Andrew G. and Bakshy, Eytan},
  booktitle = {Advances in Neural Information Processing Systems 33 (NeurIPS)},
  pages     = {21524--21538},
  year      = {2020}
}

@inproceedings{pfisterer2022yahpo,
  title     = {{YAHPO} Gym -- An Efficient Multi-Objective Multi-Fidelity Benchmark for Hyperparameter Optimization},
  author    = {Pfisterer, Florian and Schneider, Lennart and Moosbauer, Julia and Binder, Martin and Bischl, Bernd},
  booktitle = {Proceedings of the First International Conference on Automated Machine Learning (AutoML)},
  series    = {Proceedings of Machine Learning Research},
  volume    = {188},
  pages     = {3/1--39},
  year      = {2022},
  publisher = {PMLR}
}

@inproceedings{gonzalez2016glasses,
  title     = {{GLASSES}: Relieving the Myopia of {B}ayesian Optimisation},
  author    = {Gonz{\'a}lez, Javier and Osborne, Michael and Lawrence, Neil D.},
  booktitle = {Proceedings of the 19th International Conference on Artificial Intelligence and Statistics (AISTATS)},
  series    = {Proceedings of Machine Learning Research},
  volume    = {51},
  pages     = {790--799},
  year      = {2016},
  publisher = {PMLR}
}

@inproceedings{wu2019practical,
  title     = {Practical Two-Step Lookahead {B}ayesian Optimization},
  author    = {Wu, Jian and Frazier, Peter I.},
  booktitle = {Advances in Neural Information Processing Systems 32 (NeurIPS)},
  year      = {2019}
}

@inproceedings{jiang2020efficient,
  title     = {Efficient Nonmyopic {B}ayesian Optimization via One-Shot Multi-Step Trees},
  author    = {Jiang, Shali and Jiang, Daniel R. and Balandat, Maximilian and Karrer, Brian and Gardner, Jacob R. and Garnett, Roman},
  booktitle = {Advances in Neural Information Processing Systems 33 (NeurIPS)},
  pages     = {18039--18049},
  year      = {2020}
}

% Appendices
\clearpage
\appendix
\appendix
\newpage
\section{Detailed Discussion on Literature} % Corrected "Detiailed"
\label{sec:lit-review}

% --- Existing RL Subsection ---
\subsection{Reinforcement Learning in Bayesian Optimization}
Traditional Bayesian optimization (BO) frameworks rely on fixed, hand-crafted acquisition functions to determine query points, balancing exploration and exploitation of the unknown objective \citep{Jones1998,Srinivas2010}. Classical acquisition functions such as Expected Improvement or upper confidence bounds are designed a priori and remain static during the optimization. Recent work has proposed integrating reinforcement learning (RL) into BO to learn a sampling strategy dynamically rather than using a fixed rule. By formulating the selection of sample points as a sequential decision-making problem, an RL agent can adaptively improve the sampling policy based on feedback from previous observations. This approach aims to boost efficiency and outcomes by learning to trade off exploration and exploitation in an adaptive, data-driven manner.

\paragraph{RL-enhanced acquisition function strategies.}
One line of research uses RL to augment or replace the acquisition function itself, enabling dynamic selection of where to sample next. \citet{Hsieh2021} introduced a reinforced few-shot acquisition function learning (FSAF) method that treats the acquisition function as a Q-value function. They train a deep Q-network (DQN) to act as a surrogate acquisition function, using meta-learning to allow quick adaptation to new tasks. By employing a Bayesian variant of DQN (to maintain uncertainty estimates) and fine-tuning with limited data, FSAF achieved comparable or better performance than state-of-the-art hand-crafted acquisitions.

Similarly, \citet{Liu2022} proposed an RL-assisted BO framework (RLABO) that formalizes acquisition function selection as a Markov decision process. In RLABO, an RL agent is trained to choose among multiple acquisition functions at each BO iteration based on a reward signal (e.g., improvement in objective value). This learned policy adaptively balances exploration vs. exploitation in real time, outperforming any single fixed heuristic across several benchmark optimization tasks.

\paragraph{RL for non-myopic (multi-step) Bayesian optimization.}
Another avenue is using RL to extend BO beyond myopic, one-step decisions by planning multi-step lookahead strategies. \citet{Cheon2022} provided early evidence that an RL-driven policy can outperform the greedy one-step Expected Improvement strategy by optimizing over a multi-step horizon. Their approach treated BO as a sequential decision process and showed that a learned policy can yield better long-term outcomes than myopic acquisition functions.

Building on this idea, \citet{Cheon2024} recently introduced EARL-BO (Encoder-Augmented RL for BO), an RL framework for multi-step lookahead in high-dimensional optimization problems. EARL-BO represents the state of the BO (e.g., the surrogate model posterior and remaining budget) with an attention-based encoder, and it uses off-policy RL to efficiently learn a near-optimal sampling strategy over multiple future steps. This method significantly improved performance compared to traditional one-step BO and earlier limited lookahead (rollout) approaches, as demonstrated on synthetic benchmarks and real-world hyperparameter optimization tasks.

\paragraph{RL-integrated surrogate modeling and meta-BO.}
Beyond guiding point selection, researchers have also integrated RL into surrogate modeling and meta-optimization aspects of BO. \citet{Huang2024} propose RLBOF (Reinforcement Learning from Bayesian Optimization Feedback), where a reinforcement learning agent (based on proximal policy optimization) continuously adjusts the surrogate model using feedback from the BO process. In this approach, the RL agent learns to tune the surrogate model's parameters or training procedure to better fit the observations and improve prediction of the objective function. By optimizing the surrogate model itself via RL (in tandem with the standard BO loop), this method enhances the overall BO performance. Such integration of RL at the meta-level (sometimes called meta-BO) essentially allows the BO algorithm to learn how to learn---not only optimizing the decisions of where to sample, but also refining the modeling of the objective based on experience.

\paragraph{Applications and multi-objective extensions.}
The combination of RL and BO has been applied in various domains and extended to multi-objective optimization problems. For example, \citet{Deutel2025} develop an RL-assisted multi-objective BO approach for TinyML neural architecture search. Their method combines multi-objective BO with an ensemble of RL policies (trained via augmented random search) to efficiently explore trade-offs between model accuracy and resource constraints. This RL-augmented strategy was able to find better Pareto-optimal neural network architectures (balancing accuracy, memory, and latency) than conventional multi-objective BO techniques. Similarly, RL-integrated BO frameworks have been explored for automated hyperparameter tuning, engineering design optimization, and other real-world scenarios. In these applications, the adaptive exploration-exploitation control provided by RL often yields faster convergence and more robust solutions compared to static BO policies.

\paragraph{Key takeaways.}
In summary, integrating reinforcement learning into Bayesian optimization enables several key advancements over traditional BO. First, RL allows the acquisition function or strategy to be adaptive rather than fixed, selecting sampling actions based on learned experience. Second, RL-based policies can consider multi-step outcomes, mitigating the greedy, myopic nature of standard one-step BO by planning further ahead. Third, RL can be integrated into different parts of the BO pipeline (acquisition selection, surrogate model tuning, or meta-learning), effectively allowing the optimizer to improve itself over time. Across various studies and applications, RL-enhanced BO methods consistently outperform fixed heuristic strategies, demonstrating improved efficiency in finding optimal solutions for both single-objective and multi-objective problems.
% Removed todo about model misspecification, assuming it's covered sufficiently or not critical for this expanded review.

% --- Existing Deep Learning Subsection (Expanded) ---
\subsection{Deep Learning Approaches in Bayesian Optimization}
\label{subsec:deep-learning-approaches}

Recent years have seen a surge of interest in integrating deep learning techniques into Bayesian optimization to improve flexibility, scalability, and the capacity to handle more complex objective landscapes. These approaches often replace or augment the traditional Gaussian Process surrogate and hand-crafted acquisition functions with neural network models that can learn from data and past optimization experience. Below, we highlight several representative methods that combine neural networks with BO and compare them to our Decision Transformer-based strategy.

\paragraph{Optimization in latent spaces via weighted retraining.}
\citet{Tripp2020} explore Bayesian optimization in the latent space of deep generative models. Here, a generative network (e.g., a variational autoencoder) learns a lower-dimensional embedding of the input space. BO is then conducted \emph{within this latent space}, where the landscape is often smoother or easier to navigate, before mapping latent representations back to the original domain. To keep the latent space relevant to high-performing solutions, they employ a \emph{weighted retraining} scheme that prioritizes observed samples with superior objective values. Compared to our Decision Transformer-based approach, which focuses on learning a policy to make direct query decisions, \citeauthor{Tripp2020} focus on learning a \emph{representation} of the search space. Both methods harness deep learning for higher-level capabilities---either representation learning or policy learning---beyond classical GP-based BO pipelines.

\paragraph{Deep surrogate models for high-dimensional BO.}
Other recent work highlights how neural networks can be leveraged to improve surrogate modeling in high-dimensional or structured scientific problems. For instance, \citet{Kim2021} propose a deep learning framework tailored for Bayesian optimization tasks with high-dimensional structure, using neural networks to capture complex correlations and accelerate convergence in scientific applications. Their approach demonstrates how deep models can scale to large design spaces that would challenge traditional GPs. Similarly, \citet{Zhang2023ROI} introduce a neural network-based adaptive level-set estimation method for BO, termed ``Learning Regions of Interest.'' By identifying and focusing on promising sub-regions of the search space, this framework refines the surrogate model in a targeted manner, effectively allocating sampling budget to areas with higher potential for improvement. These methods align with our emphasis on harnessing deep models to handle challenging search landscapes and guide the BO process.

\paragraph{PFNs4BO and Prior-Data Fitted Networks.}
One notable recent direction is the use of \emph{Prior-Data Fitted Networks (PFNs)} for BO, exemplified by PFNs4BO~\citep{muller2023pfns4bo}. PFNs leverage transformers to \emph{amortize} Bayesian inference by meta-training on synthetic tasks drawn from a chosen prior distribution~\citep{nagler2023statistical}. In PFNs4BO, a transformer is pre-trained (offline) to approximate the posterior predictive distribution (PPD) of the objective function in-context; at runtime, this single model produces posterior predictions for any new BO dataset in one forward pass~\citep{nagler2023statistical,muller2021transformers}. This yields extremely fast surrogate updates --- the PFN surrogate does \textbf{not} require retraining at each BO iteration, unlike a Gaussian Process which must be refit or updated as data accumulates. Moreover, by training across many sampled functions (e.g., varying GP hyperparameters or even Bayesian neural network priors), the PFN inherently marginalizes over surrogate uncertainty, effectively performing Bayesian model averaging under the specified prior. Empirically, a single PFN can mimic a GP's posterior almost exactly while attaining orders-of-magnitude speedups, and it flexibly accommodates richer prior knowledge than classical kernels (for instance, one can embed hints of likely optima or ignore irrelevant dimensions). Notably, PFNs4BO can even learn the acquisition function as part of its training (enabling non-myopic policies) by incorporating the selection strategy into the prior/task simulation~\citep{muller2023pfns4bo}.

\paragraph{Meta-learning surrogates and acquisition functions with neural processes.}
Beyond PFNs, another line of work leverages meta-learning principles, often employing neural processes (NPs) and their variants (e.g., NPs, Conditional NPs, Attentive NPs) \citep{garnelo2018neural, garnelo2018conditional, kim2019attentive}, to accelerate BO. NPs are deep models trained to approximate distributions over functions, conditioning on context sets (observed data points). In the meta-BO setting \citep{swersky2020amortized}, NPs can be pre-trained on data from related optimization tasks. When deployed on a new task, the NP surrogate can rapidly adapt its predictions based on the initial samples, providing a warm start compared to training a GP or standard neural network from scratch. Some approaches jointly learn the surrogate model (like an NP or transformer-based model) and an acquisition function (or an ensemble of them) in the meta-training phase \citep{volpp2020meta, chen2022towards}. End-to-end meta-BO frameworks like OptFormer \citep{chen2022towards} and Neural Acquisition Process (NAP) \citep{maraval2023end} also leverage transformers to learn universal optimizers that generalize across tasks. % \todo{Elaborate briefly on OptFormer/NAP if desired, or ensure this mention suffices.}
This allows the BO system to learn not just how to model functions typical of the meta-training distribution, but also how to effectively explore them, transferring knowledge about both the function structure and optimal search strategies across tasks. This contrasts with methods that only meta-learn the surrogate or only meta-learn the policy.

\paragraph{Comparison of transformer-based BO approaches.}
Although PFNs4BO, Meta-BO with transformers/NPs (including OptFormer and NAP), and our Decision Transformer-based strategy employ transformer architectures or related deep sequence models, they target different stages or aspects of the BO process. PFNs4BO operates at the surrogate modeling level by using a transformer to approximate the posterior over objective functions through amortized Bayesian inference---effectively replacing the Gaussian Process for posterior estimation and enabling rapid, retraining-free updates. Meta-BO approaches using NPs or transformers, such as OptFormer and NAP, focus on learning transferable surrogate models and acquisition strategies by training across multiple related tasks to create universal optimizers. In contrast, our Decision Transformer approach learns the acquisition policy directly from problem-specific simulated trajectories by modeling the sequence of decisions, mapping the history of observations to the next query suggestion, potentially leveraging offline RL paradigms. While PFNs4BO and Meta-NPs/universal optimizers remain closer to the Bayesian paradigm by explicitly modeling uncertainty (or distributions over functions) and leveraging inter-task meta-training, our method bypasses an explicit probabilistic surrogate for decision-making in favor of directly learning sequential query strategies from intra-task simulations. Collectively, these approaches underscore the broad potential of deep neural networks in enhancing BO---whether by constructing latent representations, refining surrogate models via meta-learning, amortizing inference, or directly learning sequential decision-making policies.

\subsection{Approximation-Aware and Multi-Fidelity Bayesian Optimization}
\label{subsec:approx-aware-bo}
Classical BO often assumes access to exact evaluations of the objective function, potentially corrupted by uniform noise. However, in many real-world scenarios, particularly those involving complex simulations or physical experiments, function evaluations may come from approximations or models with varying fidelity and cost. Approximation-aware BO methods explicitly account for these complexities.

A prominent area is Multi-Fidelity Bayesian Optimization (MFBO) \citep{kandasamy2016gaussian, kandasamy2017multi}. MFBO techniques leverage cheap, low-fidelity approximations (e.g., coarse simulations, simpler models) to guide the optimization towards promising regions where expensive, high-fidelity evaluations should be performed. Methods like MF-GP-UCB \citep{kandasamy2016gaussian} and BOCA \citep{kandasamy2017multi} model the relationship between different fidelity levels (e.g., using auto-regressive GP models) and design acquisition functions (like Multi-Fidelity Entropy Search; \citealp{takeno2020multi}) that optimally trade off information gain and cost across fidelities. Other approaches might deal with approximations by explicitly modeling the discrepancy between the approximate model and the true objective \citep{kennedy2000predicting} or by incorporating evaluation cost into the acquisition function more directly \citep{snoek2012practical}. These methods aim for end-to-end optimization under realistic evaluation constraints, acknowledging that the function oracle itself might be an approximation with controllable accuracy and cost, a setting distinct from assuming a perfect black box.

\subsection{Handling Unknown Hyperparameters in Bayesian Optimization}
\label{subsec:unknown-hyperparams-bo}
The performance of BO heavily relies on the surrogate model, typically a Gaussian process, whose behavior is governed by hyperparameters (e.g., kernel lengthscales, signal variance, noise variance). Standard BO practice often involves fixing these hyperparameters based on heuristics or estimating them via MLE or MAP estimation on the observed data \citep{rasmussen2006gaussian}. However, fixing hyperparameters can lead to suboptimal performance if the initial choice is poor, and MLE/MAP estimates can be unreliable with scarce data early in the optimization.

To address this, several approaches treat hyperparameters as nuisance parameters to be marginalized out, adopting a more fully Bayesian perspective. This involves defining priors over the hyperparameters and integrating the acquisition function over the posterior distribution of these hyperparameters \citep{snoek2012practical}. Techniques like MCMC sampling can be used to approximate the integral \citep{snoek2012practical}, leading to acquisition functions like the Integrated Expected Improvement \citep{brochu2010tutorial}. While computationally more intensive, these methods avoid committing to a single hyperparameter setting and tend to be more robust, especially in the early stages of BO when data is limited. Ensemble methods, where predictions or acquisition functions from models with different hyperparameter settings are averaged \citep{wistuba2018scalable}, offer a practical alternative. Recent developments, such as those utilizing PFNs \citep{muller2023pfns4bo}, can implicitly perform marginalization by training on a distribution over hyperparameter settings drawn from the prior. This relies on extensive pre-training and does not scale well with respect to dimensionality. Another line of work incorporates uncertainty in GP hyperparameters and devises acquisition functions accounting for both hyperparameter uncertainty and conventional max-value entropy \citep{hvarfner2023self}. In contrast to our work, they stick to a conventional GP surrogate model and information-theoretic acquisitions, which pose computational challenges.

\section{Theoretical Insights on the Simulation Components}
\label{sec:theory}
\newtheorem{theorem}{Theorem}
\newtheorem{lemma}{Lemma}
\newtheorem{corollary}{Corollary}
\newtheorem{proposition}{Proposition}
\theoremstyle{definition}
\newtheorem{assumption}{Assumption}
\theoremstyle{remark}
\newtheorem{remark}{Remark}
\theoremstyle{plain}

\paragraph{Scope.} The results in this appendix concern the components that generate DRO's training data: the region of interest of \eqnref{eq:roi_definition_m} and base acquisitions maximized inside it (Lemma~\ref{lem:roi} and Theorem~\ref{thm:roi_regret}), the expected-improvement quantity behind the Bayesian early-stopping rule (Theorem~\ref{thm:ei_convergence}, Proposition~\ref{prop:bes_sound} and their corollaries), and the choice of a lengthscale ensemble (Lemma~\ref{lem:lengthscale} and Corollary~\ref{cor:lengthscale}). They do not characterize the real queries produced by the Decision Transformer, and we claim no end-to-end regret or consistency guarantee for the learned policy; establishing such guarantees for the complete system is left open. Lemma and theorem numbers of \citet{Srinivas2010} refer to the extended version of that paper (arXiv:0912.3995v4), which contains the proofs.

\subsection{Regret of Base Acquisitions Restricted to the Region of Interest}
\label{subsec:roi_theory}

The rollouts of \secref{subsec:within_model_sampling} maximize a conventional acquisition function inside the ROI of \eqnref{eq:roi_definition_m} rather than over the whole domain. Corollary~1 of \citet{Zhang2023ROI} shows that such a region retains the optimizer with high probability; building on it, we show three things. On the event on which the GP confidence bounds hold, the ROI never excludes the optimizer, and every point it retains has regret controlled by posterior widths (Lemma~\ref{lem:roi}). GP-UCB keeps its regret bound when its search is restricted to the ROI, including when the ROI is held fixed for the duration of a rollout as in \eqnref{eq:rollout} (Theorem~\ref{thm:roi_regret}(a)). And the restriction makes no-regret a rule that is not no-regret on its own, namely uncertainty sampling (Theorem~\ref{thm:roi_regret}(b)).

\paragraph{Setting.} Let $\mathcal{X}$ be a finite set, let $f:\mathcal{X}\to\mathbb{R}$ be the objective with $\mathbf{x}^* \in \operatorname*{argmax}_{\mathbf{x}\in\mathcal{X}} f(\mathbf{x})$, and let the observations be $y_t = f(\mathbf{x}_t) + \epsilon_t$. The GP model has a known prior mean, which the arguments below do not use unless it is stated to be zero, and a kernel $k$ with $k(\mathbf{x},\mathbf{x}) \le 1$ for all $\mathbf{x}$ and noise variance $\sigma_n^2 > 0$; write $\mu_{t-1}, \sigma_{t-1}$ for its posterior mean and standard deviation after $t-1$ observations. For a confidence parameter $\beta_t > 0$, let $\mathrm{UCB}_t(\mathbf{x}) = \mu_{t-1}(\mathbf{x}) + \beta_t^{1/2}\sigma_{t-1}(\mathbf{x})$, $\mathrm{LCB}_t(\mathbf{x}) = \mu_{t-1}(\mathbf{x}) - \beta_t^{1/2}\sigma_{t-1}(\mathbf{x})$,
\begin{equation}
\hat{\mathcal{X}}_t = \bigl\{\mathbf{x}\in\mathcal{X} : \mathrm{UCB}_t(\mathbf{x}) \ge \max_{\mathbf{x}'\in\mathcal{X}} \mathrm{LCB}_t(\mathbf{x}')\bigr\},
\label{eq:roi_theory}
\end{equation}
which is \eqnref{eq:roi_definition_m} for a single surrogate, and let $\mathbf{x}_t^{\mathrm{U}} \in \operatorname*{argmax}_{\mathbf{x}\in\mathcal{X}} \mathrm{UCB}_t(\mathbf{x})$ denote an unrestricted GP-UCB maximizer. The instantaneous and cumulative regrets are $r_t = f(\mathbf{x}^*) - f(\mathbf{x}_t)$ and $R_T = \sum_{t=1}^T r_t$, and $\gamma_T$ is the maximum information gain of the GP model on $\mathcal{X}$ after $T$ observations \citep{Srinivas2010}. Finiteness of $\mathcal{X}$ is used only to ensure that the maxima below are attained; the arguments apply verbatim to a compact $\mathcal{X}$ and a continuous kernel whenever Assumption~\ref{ass:confidence} holds there.

\begin{assumption}[Uniform confidence bounds]
\label{ass:confidence}
For a given $\varrho \in (0,1)$, the non-decreasing sequence $(\beta_t)_{t\ge1}$ is such that, with probability at least $1-\varrho$, simultaneously for all $t \ge 1$ and all $\mathbf{x}\in\mathcal{X}$,
\begin{equation}
\bigl|\mu_{t-1}(\mathbf{x}) - f(\mathbf{x})\bigr| \le \beta_t^{1/2}\,\sigma_{t-1}(\mathbf{x}).
\label{eq:confidence}
\end{equation}
\end{assumption}

Assumption~\ref{ass:confidence} is the starting point of the GP-UCB analysis. \citet{Srinivas2010} establish it for a sample path of the GP prior on a finite domain, with $\beta_t = 2\log(|\mathcal{X}|\pi_t/\varrho)$ for any $\pi_t > 0$ with $\sum_{t\ge1}\pi_t^{-1} = 1$ (their Lemma~5.1), where a non-decreasing choice such as $\pi_t = \pi^2t^2/6$ makes $(\beta_t)$ non-decreasing as required here. They also establish it on an arbitrary domain for an objective with reproducing kernel Hilbert space (RKHS) norm $\|f\|_k^2 \le B$ observed under noise that is a martingale difference sequence uniformly bounded by $\sigma_n$, with $\beta_t = 2B + 300\gamma_t\log^3(t/\varrho)$ (their Theorem~6, stated with the smaller $2\|f\|_k^2$ in place of $2B$, and enlarging $\beta_t$ preserves the bound; it is the confidence bound behind their Theorem~3). For noise that is only conditionally $R$-sub-Gaussian, \citet[Theorem~2]{Chowdhury2017} give an RKHS bound whose $\beta_t$ has no factor $\log^3(t/\varrho)$. For a zero-mean GP model with kernel $k$, noise-variance parameter $\sigma_n^2 = 1 + 2/T_0$ (their $\lambda = 1 + \eta$ with $\eta = 2/T_0$, a parameter of the model and not the variance of $\epsilon_t$) and a horizon $T_0$ fixed in advance, with probability at least $1-\varrho$, \eqref{eq:confidence} holds simultaneously for all $\mathbf{x}$ and all $t \le T_0$ with $\beta_t^{1/2} = B^{1/2} + R\sqrt{2(\gamma_{t-1} + 1 + \log(1/\varrho))}$. Here $\gamma$ is computed at that noise variance and their $B$ bounds $\|f\|_k$ rather than its square. Their theorem is stated for all $t \ge 1$, but its proof, which ties the noise variance to the horizon, covers $t \le T_0$. This $(\beta_t)$ is non-decreasing on $t \le T_0$ because $\gamma_{t-1}$ is. Since the proof of Theorem~\ref{thm:roi_regret} for a horizon $T$ uses \eqref{eq:confidence} only at iterations $t \le T$, including through Lemma~\ref{lem:roi}(ii) for regions computed at $s \le t$, this bound supports Theorem~\ref{thm:roi_regret} for $T \le T_0$. Like Theorem~6 of \citet{Srinivas2010}, it concerns a fixed objective of bounded RKHS norm. In each case the probability $1-\varrho$ already accounts for the union over all iterations, which is why a single failure probability appears below rather than a per-iteration one. For sample paths on a continuous domain, \citet{Srinivas2010} obtain confidence bounds only at the queried points and on a sequence of discretizations, which they relate to the optimizer through its nearest grid point (Lemmas~5.5--5.7 in the proof of their Theorem~2); this is weaker than Assumption~\ref{ass:confidence}, and we do not pursue that case. We also use the following standard fact.

\begin{lemma}[\citealt{Srinivas2010}, Lemma~5.3 and the proof of Lemma~5.4]
\label{lem:infogain}
If $k(\mathbf{x},\mathbf{x}) \le 1$ for all $\mathbf{x}$, then for any sequence of points $\mathbf{x}_1, \dots, \mathbf{x}_T \in \mathcal{X}$ and any non-decreasing $(\beta_t)$, $\sum_{t=1}^T 4\beta_t\,\sigma_{t-1}^2(\mathbf{x}_t) \le C_1\,\beta_T\,\gamma_T$ with $C_1 = 8/\log(1+\sigma_n^{-2})$.
\end{lemma}

For a kernel with signal variance $s^2$, i.e., $k(\mathbf{x},\mathbf{x}) \le s^2$, Lemma~\ref{lem:infogain}, and hence every bound below that uses it, holds with $C_1 = 8s^2/\log(1+s^2\sigma_n^{-2})$, by applying Lemma~\ref{lem:infogain} to the model for $f/s$, whose kernel is $k/s^2$, whose noise variance is $\sigma_n^2/s^2$, and whose information gain is unchanged.

\begin{lemma}[Properties of the region of interest]
\label{lem:roi}
For every $t \ge 1$:
\begin{enumerate}[label=(\roman*), leftmargin=*]
    \item every maximizer of $\mathrm{UCB}_t$ over $\mathcal{X}$ lies in $\hat{\mathcal{X}}_t$; in particular $\hat{\mathcal{X}}_t \neq \emptyset$, and maximizing $\mathrm{UCB}_t$ over $\hat{\mathcal{X}}_t$ selects exactly the same points as maximizing it over $\mathcal{X}$.
\end{enumerate}
On the event of Assumption~\ref{ass:confidence}, moreover, for every $t \ge 1$:
\begin{enumerate}[label=(\roman*), leftmargin=*, resume]
    \item $\mathbf{x}^* \in \hat{\mathcal{X}}_t$; hence $\mathbf{x}^*$ also belongs to any region $\hat{\mathcal{X}}_s$ computed at another iteration $s$ and to any intersection of such regions;
    \item every $\mathbf{x} \notin \hat{\mathcal{X}}_t$ is certifiably suboptimal: $f(\mathbf{x}^*) - f(\mathbf{x}) \ge \max_{\mathbf{x}'\in\mathcal{X}}\mathrm{LCB}_t(\mathbf{x}') - \mathrm{UCB}_t(\mathbf{x}) > 0$;
    \item every $\mathbf{x} \in \hat{\mathcal{X}}_t$ satisfies $f(\mathbf{x}^*) - f(\mathbf{x}) \le 2\beta_t^{1/2}\bigl(\sigma_{t-1}(\mathbf{x}) + \sigma_{t-1}(\mathbf{x}_t^{\mathrm{U}})\bigr)$.
\end{enumerate}
\end{lemma}

\begin{proof}
(i) Let $\tilde{\mathbf{x}}$ maximize $\mathrm{UCB}_t$ over $\mathcal{X}$. Since $\sigma_{t-1} \ge 0$, $\mathrm{UCB}_t(\tilde{\mathbf{x}}) \ge \mathrm{UCB}_t(\mathbf{x}') \ge \mathrm{LCB}_t(\mathbf{x}')$ for every $\mathbf{x}'\in\mathcal{X}$, so $\mathrm{UCB}_t(\tilde{\mathbf{x}}) \ge \max_{\mathbf{x}'}\mathrm{LCB}_t(\mathbf{x}')$ and $\tilde{\mathbf{x}} \in \hat{\mathcal{X}}_t$. Hence $\max_{\hat{\mathcal{X}}_t}\mathrm{UCB}_t = \max_{\mathcal{X}}\mathrm{UCB}_t$, and the maximizers over $\hat{\mathcal{X}}_t$ are the maximizers over $\mathcal{X}$.

For (ii)--(iv), work on the event of Assumption~\ref{ass:confidence}, on which $\mathrm{LCB}_t(\mathbf{x}) \le f(\mathbf{x}) \le \mathrm{UCB}_t(\mathbf{x})$ for all $t$ and $\mathbf{x}$.

(ii) For every $\mathbf{x}' \in \mathcal{X}$, $\mathrm{UCB}_t(\mathbf{x}^*) \ge f(\mathbf{x}^*) \ge f(\mathbf{x}') \ge \mathrm{LCB}_t(\mathbf{x}')$, so $\mathrm{UCB}_t(\mathbf{x}^*) \ge \max_{\mathbf{x}'}\mathrm{LCB}_t(\mathbf{x}')$, which is the membership condition in \eqref{eq:roi_theory}. As the event covers all iterations at once, the same holds for every $\hat{\mathcal{X}}_s$, hence for their intersections.

(iii) Let $\hat{\mathbf{x}}$ maximize $\mathrm{LCB}_t$. Then $f(\mathbf{x}^*) - f(\mathbf{x}) \ge f(\hat{\mathbf{x}}) - f(\mathbf{x}) \ge \mathrm{LCB}_t(\hat{\mathbf{x}}) - \mathrm{UCB}_t(\mathbf{x})$, which is positive when $\mathbf{x} \notin \hat{\mathcal{X}}_t$.

(iv) Write $b = \beta_t^{1/2}$. For $\mathbf{x} \in \hat{\mathcal{X}}_t$, $\mathrm{UCB}_t(\mathbf{x}) \ge \max_{\mathbf{x}'}\mathrm{LCB}_t(\mathbf{x}') \ge \mathrm{LCB}_t(\mathbf{x}_t^{\mathrm{U}})$. Therefore
$$ f(\mathbf{x}) \ge \mathrm{LCB}_t(\mathbf{x}) = \mathrm{UCB}_t(\mathbf{x}) - 2b\,\sigma_{t-1}(\mathbf{x}) \ge \mathrm{UCB}_t(\mathbf{x}_t^{\mathrm{U}}) - 2b\,\sigma_{t-1}(\mathbf{x}_t^{\mathrm{U}}) - 2b\,\sigma_{t-1}(\mathbf{x}), $$
and $\mathrm{UCB}_t(\mathbf{x}_t^{\mathrm{U}}) \ge \mathrm{UCB}_t(\mathbf{x}^*) \ge f(\mathbf{x}^*)$.
\end{proof}

\begin{theorem}[Regret of rules restricted to the region of interest]
\label{thm:roi_regret}
Let Assumption~\ref{ass:confidence} hold. Then, with probability at least $1-\varrho$, simultaneously for all $T \ge 1$:
\begin{enumerate}[label=(\alph*), leftmargin=*]
    \item \emph{GP-UCB on any search set that keeps the optimizer.} Let $(\mathcal{S}_t)_{t\ge1}$ be search sets such that $\mathbf{x}^* \in \mathcal{S}_t$ for all $t$ whenever the event of Assumption~\ref{ass:confidence} holds; by Lemma~\ref{lem:roi}(ii) this is the case for the current region $\mathcal{S}_t = \hat{\mathcal{X}}_t$, for a region $\hat{\mathcal{X}}_s$ computed at an earlier iteration $s \le t$ and held fixed, and for the running intersection $\bigcap_{s\le t}\hat{\mathcal{X}}_s$. If $\mathbf{x}_t \in \operatorname*{argmax}_{\mathbf{x}\in\mathcal{S}_t}\mathrm{UCB}_t(\mathbf{x})$ (chosen arbitrarily if $\mathcal{S}_t = \emptyset$), then $r_t \le 2\beta_t^{1/2}\sigma_{t-1}(\mathbf{x}_t)$ for every $t \le T$, and $R_T \le \sqrt{C_1\,T\,\beta_T\,\gamma_T}$, which is the bound of Theorem~1 of \citet{Srinivas2010} for unrestricted GP-UCB.
    \item \emph{Any rule inside the region that keeps comparable uncertainty.} If $\mathbf{x}_t \in \hat{\mathcal{X}}_t$ and $\sigma_{t-1}(\mathbf{x}_t) \ge c\,\sigma_{t-1}(\mathbf{x}_t^{\mathrm{U}})$ for every $t \le T$ and a constant $c \in (0, 1]$, then $R_T \le (1 + 1/c)\sqrt{C_1\,T\,\beta_T\,\gamma_T}$. In particular, uncertainty sampling inside the region, $\mathbf{x}_t \in \operatorname*{argmax}_{\mathbf{x}\in\hat{\mathcal{X}}_t}\sigma_{t-1}(\mathbf{x})$, satisfies $R_T \le 2\sqrt{C_1\,T\,\beta_T\,\gamma_T}$.
\end{enumerate}
Consequently, if $f$ takes values in an interval of length $\Delta$, then $\mathbb{E}[R_T] \le \bar R_T + \varrho\,\Delta\,T$, where $\bar R_T$ is the bound of (a) or (b) for the rule in use.
\end{theorem}

\begin{proof}
Work on the event of Assumption~\ref{ass:confidence}, which has probability at least $1-\varrho$; everything below holds deterministically on this event, for all $t$ at once.

(a) Since $\mathbf{x}^* \in \mathcal{S}_t$ and $\mathbf{x}_t$ maximizes $\mathrm{UCB}_t$ over $\mathcal{S}_t$, $\mathrm{UCB}_t(\mathbf{x}_t) \ge \mathrm{UCB}_t(\mathbf{x}^*) \ge f(\mathbf{x}^*)$. Together with $f(\mathbf{x}_t) \ge \mathrm{LCB}_t(\mathbf{x}_t)$,
$$ 0 \le r_t = f(\mathbf{x}^*) - f(\mathbf{x}_t) \;\le\; \mathrm{UCB}_t(\mathbf{x}_t) - \mathrm{LCB}_t(\mathbf{x}_t) \;=\; 2\beta_t^{1/2}\,\sigma_{t-1}(\mathbf{x}_t). $$
This is the only step of the GP-UCB analysis in which the search set enters (Lemma~5.2 of \citealt{Srinivas2010}), and it uses nothing about $\mathcal{S}_t$ beyond $\mathbf{x}^* \in \mathcal{S}_t$. By the Cauchy--Schwarz inequality and Lemma~\ref{lem:infogain},
$$ R_T = \sum_{t=1}^T r_t \;\le\; \sqrt{T \sum_{t=1}^T r_t^2} \;\le\; \sqrt{T \sum_{t=1}^T 4\beta_t\,\sigma_{t-1}^2(\mathbf{x}_t)} \;\le\; \sqrt{C_1\,T\,\beta_T\,\gamma_T}. $$

(b) By Lemma~\ref{lem:roi}(iv) and the hypothesis,
$$ 0 \le r_t \le 2\beta_t^{1/2}\bigl(\sigma_{t-1}(\mathbf{x}_t) + \sigma_{t-1}(\mathbf{x}_t^{\mathrm{U}})\bigr) \le (1 + 1/c)\,2\beta_t^{1/2}\sigma_{t-1}(\mathbf{x}_t), $$
and the conclusion follows as in (a). For uncertainty sampling, $\mathbf{x}_t^{\mathrm{U}} \in \hat{\mathcal{X}}_t$ by Lemma~\ref{lem:roi}(i), so $\sigma_{t-1}(\mathbf{x}_t) = \max_{\hat{\mathcal{X}}_t}\sigma_{t-1} \ge \sigma_{t-1}(\mathbf{x}_t^{\mathrm{U}})$ and $c = 1$.

For the expectation, $R_T \le \Delta T$ holds always, so $\mathbb{E}[R_T] \le \bar R_T\cdot 1 + \Delta T\cdot\varrho$. In the RKHS case, $|f(\mathbf{x})| \le \|f\|_k\,k(\mathbf{x},\mathbf{x})^{1/2} \le \sqrt{B}$, so $\Delta \le 2\sqrt{B}$; GP sample paths have no deterministic bound (their range is random), and there the Cauchy--Schwarz inequality $\mathbb{E}[R_T\,\mathbf{1}_{E^c}] \le (\varrho\,\mathbb{E}[R_T^2])^{1/2}$, with $E$ the event of Assumption~\ref{ass:confidence}, replaces the term $\varrho\,\Delta\,T$.

For a fixed $\varrho$, the term $\varrho\,\Delta\,T$ grows linearly in $T$. For a horizon $T_0 \ge 2$ fixed in advance, running the rule with $\varrho = 1/T_0$ gives $\mathbb{E}[R_T] \le \bar R_T + \Delta$ for all $T \le T_0$, with $\bar R_T$ evaluated at $\varrho = 1/T_0$; this is a fixed-horizon statement, because $\beta_t$, and with it the rule, then depends on $T_0$. On a finite domain, this choice turns the $\beta_t$ of Lemma~5.1 of \citet{Srinivas2010} into $2\log(|\mathcal{X}|\pi_t) + 2\log T_0$, which for $\pi_t = \pi^2t^2/6$ gives $\beta_{T_0} = 2\log(|\mathcal{X}|\pi^2T_0^3/6)$. There $f$ is a Gaussian sample path with no deterministic range, but $R_{T_0} \le T_0 D$ with $D = \max_{\mathcal{X}} f - \min_{\mathcal{X}} f$, and $\mathbb{E}[D^2] \le 4\,\mathbb{E}[\max_{\mathbf{x}\in\mathcal{X}} f(\mathbf{x})^2] \le 16\log|\mathcal{X}| + 8\log 2$ for a zero-mean prior with $k(\mathbf{x},\mathbf{x}) \le 1$ (the first inequality because $D \le 2\max_{\mathbf{x}}|f(\mathbf{x})|$, the second by Jensen's inequality, since $\mathbb{E}\,e^{\max_{\mathbf{x}} f(\mathbf{x})^2/4} \le \sum_{\mathbf{x}}\mathbb{E}\,e^{f(\mathbf{x})^2/4} = \sum_{\mathbf{x}}(1 - k(\mathbf{x},\mathbf{x})/2)^{-1/2} \le \sqrt{2}\,|\mathcal{X}|$); the Cauchy--Schwarz term is then at most $\bigl(T_0(16\log|\mathcal{X}| + 8\log 2)\bigr)^{1/2}$, and $\mathbb{E}[R_{T_0}] = O\bigl((T_0\,\gamma_{T_0}\log(|\mathcal{X}|\,T_0))^{1/2}\bigr)$. In the RKHS case of Theorem~6 of \citet{Srinivas2010}, with $\|f\|_k^2 \le B$, $\beta_{T_0} = 2B + 2400\,\gamma_{T_0}\log^3 T_0$, at most eight times its value for any fixed $\varrho \in (0,1)$, and $\mathbb{E}[R_{T_0}] = O\bigl((B\,T_0\,\gamma_{T_0})^{1/2} + \gamma_{T_0}(T_0\log^3 T_0)^{1/2}\bigr)$, which is sublinear when $\gamma_T\log^{3/2}T = o(\sqrt{T})$, the same condition under which the high-probability bound $\bar R_T$ is sublinear for a fixed $\varrho$.
\end{proof}

\begin{remark}[What the restriction does and does not buy]
\label{rem:roi}
By Lemma~\ref{lem:roi}(i), GP-UCB restricted to the current region coincides with GP-UCB, so for that rule Theorem~\ref{thm:roi_regret}(a) is the guarantee of \citet{Srinivas2010}. The content of part (a) is that the restriction also costs nothing in the forms in which the restricted and unrestricted maximizers can differ: a region held fixed while the posterior evolves, as in the rollouts of \eqnref{eq:rollout}, or a running intersection. Lemma~\ref{lem:roi}(ii) also holds, and hence so does part (a), when the region is built with any width larger than $\beta_t^{1/2}$, since enlarging the width only enlarges the set. Part (b) is where the restriction does work: uncertainty sampling ignores the posterior mean and has no regret guarantee on its own, but inside the region the mean enters through the set, and the rule inherits the GP-UCB rate up to a factor of two (this rule is a single-GP form of the BALLET-RCI acquisition, Eq.~(9) of \citealt{Zhang2023ROI}, which runs uncertainty sampling inside the region). For EI, PI and MES, the other members of the rotation in \secref{subsec:within_model_sampling}, the restriction is active but neither part gives a rate; Lemma~\ref{lem:roi}(iv) still bounds the regret of every point the region retains, and hence of whichever point they select, by posterior widths at the time the region is computed, a bound that does not hold for the unrestricted rules. In a rollout, where the region is held fixed, this bound is computed from the posterior at the start of the rollout and does not shrink as the rollout proceeds. None of these results claims a faster rate than GP-UCB: the information gain is that of $\mathcal{X}$ (the queried points lie in $\bigcup_t \hat{\mathcal{X}}_t \subseteq \mathcal{X}$, so any information-gain bound for $\mathcal{X}$ applies to them), and the benefit of the ROI in DRO is that the finite simulation budget is spent where the optimizer can still be (\secref{sec:ablations}).
\end{remark}

\begin{remark}[Application to the rollouts, and gaps to the implementation]
\label{rem:roi_rollout}
Consider a rollout of member $m$ started at real iteration $t$. Suppose that its observations are drawn sequentially from the posterior predictive of $\mathrm{GP}_m$, noise included, that $\mathrm{GP}_m$ is conditioned on each of them with the same noise variance and with its hyperparameters held fixed during the rollout, as in \secref{subsec:within_model_sampling}, and that the randomness of the acquisition choice is independent of the simulated observations. Then, conditionally on $\mathcal{D}_{t-1}$, the rollout has the same law as drawing a simulated objective $\tilde f$ from the posterior of $\mathrm{GP}_m$ given $\mathcal{D}_{t-1}$ and observing it with $\mathcal{N}(0,\sigma_n^2)$ noise. The model is therefore well specified in the simulated world, and on a finite domain Assumption~\ref{ass:confidence} holds for the rollout by Lemma~5.1 of \citet{Srinivas2010}, applied with the posterior of $\mathrm{GP}_m$ in place of the prior. The region $\hat{\mathcal{X}}_{m,t}$ is computed once from that posterior and held fixed, which is the fixed-region case of Theorem~\ref{thm:roi_regret}(a) provided its width is at least the confidence width at the first simulated step. A UCB-driven rollout of $L$ steps inside $\hat{\mathcal{X}}_{m,t}$ thus has simulated cumulative regret at most $\sqrt{C_1 L\beta_L\gamma_L}$ with probability at least $1-\varrho$, where $C_1$ is computed with the signal variance of $\mathrm{GP}_m$ as above and $\gamma_L$ may be taken to be the maximum information gain of the prior kernel of $\mathrm{GP}_m$: the information gain of a set $A$ of $L$ points is $\frac12\log\det(I + \sigma_n^{-2}K_A)$, which is non-decreasing in $K_A$ in the Loewner order, and the posterior covariance matrix of $f$ on $A$ given $\mathcal{D}_{t-1}$ is dominated in that order by the prior one. Since the per-step bound of (a) holds at every step at which UCB is used, and Lemma~\ref{lem:infogain} bounds the sum over any subset of steps, the regret accumulated on the $L_{\mathrm{U}}$ UCB steps of a rollout that rotates acquisitions is at most $\sqrt{C_1 L_{\mathrm{U}}\beta_L\gamma_L}$; for its EI, PI and MES steps the result is a design justification, not a guarantee. These statements concern the simulated world of each member and say nothing about the real objective, on which a member may be misspecified. Finally, the implementation uses fixed widths in place of a growing $\beta_t^{1/2}$ (\appref{sec:implementation_details}); in the simulated world on a finite domain, a fixed width $\kappa$ makes Assumption~\ref{ass:confidence} hold over $L$ steps with failure probability at most $L\,|\mathcal{X}|\,e^{-\kappa^2/2}$, by the union bound in the proof of Lemma~5.1 of \citet{Srinivas2010} without the schedule $\pi_t$.
\end{remark}

\textit{Insight for Theorem~\ref{thm:roi_regret}:} Restricting a base acquisition to the ROI of \eqnref{eq:roi_definition_m} never removes the optimizer on the event on which the confidence bounds hold, keeps the GP-UCB guarantee whether the region is recomputed or held fixed over a rollout, and turns even a rule that ignores the posterior mean into a no-regret one. This justifies using ROI filtering in DRO to focus simulated rollouts without undermining the soundness of the underlying acquisition strategy.

\subsection{Expected Improvement and the Bayesian Early-Stopping Rule}
\label{subsec:ei_theory}

The Bayesian early-stopping rule of \secref{subsec:early_stop_method} terminates a rollout once the maximal expected improvement inside the ROI falls below a threshold $\delta$, the stopping condition for EI analyzed by \citet{nguyen2017regret}. The results below make precise in what sense this is a sound termination signal, in both directions. Completeness (Theorem~\ref{thm:ei_convergence}): the expected improvement at any point is controlled by three quantities, namely the improvement still available at that point (at most the current simple regret), the surrogate's mean error there, and the surrogate's posterior standard deviation there, so on any region where the surrogate is consistent, vanishing simple regret forces the maximal expected improvement to vanish uniformly and the rule fires. Soundness (Proposition~\ref{prop:bes_sound}): when the rule fires, no point of the region can improve on the incumbent by more than an explicit multiple $c_b\,\delta$ of the threshold on the confidence event, and the multiple $c_b$ cannot be improved. We first fix the setting and the assumptions.

\paragraph{Setting.} Let $\mathcal{X}$ be a compact subset of $\mathbb{R}^d$ and let $f:\mathcal{X}\to\mathbb{R}$ be continuous, so that $f^* = \max_{\mathbf{x}\in\mathcal{X}} f(\mathbf{x})$ is finite and attained. After $t$ observations, a Gaussian-process surrogate provides a posterior mean $\mu_t:\mathcal{X}\to\mathbb{R}$ and a posterior standard deviation $\sigma_t:\mathcal{X}\to[0,\infty)$, so that the surrogate's belief about the value at $\mathbf{x}$ is $\mathcal{N}(\mu_t(\mathbf{x}), \sigma_t^2(\mathbf{x}))$. As in \appref{subsec:roi_theory}, the subscript counts observations, but the iteration index is shifted by one: there, iteration $t$ uses $(\mu_{t-1}, \sigma_{t-1})$ and $\beta_t$, so the posterior $(\mu_t, \sigma_t)$ here is the one that \appref{subsec:roi_theory} uses at iteration $t+1$, with confidence parameter $\beta_{t+1}$. Let $\xi_t \in \mathbb{R}$ be the incumbent value against which improvement is measured. In the formulation of \secref{sec:problem-formulation} this is the true value of the best evaluated point, $\xi_t = f(\mathbf{x}_t^+) = \max_{i \le t} f(\mathbf{x}_i)$, whereas \citet{nguyen2017regret} use the best noisy observation $\max_{i\le t} y_i$; the results below hold verbatim for either choice and for any other incumbent rule, such as the largest posterior mean over the evaluated points. The expected improvement is
\begin{equation}
\mathrm{EI}_t(\mathbf{x}) = \mathbb{E}_{Y \sim \mathcal{N}(\mu_t(\mathbf{x}),\, \sigma_t^2(\mathbf{x}))}\bigl[\max\bigl(0,\, Y - \xi_t\bigr)\bigr],
\label{eq:ei_def}
\end{equation}
with the convention $\mathrm{EI}_t(\mathbf{x}) = \max(0, \mu_t(\mathbf{x}) - \xi_t)$ when $\sigma_t(\mathbf{x}) = 0$. We write $\Phi$ and $\phi$ for the standard normal distribution and density functions, $\psi(z) = z\,\Phi(z) + \phi(z)$, and, for a set $S \subseteq \mathcal{X}$,
\begin{equation}
r^+_t = f^* - \xi_t, \quad
r^+_t(S) = \sup_{\mathbf{x}\in S} f(\mathbf{x}) - \xi_t, \quad
\varepsilon_t(S) = \sup_{\mathbf{x}\in S} \bigl|\mu_t(\mathbf{x}) - f(\mathbf{x})\bigr|, \quad
s_t(S) = \sup_{\mathbf{x}\in S} \sigma_t(\mathbf{x}),
\label{eq:ei_quantities}
\end{equation}
for the simple regret of the incumbent (marked $+$, like the best point $\mathbf{x}_t^+$, to distinguish it from the regret $r_t$ of a query in \appref{subsec:roi_theory}), its regret relative to the best value on $S$ (so that $r^+_t(S) \le r^+_t$), the mean error of the surrogate on $S$, and the posterior width of the surrogate on $S$. The regret $r^+_t$ can be negative for an incumbent based on noisy observations, which is why it enters below through $\max(0, r^+_t)$.

\begin{assumption}[Surrogate consistency on $S$]
\label{ass:consistency}
For a set $S \subseteq \mathcal{X}$, $\varepsilon_t(S) \to 0$ and $s_t(S) \to 0$ as $t \to \infty$.
\end{assumption}

\begin{assumption}[Vanishing simple regret]
\label{ass:regret}
$\max(0, r^+_t) \to 0$ as $t \to \infty$.
\end{assumption}

Assumption~\ref{ass:regret} is the hypothesis of the statement made in the main text. Assumption~\ref{ass:consistency} is the additional hypothesis needed to pass from vanishing regret to vanishing expected improvement, and it cannot be dropped (Remark~\ref{rem:necessity}). It concerns the surrogate on $S$ only; in DRO the relevant set is the ROI, inside which both the rollouts and the stopping test live. Both assumptions may hold deterministically or only with high probability; the theorem below is a deterministic inequality, so probabilistic versions of its corollaries follow at once.

\begin{lemma}[Closed form and bounds for the expected improvement]
\label{lem:tau}
With $\psi(z) = z\,\Phi(z) + \phi(z)$ and $Z \sim \mathcal{N}(0,1)$:
\begin{enumerate}[label=(\roman*), leftmargin=*]
    \item $\psi$ is strictly increasing and convex on $\mathbb{R}$, with $\psi'(z) = \Phi(z)$, $\lim_{z \to -\infty}\psi(z) = 0$, and $\psi(z) > 0$ for every $z \in \mathbb{R}$;
    \item $\max(z, 0) \le \psi(z) \le \max(z, 0) + \phi(z) \le \max(z, 0) + \phi(0)$ for every $z \in \mathbb{R}$, where $\phi(0) = 1/\sqrt{2\pi}$;
    \item $\psi(z) = \mathbb{E}[\max(0, Z + z)]$, and if $\sigma_t(\mathbf{x}) > 0$, then $\mathrm{EI}_t(\mathbf{x}) = \sigma_t(\mathbf{x})\,\psi\bigl(z_t(\mathbf{x})\bigr)$ with $z_t(\mathbf{x}) = \bigl(\mu_t(\mathbf{x}) - \xi_t\bigr)/\sigma_t(\mathbf{x})$ \citep{Jones1998}.
\end{enumerate}
\end{lemma}

\begin{proof}
(iii) $\mathbb{E}[\max(0, Z + z)] = \int_{-z}^{\infty} (u + z)\,\phi(u)\,du = \phi(z) + z\,\Phi(z) = \psi(z)$, using $\int_{-z}^{\infty} u\,\phi(u)\,du = \phi(-z) = \phi(z)$ and $\int_{-z}^{\infty} \phi(u)\,du = 1 - \Phi(-z) = \Phi(z)$. If $\sigma_t(\mathbf{x}) > 0$, write $Y = \mu_t(\mathbf{x}) + \sigma_t(\mathbf{x})\, Z$ in \eqref{eq:ei_def}; then $\max(0, Y - \xi_t) = \sigma_t(\mathbf{x})\max(0, Z + z_t(\mathbf{x}))$, and taking expectations gives the closed form.

(i) Since $\phi'(z) = -z\,\phi(z)$, we have $\psi'(z) = \Phi(z) + z\,\phi(z) - z\,\phi(z) = \Phi(z) \in (0, 1)$, so $\psi$ is strictly increasing, and $\psi''(z) = \phi(z) > 0$, so $\psi$ is convex. For $z < 0$, the Mills-ratio bound $\Phi(z) \le \phi(z)/|z|$ gives $|z\,\Phi(z)| \le \phi(z)$, hence $0 \le \psi(z) \le 2\phi(z) \to 0$ as $z \to -\infty$. A strictly increasing function with limit $0$ at $-\infty$ is positive everywhere.

(ii) The lower bound is Jensen's inequality for the convex function $u \mapsto \max(0, u)$ applied to (iii): $\psi(z) = \mathbb{E}[\max(0, Z + z)] \ge \max(0, \mathbb{E}[Z + z]) = \max(z, 0)$. For the upper bound, if $z \le 0$, then $z\,\Phi(z) \le 0$, hence $\psi(z) \le \phi(z) = \max(z, 0) + \phi(z)$; if $z > 0$, then $\psi(z) - z = z\,(\Phi(z) - 1) + \phi(z) = \phi(z) - z\,\Phi(-z) \le \phi(z)$. In both cases, $\phi(z) \le \phi(0)$ because $\phi$ is maximized at $0$.
\end{proof}

\begin{theorem}[Expected improvement is controlled by regret and surrogate error]
\label{thm:ei_convergence}
In the setting above, for every iteration $t$ and every $\mathbf{x} \in \mathcal{X}$,
\begin{equation}
\mathrm{EI}_t(\mathbf{x}) \;\le\; \bigl|\mu_t(\mathbf{x}) - f(\mathbf{x})\bigr| \;+\; \max\bigl(0, f(\mathbf{x}) - \xi_t\bigr) \;+\; \frac{\sigma_t(\mathbf{x})}{\sqrt{2\pi}},
\label{eq:ei_bound}
\end{equation}
where the middle term is at most $\max(0, r^+_t)$. Consequently, for every $S \subseteq \mathcal{X}$,
\begin{equation}
\sup_{\mathbf{x} \in S} \mathrm{EI}_t(\mathbf{x}) \;\le\; \varepsilon_t(S) \;+\; \max\bigl(0, r^+_t(S)\bigr) \;+\; \frac{s_t(S)}{\sqrt{2\pi}}
\;\le\; \varepsilon_t(S) \;+\; \max(0, r^+_t) \;+\; \frac{s_t(S)}{\sqrt{2\pi}}.
\label{eq:ei_sup_bound}
\end{equation}
\end{theorem}

\begin{proof}
Fix $t$ and $\mathbf{x}$. If $\sigma_t(\mathbf{x}) > 0$, Lemma~\ref{lem:tau}(iii) followed by Lemma~\ref{lem:tau}(ii) gives
\begin{equation*}
\mathrm{EI}_t(\mathbf{x}) = \sigma_t(\mathbf{x})\,\psi\bigl(z_t(\mathbf{x})\bigr)
\le \sigma_t(\mathbf{x}) \max\bigl(z_t(\mathbf{x}), 0\bigr) + \sigma_t(\mathbf{x})\,\phi(0)
= \max\bigl(0, \mu_t(\mathbf{x}) - \xi_t\bigr) + \frac{\sigma_t(\mathbf{x})}{\sqrt{2\pi}} .
\end{equation*}
If $\sigma_t(\mathbf{x}) = 0$, the same inequality holds by the convention in \eqref{eq:ei_def}, with the last term equal to zero. It remains to bound the first term. Write $\mu_t(\mathbf{x}) - \xi_t = \bigl(\mu_t(\mathbf{x}) - f(\mathbf{x})\bigr) + \bigl(f(\mathbf{x}) - \xi_t\bigr)$. Since $u \mapsto \max(0, u)$ is subadditive, $\max(0, a + b) \le \max(0, a) + \max(0, b)$ for all real $a, b$, and therefore
\begin{equation*}
\max\bigl(0, \mu_t(\mathbf{x}) - \xi_t\bigr) \le \max\bigl(0, \mu_t(\mathbf{x}) - f(\mathbf{x})\bigr) + \max\bigl(0, f(\mathbf{x}) - \xi_t\bigr) \le \bigl|\mu_t(\mathbf{x}) - f(\mathbf{x})\bigr| + \max\bigl(0, f(\mathbf{x}) - \xi_t\bigr),
\end{equation*}
which proves \eqref{eq:ei_bound}. Because $u \mapsto \max(0, u)$ is non-decreasing and $f(\mathbf{x}) \le \sup_{S} f \le f^*$ for $\mathbf{x} \in S$, the middle term of \eqref{eq:ei_bound} is at most $\max(0, r^+_t(S)) \le \max(0, r^+_t)$. Taking the supremum over $\mathbf{x} \in S$ on both sides of \eqref{eq:ei_bound} and bounding each term on the right by its own supremum over $S$ gives \eqref{eq:ei_sup_bound}.
\end{proof}

\begin{corollary}[Uniform convergence of the expected improvement]
\label{cor:ei_uniform}
Let $S \subseteq \mathcal{X}$. Under Assumptions~\ref{ass:consistency} and \ref{ass:regret}, $\sup_{\mathbf{x} \in S} \mathrm{EI}_t(\mathbf{x}) \to 0$ as $t \to \infty$, that is, $\mathrm{EI}_t \to 0$ uniformly on $S$. If, instead, at iteration $t$ the event $\{\varepsilon_t(S) + \max(0, r^+_t(S)) + s_t(S)/\sqrt{2\pi} \le \eta_t\}$ has probability at least $1 - \zeta_t$, then $\sup_{\mathbf{x} \in S} \mathrm{EI}_t(\mathbf{x}) \le \eta_t$ with probability at least $1 - \zeta_t$. Applied within a rollout, with the simulation step in place of $t$ and the region $S = \hat{\mathcal{X}}_{m,t}$ of real iteration $t$, which is fixed for the duration of the rollout, this is the asymptotic form of the statement made in \secref{subsec:early_stop_method}, under the two stated assumptions; since rollouts have finite length, Corollary~\ref{cor:bes} below is the form that applies to them.
\end{corollary}

\begin{proof}
Both claims are immediate from \eqref{eq:ei_sup_bound}: its right-hand side tends to zero under the assumptions, and its middle expression is at most $\eta_t$ on the stated event.
\end{proof}

\begin{corollary}[Bayesian early stopping inside the ROI]
\label{cor:bes}
Consider a rollout generated from $\mathrm{GP}_m$ inside its region $S = \hat{\mathcal{X}}_{m,t}$. Let $\tilde f$ be the objective of the simulated world (the function whose values the rollout observes) and $\tilde f^*_S = \max_{\mathbf{x} \in S} \tilde f(\mathbf{x})$. After $\tau$ simulated steps, let $\mu^{(\tau)}, \sigma^{(\tau)}$ be the posterior of $\mathrm{GP}_m$ conditioned on $\mathcal{D}_{t-1}$ and the $\tau$ simulated observations, $\xi^{(\tau)} = f_{m,\text{best}}^{(\tau)}$ the simulated incumbent, and $\varepsilon^{(\tau)}(S), s^{(\tau)}(S)$ the quantities of \eqref{eq:ei_quantities} computed with $\tilde f$ and this posterior. If
\begin{equation*}
\varepsilon^{(\tau)}(S) + \max\bigl(0, \tilde f^*_S - \xi^{(\tau)}\bigr) + \frac{s^{(\tau)}(S)}{\sqrt{2\pi}} < \delta,
\end{equation*}
then the stopping condition \eqnref{eq:bes_ei} holds at step $\tau$, i.e., $\max_{\mathbf{x} \in S} \mathrm{EI}^{(\tau)}(\mathbf{x}) < \delta$. Thus a rollout is stopped no later than the first step at which, by its own model, the improvement left inside the ROI, the model's error on the ROI, and the model's uncertainty on the ROI are jointly small.
\end{corollary}

\begin{proof}
Apply the first inequality of \eqref{eq:ei_sup_bound} with the objective $\tilde f$, the posterior $(\mu^{(\tau)}, \sigma^{(\tau)})$, the incumbent $\xi^{(\tau)}$ and the set $S$, for which $r^+(S) = \tilde f^*_S - \xi^{(\tau)}$; the maximum over the compact set $S$ is at most the supremum.
\end{proof}

Corollary~\ref{cor:bes} and the following Proposition~\ref{prop:bes_sound} concern the max-EI rule \eqnref{eq:bes_ei}. The released implementation uses a cheaper proxy that tests the realized improvement of the sampled point (\appref{sec:implementation_details}). Given the data before that step, the proxy's expectation is the expected improvement of a noisy observation at that single point, not the maximal expected improvement over the region, and the proxy is zero whenever the sampled point fails to improve on the incumbent. It can therefore stop a rollout earlier than the max-EI rule, and it is not covered by these results.

\begin{proposition}[Soundness of the max-EI stopping rule]
\label{prop:bes_sound}
Fix $S \subseteq \mathcal{X}$ and $b > 0$, and suppose that at iteration $t$ the confidence bound $|\mu_t(\mathbf{x}) - f(\mathbf{x})| \le b\,\sigma_t(\mathbf{x})$ holds for all $\mathbf{x} \in S$. (Under Assumption~\ref{ass:confidence}, in whose notation the posterior after $t$ observations is $(\mu_t, \sigma_t)$, this holds for all $t$ simultaneously with $b = b_t = \beta_{t+1}^{1/2}$, with probability at least $1-\varrho$; the pointwise bound \eqref{eq:bes_pointwise} below needs the confidence bound only at the point $\mathbf{x}$ itself.) Let $z_b$ be the unique solution of $\phi(z) = b\,\Phi(z)$ on $(-b, \infty)$, and let $c_b = 1/\Phi(z_b)$. Then, for every $\mathbf{x} \in S$,
\begin{equation}
f(\mathbf{x}) - \xi_t \;\le\; c_b\,\mathrm{EI}_t(\mathbf{x}).
\label{eq:bes_pointwise}
\end{equation}
Consequently, if $\sup_{\mathbf{x}\in S}\mathrm{EI}_t(\mathbf{x}) < \delta$, then
\begin{equation}
\sup_{\mathbf{x}\in S} \bigl(f(\mathbf{x}) - \xi_t\bigr) \;\le\; c_b \sup_{\mathbf{x}\in S}\mathrm{EI}_t(\mathbf{x}) \;<\; c_b\,\delta .
\label{eq:bes_sound}
\end{equation}
The constant $c_b$ is the best possible: for every $b > 0$ there are values of $\mu_t(\mathbf{x})$, $\sigma_t(\mathbf{x}) > 0$, $\xi_t$ and $f(\mathbf{x})$ that satisfy the confidence bound and attain equality in \eqref{eq:bes_pointwise}. Moreover, $1 < c_b \le 1 + b/\psi(-b)$, and $c_b = b/\phi(z_b) < \sqrt{2\pi}\, b\, e^{b^2/2}$ whenever $b \ge 2\phi(0)$.
\end{proposition}

\begin{proof}
\emph{The constant.} For $z > -b$, let $g(z) = (z + b)/\psi(z)$, which is well defined because $\psi > 0$. Using $\psi' = \Phi$ and $\psi(z) = z\,\Phi(z) + \phi(z)$,
$$ g'(z) = \frac{\psi(z) - (z + b)\,\Phi(z)}{\psi(z)^2} = \frac{h(z)}{\psi(z)^2}, \qquad h(z) = \phi(z) - b\,\Phi(z). $$
The function $h$ is continuous with $h'(z) = -(z + b)\,\phi(z) < 0$ on $(-b, \infty)$, $h(-b) = \phi(b) - b\,\Phi(-b) = \psi(-b) > 0$, and $h(z) \to -b < 0$ as $z \to \infty$. Hence $h$ has a unique zero $z_b$ on $(-b, \infty)$, $g$ is increasing on $(-b, z_b]$ and decreasing on $[z_b, \infty)$, and $\sup_{z > -b} g(z) = g(z_b)$. Since $\phi(z_b) = b\,\Phi(z_b)$ gives $\psi(z_b) = (z_b + b)\,\Phi(z_b)$, we obtain $g(z_b) = 1/\Phi(z_b) = c_b$.

\emph{The pointwise bound.} Fix $\mathbf{x}\in S$ and write $\sigma = \sigma_t(\mathbf{x})$. By the confidence bound, $f(\mathbf{x}) - \xi_t \le (\mu_t(\mathbf{x}) - \xi_t) + b\,\sigma$. If $\sigma = 0$, then $f(\mathbf{x}) - \xi_t \le \max(0, \mu_t(\mathbf{x}) - \xi_t) = \mathrm{EI}_t(\mathbf{x}) \le c_b\,\mathrm{EI}_t(\mathbf{x})$ because $c_b > 1$. Otherwise let $z = z_t(\mathbf{x})$, so that $f(\mathbf{x}) - \xi_t \le \sigma\,(z + b)$. If $z \le -b$, the right-hand side is non-positive and \eqref{eq:bes_pointwise} holds because $\mathrm{EI}_t \ge 0$. If $z > -b$, Lemma~\ref{lem:tau}(iii) gives $\sigma = \mathrm{EI}_t(\mathbf{x})/\psi(z)$, so $f(\mathbf{x}) - \xi_t \le \mathrm{EI}_t(\mathbf{x})\, g(z) \le c_b\,\mathrm{EI}_t(\mathbf{x})$. Taking the supremum over $S$ gives \eqref{eq:bes_sound}.

\emph{Tightness.} Take $\sigma_t(\mathbf{x}) = 1$, $\xi_t = 0$, $\mu_t(\mathbf{x}) = z_b$ and $f(\mathbf{x}) = z_b + b$. The confidence bound holds with equality, $\mathrm{EI}_t(\mathbf{x}) = \psi(z_b)$, and $f(\mathbf{x}) - \xi_t = z_b + b = c_b\,\psi(z_b)$.

\emph{Bounds on $c_b$.} $c_b > 1$ because $\Phi(z_b) < 1$. For $z > -b$, Lemma~\ref{lem:tau}(ii) gives $z \le \max(z, 0) \le \psi(z)$, and monotonicity gives $\psi(z) \ge \psi(-b)$, so $g(z) = z/\psi(z) + b/\psi(z) \le 1 + b/\psi(-b)$. Finally, $\Phi(z_b) = \phi(z_b)/b$ gives $c_b = b/\phi(z_b) = \sqrt{2\pi}\, b\, e^{z_b^2/2}$. If $b \ge 2\phi(0)$, then $h(0) = \phi(0) - b/2 \le 0$, so $-b < z_b \le 0$, hence $z_b^2 < b^2$ and $c_b < \sqrt{2\pi}\, b\, e^{b^2/2}$.
\end{proof}

Numerically, $c_1 \approx 2.6$, $c_2 \approx 17.2$ and $c_3 \approx 283$. The exponential growth in $b$ is therefore a property of the max-EI rule, not an artifact of the analysis: a point whose posterior mean lies far below the incumbent but which carries large uncertainty has little expected improvement and yet, on the confidence event, may exceed the incumbent by $c_b$ times that amount. The constant also inherits the growth of the confidence width: with the union-bound choice of Lemma~5.1 of \citet{Srinivas2010} on a finite domain, $e^{b_t^2/2} = |\mathcal{X}|\pi_{t+1}/\varrho$, so $c_{b_t}$ is of order $b_t|\mathcal{X}|\pi_{t+1}/\varrho$ and the uniform statement \eqref{eq:bes_sound} is qualitative. In the simulated world of a rollout, where the model is well specified (Remark~\ref{rem:roi_rollout}) and the incumbent is a function of the simulated data, a per-point statement needs no union bound. There $\mathrm{EI}^{(\tau)}(\mathbf{x})$ is exactly the posterior expectation of $\max(0, \tilde f(\mathbf{x}) - \xi^{(\tau)})$, so once the rule fires, Markov's inequality gives $\Pr\bigl(\tilde f(\mathbf{x}) - \xi^{(\tau)} \ge a \mid \text{simulated data}\bigr) \le \mathrm{EI}^{(\tau)}(\mathbf{x})/a < \delta/a$ for every $\mathbf{x} \in S$ and $a > 0$. Where the posterior width is small, the additive bound $\max(0, f(\mathbf{x}) - \xi_t) \le \mathrm{EI}_t(\mathbf{x}) + b\,\sigma_t(\mathbf{x})$ of \citet[Lemma~6]{nguyen2017regret} is the sharper statement. Combining \eqref{eq:bes_pointwise} with \eqref{eq:ei_bound} gives, on the confidence event and for every $\mathbf{x} \in S$,
\begin{equation*}
\frac{\max\bigl(0, f(\mathbf{x}) - \xi_t\bigr)}{c_b} \;\le\; \mathrm{EI}_t(\mathbf{x}) \;\le\; \max\bigl(0, f(\mathbf{x}) - \xi_t\bigr) + \Bigl(b + \frac{1}{\sqrt{2\pi}}\Bigr)\sigma_t(\mathbf{x}),
\end{equation*}
so the expected improvement tracks the improvement actually available at $\mathbf{x}$, up to the factor $c_b$ from below and up to a multiple of the posterior width from above. This two-sided relation is what makes the maximal expected improvement a principled stopping statistic on regions where the surrogate is consistent.

Because the region of interest keeps the optimizer on the confidence event (Lemma~\ref{lem:roi}(ii)), a max-EI test that inspects only the region also bounds the gap between the incumbent and the global optimum.

\begin{corollary}[The max-EI rule on the region of interest bounds the gap to the optimum]
\label{cor:bes_global}
Let Assumption~\ref{ass:confidence} hold, and work in the setting of \appref{subsec:roi_theory}, with $(\mu_t, \sigma_t)$ the posterior after $t$ observations, any incumbent $\xi_t$, and $b_t = \beta_{t+1}^{1/2}$, so that $\mu_t + b_t\,\sigma_t = \mathrm{UCB}_{t+1}$. With probability at least $1-\varrho$, the following holds simultaneously for all $t \ge 1$, all $\delta > 0$, and every set $S \subseteq \mathcal{X}$ that contains $\mathbf{x}^*$ whenever the event of Assumption~\ref{ass:confidence} holds; by Lemma~\ref{lem:roi}(ii), such sets include every region $\hat{\mathcal{X}}_s$, whether recomputed or held fixed, and every intersection of such regions. If $\sup_{\mathbf{x}\in S}\mathrm{EI}_t(\mathbf{x}) < \delta$, then
\begin{equation}
f^* - \xi_t \;\le\; \sup_{\mathbf{x}\in S}\mathrm{UCB}_{t+1}(\mathbf{x}) - \xi_t \;<\; \min\bigl\{c_{b_t}\,\delta,\; \delta + b_t\,s_t(S)\bigr\}.
\label{eq:bes_global}
\end{equation}
\end{corollary}

\begin{proof}
On the event of Assumption~\ref{ass:confidence}, $\mathbf{x}^* \in S$, and \eqref{eq:confidence} at index $t+1$ gives $f^* = f(\mathbf{x}^*) \le \mathrm{UCB}_{t+1}(\mathbf{x}^*) \le \sup_{\mathbf{x}\in S}\mathrm{UCB}_{t+1}(\mathbf{x})$, which is the first inequality. For the second, fix $\mathbf{x} \in S$ and write $b = b_t$. In each of its three cases, the proof of \eqref{eq:bes_pointwise} bounds $(\mu_t(\mathbf{x}) - \xi_t) + b\,\sigma_t(\mathbf{x})$ by $c_b\,\mathrm{EI}_t(\mathbf{x})$ without using the confidence bound, and $\mu_t(\mathbf{x}) - \xi_t \le \mathrm{EI}_t(\mathbf{x})$ by Lemma~\ref{lem:tau}(ii)--(iii) and the convention in \eqref{eq:ei_def}. Hence $\mathrm{UCB}_{t+1}(\mathbf{x}) - \xi_t \le \min\{c_b\,\mathrm{EI}_t(\mathbf{x}),\, \mathrm{EI}_t(\mathbf{x}) + b\,\sigma_t(\mathbf{x})\}$, and taking the supremum over $S$, with $\sup_S \mathrm{EI}_t < \delta$ and $\sup_S \sigma_t = s_t(S)$, gives the second inequality.
\end{proof}

The multiplicative branch of \eqref{eq:bes_global} is the bound that \eqref{eq:bes_sound} gives for a set containing $\mathbf{x}^*$, for which $\sup_S f = f^*$, so passing from the region to the global optimum loses nothing; its additive branch is the bound of \citet[Lemma~6]{nguyen2017regret} quoted above, applied at $\mathbf{x}^*$. Under union-bound widths, $c_{b_t}$ is of order $b_t|\mathcal{X}|\pi_{t+1}/\varrho$; the additive alternative $\delta + b_t\,s_t(S)$ is informative only once the posterior width on $S$ is small; and neither is sharper than the gap $\sup_S\mathrm{UCB}_{t+1} - \xi_t$, which is computable whenever $\xi_t$ and $\beta_{t+1}$ are known. The difference $f^* - \xi_t$ is the simple regret of a point when $\xi_t = f(\mathbf{x}_t^+)$. For an incumbent $\xi_t = \mu_t(\hat{\mathbf{x}}_t)$ defined by a point $\hat{\mathbf{x}}_t$, the same event gives $f^* - f(\hat{\mathbf{x}}_t) < c_{b_t}\delta + b_t\,\sigma_t(\hat{\mathbf{x}}_t)$. Under the premises of Remark~\ref{rem:roi_rollout}, and provided the width of $\hat{\mathcal{X}}_{m,t}$ is at least the rollout's confidence width at its first simulated step, the same conclusion holds conditionally on $\mathcal{D}_{t-1}$ at every simulated step $\tau$. It then holds with $\max_{\mathcal{X}}\tilde f$ in place of $f^*$, $\xi^{(\tau)}$ in place of $\xi_t$, $S = \hat{\mathcal{X}}_{m,t}$, and $b_t$ replaced by the rollout's confidence width at simulated step $\tau+1$. With a fixed width $\kappa$ and at most $L$ simulated steps, the failure probability is at most $(L+1)\,|\mathcal{X}|\,e^{-\kappa^2/2}$. This concerns the simulated objective only. It does not apply to the realized-improvement proxy of the released implementation.

\begin{remark}[The surrogate-consistency assumption cannot be dropped]
\label{rem:necessity}
Assumption~\ref{ass:regret} alone does not imply that the expected improvement vanishes, even pointwise. Suppose there are a point $\bar{\mathbf{x}} \in \mathcal{X}$ and constants $\underline{\sigma} > 0$ and $C < \infty$ such that $\sigma_t(\bar{\mathbf{x}}) \ge \underline{\sigma}$ and $|\mu_t(\bar{\mathbf{x}}) - \xi_t| \le C$ for all $t$. This is the situation of a region that the algorithm never samples, where the posterior keeps its prior scale, and it is compatible with $r^+_t \to 0$ because the incumbent may approach $f^*$ elsewhere. By Lemma~\ref{lem:tau}(i) and (iii), since $z_t(\bar{\mathbf{x}}) \ge -C/\underline{\sigma}$,
\begin{equation*}
\mathrm{EI}_t(\bar{\mathbf{x}}) = \sigma_t(\bar{\mathbf{x}})\,\psi\bigl(z_t(\bar{\mathbf{x}})\bigr) \ge \underline{\sigma}\,\psi\bigl(-C/\underline{\sigma}\bigr) > 0 \quad \text{for all } t,
\end{equation*}
so $\sup_{\mathbf{x} \in \mathcal{X}} \mathrm{EI}_t(\mathbf{x})$ does not converge to zero. The bound \eqref{eq:ei_bound} is moreover attained: when $\mu_t(\mathbf{x}) = f(\mathbf{x}) = \xi_t$, both sides equal $\sigma_t(\mathbf{x})/\sqrt{2\pi}$, and when $\sigma_t(\mathbf{x}) = 0$ and $\mu_t(\mathbf{x}) \ge f(\mathbf{x}) \ge \xi_t$, both sides equal $\mu_t(\mathbf{x}) - \xi_t$. Finally, pointwise convergence would not by itself yield uniform convergence: $\mathrm{EI}_t$ is not monotone in $t$ (a new high observation raises the posterior mean nearby and can raise the expected improvement), so Dini's theorem does not apply; the uniformity in Corollary~\ref{cor:ei_uniform} comes from the uniform quantities $\varepsilon_t(S)$ and $s_t(S)$ in the bound.
\end{remark}

\begin{remark}[When the assumptions hold]
\label{rem:when}
Assumption~\ref{ass:regret} is the goal of any sound BO policy; it holds, for instance, whenever the cumulative regret is sublinear and the incumbent is the best evaluated value $\xi_t = \max_{i \le t} f(\mathbf{x}_i)$, since then $0 \le r^+_t \le \frac{1}{t}\sum_{i \le t}\bigl(f^* - f(\mathbf{x}_i)\bigr)$. Assumption~\ref{ass:consistency} is a property of the surrogate and of the design, not of the policy. If confidence bounds $|\mu_t(\mathbf{x}) - f(\mathbf{x})| \le b_t\,\sigma_t(\mathbf{x})$ hold for all $\mathbf{x} \in S$, as on the event of Assumption~\ref{ass:confidence}, then $\varepsilon_t(S) \le b_t\, s_t(S)$, and the assumption is implied by $b_t\, s_t(S) \to 0$: the posterior width vanishing uniformly on $S$ faster than the confidence parameter grows. With noisy observations a single evaluation leaves positive posterior variance, so this requires the design to keep evaluating every part of $S$, at a rate that outpaces the growth of $b_t$; it is a property of the design that the stopping rule cannot itself guarantee. Within a DRO rollout the simulated objective is a draw from the posterior of $\mathrm{GP}_m$ (Remark~\ref{rem:roi_rollout}), so the surrogate is well specified by construction and the mean-error term is controlled by the posterior width on the confidence event. The width term, however, vanishes on the ROI only if the rollout keeps evaluating all of it, which acquisitions that concentrate near the incumbent need not do, and rollouts have finite length. For the rollouts, Assumption~\ref{ass:consistency} is therefore an idealization behind the asymptotic Corollary~\ref{cor:ei_uniform}; the non-asymptotic Theorem~\ref{thm:ei_convergence} and Corollary~\ref{cor:bes}, which concern only the set that the stopping rule inspects, are the statements that apply to them.
\end{remark}

\textit{Insight for Theorem~\ref{thm:ei_convergence} and Proposition~\ref{prop:bes_sound}:} The theorem decomposes the expected improvement into the improvement still available, a mean-error term and a posterior-width term. A successful policy drives the first to zero; the other two are properties of the surrogate on the region of interest, not of the policy. The maximal expected improvement is therefore a sound indicator of exhausted improvement precisely on regions where the surrogate is consistent, and, conversely, once it falls below $\delta$, no point of the region can improve on the incumbent by more than $c_b\,\delta$ on the confidence event of width $b$. Because the region keeps the optimizer on the event of Assumption~\ref{ass:confidence}, which covers all of $\mathcal{X}$ and every iteration, including the one at which the region was computed, the same bound applies to the gap to the global optimum of the objective for which the model is calibrated (Corollary~\ref{cor:bes_global}); in a rollout this is the simulated objective, and nothing follows for the real one (Remark~\ref{rem:roi_rollout}). This justifies its use as the termination signal of the ROI-constrained rollouts in \secref{subsec:early_stop_method}. Remark~\ref{rem:necessity} shows the limitation in the other direction: where parts of the region remain unexplored, the expected improvement need not vanish, so the rule need not fire.

\subsection{Nested Hypothesis Classes behind the Lengthscale Ensemble}
\label{subsec:lengthscale_theory}

Let $k_\theta(\mathbf{x},\mathbf{x}') = K\bigl((x_1 - x'_1)/[\theta]_1, \dots, (x_d - x'_d)/[\theta]_d\bigr)$ be a squared-exponential or Mat\'ern kernel with lengthscale vector $\theta \in \mathbb{R}^d_{>0}$ and unit signal variance (a common signal variance rescales all RKHS norms by the same factor and does not affect the statements below), let $\mathcal{H}_\theta$ be its RKHS on $\mathcal{X}$, let $[\theta]_i$ be the $i$-th coordinate of $\theta$, and read $\theta' \le \theta$ coordinatewise. Both kernel families satisfy the assumptions of \citet{Bull2011} (in particular, the Fourier transform of $K$ is isotropic and radially non-increasing), under which the following holds.

\begin{lemma}[Lemma~4 of \citealt{Bull2011}; Lemma~2 of \citealt{berkenkamp2019no}]
\label{lem:lengthscale}
    If $f \in \mathcal{H}_\theta$, then $f \in \mathcal{H}_{\theta'}$ for all $0 < \theta' \le \theta$, and
    \begin{equation}
        \|f\|^2_{\mathcal{H}_{\theta'}} \le \left( \prod_{i=1}^d \frac{[\theta]_i}{[\theta']_i} \right) \|f\|^2_{\mathcal{H}_\theta}.
    \end{equation}
\end{lemma}

\begin{corollary}[Coverage by the lengthscales of the ensemble]
\label{cor:lengthscale}
Let the ensemble use isotropic lengthscales $\ell_1 < \dots < \ell_M$, let $f \in \mathcal{H}_{\ell\mathbf{1}}$ for some $\ell \ge \ell_1$, where $\mathbf{1}$ is the all-ones vector, and let $m(\ell) = \max\{m : \ell_m \le \ell\}$. Then $f \in \mathcal{H}_{\ell_m\mathbf{1}}$ for every $m \le m(\ell)$, with $\|f\|^2_{\mathcal{H}_{\ell_m\mathbf{1}}} \le (\ell/\ell_m)^d\,\|f\|^2_{\mathcal{H}_{\ell\mathbf{1}}}$; the inflation factor is smallest for $m = m(\ell)$. If the lengthscales are spaced geometrically, $\ell_{m+1}/\ell_m = \rho$, and $\ell \le \ell_M$, this smallest factor is at most $\rho^d$.
\end{corollary}

\begin{proof}
For $m \le m(\ell)$, apply Lemma~\ref{lem:lengthscale} with $\theta = \ell\mathbf{1}$ and $\theta' = \ell_m\mathbf{1} \le \theta$, for which the product equals $(\ell/\ell_m)^d$; it decreases in $\ell_m$. For a geometric grid and $\ell \le \ell_M$, either $m(\ell) = M$, in which case $\ell = \ell_M$ and the factor is $1$, or $\ell_{m(\ell)} \le \ell < \ell_{m(\ell)+1} = \rho\,\ell_{m(\ell)}$, in which case $\ell/\ell_{m(\ell)} < \rho$.
\end{proof}

\textit{Justification for the lengthscale ensemble.} Lemma~\ref{lem:lengthscale} and Corollary~\ref{cor:lengthscale} say that the function classes are nested: shrinking the lengthscale can only enlarge the RKHS, at the price of inflating the squared norm by $\prod_i [\theta]_i/[\theta']_i$. Two consequences motivate the design of \secref{subsec:ensemble_gps}. First, an ensemble whose isotropic lengthscales span $[\ell_1, \ell_M]$ contains, for every objective whose isotropic lengthscale is at least $\ell_1$, members in whose RKHS the objective lies. For those members the RKHS confidence bounds (Theorem~6 of \citealt{Srinivas2010}) are valid for the real objective once their confidence parameter accounts for the inflated bound $(\ell/\ell_m)^d\|f\|^2_{\mathcal{H}_{\ell\mathbf{1}}}$ on the squared norm, which a fixed width does not do automatically, so the regions of interest of at least one member rest on a correct, if conservative, hypothesis without committing to a maximum-likelihood estimate that is unreliable early on. (The rollouts themselves sample posterior paths, which almost surely lie outside the RKHS; the coverage statement concerns the hypothesis class, not the simulated paths.) The lower end of the grid provides this coverage, which is why the ensemble includes small lengthscales, and the spacing of the grid controls the squared-norm inflation of the best-matched member. The inflation is exponential in $d$ (for instance, $\rho^d \approx 1.67^d$ for ten lengthscales spaced geometrically in $[0.1, 10]$), so the corollary describes which hypothesis classes the ensemble covers rather than providing a useful norm bound in high dimension. An objective with anisotropic lengthscales $\theta$ is covered in the same sense by every isotropic member with $\ell_m \le \min_i [\theta]_i$, at inflation $\prod_{i=1}^d ([\theta]_i/\ell_m)$. Second, the no-regret result of \citet{berkenkamp2019no} additionally requires the lengthscales to \emph{decrease} and the norm bound to grow \emph{over time}, so that every objective in the hypothesis class eventually lies in the candidate function class; a fixed grid approximates that schedule only down to its smallest lengthscale ($0.1$ in our implementation) and offers no guarantee for objectives that require a shorter one, which we state as a limitation rather than a transferred guarantee. We use isotropic lengthscales so that the grid is one-dimensional and the ensemble stays small ($M \le 10$) in high-dimensional spaces; the mixed-kernel variant of \appref{subsec:kernel_diversity} shows that the same mechanism tolerates kernel diversity as well.

\section{Extensibility of the DRO Framework: Constrained and Multi-Fidelity Optimization}
\label{sec:dro_extensions}

A significant benefit of the DRO framework is its inherent flexibility and potential for extension to more complex optimization scenarios. The core principle of learning a non-myopic decision-making policy from simulated trajectories can be adapted by appropriately modifying the simulation environment, state representation, and action space. This section outlines conceptual extensions of DRO to two common and important settings: Bayesian optimization with unknown constraints and multi-fidelity Bayesian optimization.

\subsection{DRO for Bayesian Optimization with Unknown Constraints}

When optimizing an objective $f(\mathbf{x})$ subject to $J$ unknown black-box constraints $\mathcal{C}_j(\mathbf{x}) \ge 0$, DRO can be adapted by primarily modifying the simulation phase and state representation. The objective remains to train a Decision Transformer to directly minimize the simple regret with respect to the true (constrained) optimum. Key considerations include:

\begin{enumerate}
    \item \textbf{Ensemble GP models for objective and constraints:} Independent ensembles of Gaussian Process (GP) models would be maintained for the objective function and for each of the $J$ constraints. This allows for capturing model uncertainty for all unknown functions. All GPs would be conditioned on all available real observations (objective and constraint values).

    \item \textbf{Constrained region of interest for simulations:} During the simulation (rollout generation) for a specific set of sampled models (one objective GP and one GP for each constraint from their respective ensembles), a \textbf{constrained ROI} must be identified. This can be achieved by adapting ROI identification strategies from the constrained Bayesian optimization (CBO) literature, such as the approach in \citet{zhang2024finding}. This involves:
    \begin{itemize}
        \item Defining an ROI for each constraint $\mathcal{C}_j$ (based on its current GP model $\mathrm{GP}_{\mathcal{C}_j}$) to include regions likely to satisfy the constraint (e.g., where $\mathrm{UCB}_{\mathcal{C}_j}(\mathbf{x}) \ge 0$).
        \item Forming a joint feasible ROI by intersecting these individual constraint ROIs.
        \item Defining an objective ROI based on the current objective GP model ($\mathrm{GP}_f$), potentially using a threshold derived from its LCB within the high-confidence jointly feasible region.
        \item The final ROI for simulation is the intersection of the objective ROI and the joint feasible ROI. This ensures simulated rollouts explore regions deemed both promising for the objective and likely feasible by the set of models defining that particular simulated world.
    \end{itemize}

    \item \textbf{Constraint-aware rollout generation:} Simulated trajectories are generated by iteratively selecting points within the dynamically identified constrained ROI. The acquisition function used within the simulation must be constraint-aware (e.g., by multiplying a standard acquisition function with the joint probability of feasibility estimated from the constraint GPs, or by using specialized CBO acquisition functions). Observations for both the objective and constraints are sampled from their respective GP posterior predictive distributions.

    \item \textbf{Augmented state representation:} The state $\mathbf{s}_\tau$ for the Decision Transformer must include information about the constraints, such as hyperparameters of the constraint GPs, historical constraint observations, and potentially features describing the estimated feasible region.

    \item \textbf{Training objective:} The Decision Transformer is trained to predict actions that minimize the final simple regret with respect to the true \emph{constrained} optimum, using return-to-go signals calculated based on the best \emph{feasible} objective value found in simulated trajectories.
\end{enumerate}
This conceptual extension enables DRO to learn a non-myopic policy that navigates the trade-offs inherent in constrained optimization.

\subsection{DRO for Multi-Fidelity Bayesian Optimization (MFBO)}

DRO can also be extended to MFBO settings, where the objective function can be evaluated at different fidelity levels $q \in \mathcal{Q}$, each with an associated cost $c(q)$. The goal is to find the high-fidelity optimum while minimizing total evaluation cost. This extension builds upon the core DRO machinery with the following key adaptations:

\begin{enumerate}
    \item \textbf{Fidelity-specific GP ensembles:} Independent ensembles of GPs are maintained for each fidelity level, i.e., $\{\mathrm{GP}_{m}^{(q)}\}_{m=1}^{M_q}$ for each $q \in \mathcal{Q}$. Each $\mathrm{GP}_{m}^{(q)}$ is trained on data observed at fidelity $q$. Relationships between fidelities (e.g., via auto-regressive models or deep kernels as in \citealt{zhang2025robust_mfbo}) could also be incorporated into these GP models if desired, though simpler independent models are a first step.

    \item \textbf{Fidelity-aware and cost-aware ROI for simulations:} When generating rollouts, the ROI identification can be made fidelity-aware. For instance, drawing inspiration from methods like those in \citet{zhang2025robust_mfbo} that identify robust partitions or regions of agreement across fidelities, the simulation environment for DRO could define ROIs based on information from one or more fidelity-specific GPs. The decision of which fidelity's GP(s) to use for ROI definition within a simulation could be part of the diversity generation.

    \item \textbf{Joint action space (point and fidelity):} The Decision Transformer's action becomes a pair $(\mathbf{x}_t, q_t)$, selecting both the point to evaluate and the fidelity level at which to perform the evaluation.

    \item \textbf{Augmented state representation:} The state $\mathbf{s}_\tau$ must include fidelity-related information:
    \begin{itemize}
        \item Historical data, including the fidelity level and cost of each past evaluation.
        \item Current cumulative cost and remaining budget (if applicable).
        \item Hyperparameters or summary statistics from the fidelity-specific GP ensembles.
    \end{itemize}

    \item \textbf{Cost-sensitive rollout simulation and policy learning:} Simulated trajectories now involve selecting fidelities for simulated queries, incurring simulated costs. The Bayesian early stopping criterion might be adapted to consider the cost-benefit of continuing a rollout. The return-to-go signal for training the Decision Transformer would still relate to the final (high-fidelity) simple regret, but the policy must learn to achieve this efficiently with respect to the costs associated with different fidelities.
\end{enumerate}
This extension would allow DRO to learn a non-myopic policy that strategically chooses both query locations and fidelity levels to optimize the high-fidelity objective under budget or cost constraints.

\vspace{1em} % Add some vertical space before the next section
These two examples illustrate that the DRO paradigm---leveraging ensemble-based simulations to train a Decision Transformer for direct regret minimization---provides a flexible foundation. By appropriately designing the components of the simulated environment (surrogate models, ROI definitions, state-action spaces) and the information fed to the Decision Transformer, DRO has the potential to be tailored to a variety of complex Bayesian optimization settings beyond the standard unconstrained case.

\section{Additional Implementation Details}
\label{sec:implementation_details}
The following subsections provide further details on key hyperparameters and configurations used for the DRO framework components. These correspond to the settings in the provided configuration files and are reflected in the implementation.

\subsection{Decision Transformer}
The collection of simulated trajectories from all GPs in the ensemble forms an offline dataset for training a \emph{Decision Transformer} \citep{chen2021decision}. This model is chosen for its ability to handle sequential decision-making problems by framing them as sequence modeling tasks. Each trajectory $(\mathbf{s}_0, \mathbf{a}_0, r_0, \mathbf{s}_1, \mathbf{a}_1, r_1, \dots)$ is processed into sequences of \emph{(state, action, return-to-go)} tuples suitable for the Decision Transformer:
\begin{itemize}
    \item \textbf{State ($\mathbf{s}_\tau$)}: An embedding representing the state of the BO process at simulation step $\tau$. This can include features derived from the current $\mathrm{GP}_m$ (used for that rollout), the history of simulated observations $(\mathbf{x}_i^{(m)}, y_i^{(m)})$ within that rollout (e.g., best value found, % $f(\mathbf{x}_{\text{best\_simulated}}^{(m)})$, 
    number of steps taken), and context from the real BO process (e.g., current real iteration $t$, real historical data $\mathcal{D}_{t-1}$).
    \item \textbf{Action ($\mathbf{a}_\tau$)}: The point $\mathbf{x}_\tau^{(m)}$ selected by the conventional acquisition function (within $\hat{\mathcal{X}}_{m,t}$) during the simulation step $\tau$ for $\mathrm{GP}_m$.
    \item \textbf{Return-to-go ($R_\tau$)}: A crucial signal for the Decision Transformer, representing the cumulative future reward from step $\tau$ to the end of the trajectory. It is the improvement still to be realized by the end of the trajectory: for a simulated trajectory of at most $L$ steps with incumbent $f_{m,\text{best}}^{(\tau)}$ after step $\tau$ (held at its final value once the rollout stops), $R_\tau = f_{m,\text{best}}^{(L)} - f_{m,\text{best}}^{(\tau)}$, i.e., the sum of the future simulated improvements, exactly as in \secref{subsec:dt}; it is zero at the end of every trajectory and equals the trajectory's remaining simple regret relative to its own terminal incumbent.
\end{itemize}
The Decision Transformer is trained using this offline data to predict an action $\mathbf{a}_\tau$ given a history of states, actions, and a target return-to-go. During the \emph{real} BO procedure, the trained Decision Transformer is conditioned on the current real BO state (constructed similarly, using the ensemble and real data $\mathcal{D}_{t-1}$) and a high target return (i.e., low target regret) to propose the next query point $\mathbf{x}_t$.

\begin{itemize}
    \item \textbf{Architecture}:
    \begin{itemize}
        \item \textbf{Hidden size}: The embedding dimension or hidden size of the transformer is 128.
        \item \textbf{Number of layers}: The transformer has 4 layers.
        \item \textbf{Number of heads}: 4 attention heads are used in the multi-head attention mechanisms.
        \item \textbf{Dropout}: A dropout rate of 0.1 is applied.
    \end{itemize}
    \item \textbf{Training}:
    \begin{itemize}
        \item \textbf{Learning rate}: The Decision Transformer is trained with a learning rate of $1 \times 10^{-4}$.
        \item \textbf{Weight decay}: A weight decay of $1 \times 10^{-5}$ is used.
        \item \textbf{Batch size}: The batch size for training is 32.
        \item \textbf{Number of epochs}: The transformer is trained for 100 epochs on the simulated trajectory data collected in each BO iteration.
    \end{itemize}
    \item \textbf{Sequence length}: The maximum sequence length processed by the transformer is 20.
    \item \textbf{State dimension}: The state dimension for the transformer input is dynamically calculated. It includes features for each GP in the ensemble (e.g., key hyperparameters like lengthscale and outputscale, typically 2 parameters per GP), the best observed objective value so far, the current iteration number (or a normalized version), and the coordinates of the best-known points.
    \item \textbf{Action dimension}: This corresponds to the input dimension of the black-box function being optimized.
    \item \textbf{Return-to-go calculation}: During training, the return-to-go for a step in a simulated trajectory is the sum of future rewards (improvements achieved in the simulation in terms of simple regret) until the end of that trajectory. For inference (proposing the next real query point), a high target return-to-go (e.g., 1.0, assuming rewards are normalized or scaled appropriately, or equivalently zero simple regret) is used to prompt the transformer to generate an action aimed at achieving this high return.
\end{itemize}

\subsection{Gaussian Process Ensemble}
\begin{itemize}
    \item \textbf{Number of models ($M$)}: The ensemble consists of $M=10$ Gaussian Process (GP) models by default.
    \item \textbf{Kernel}: The GPs utilize an RBF (Radial Basis Function) kernel by default. The configuration allows for other kernels like Mat\'ern or rational-quadratic, though RBF is the specified default.
    \item \textbf{Kernel hyperparameters}:
    \begin{itemize}
        \item \textbf{Lengthscale}: Initial lengthscales for the ensemble models are sampled from a range between a minimum of 0.1 and a maximum of 10.0.
        \item \textbf{ARD (automatic relevance determination)}: ARD is not enabled by default (\texttt{ard: False}).
    \end{itemize}
    \item \textbf{Likelihood noise}: The Gaussian likelihood for each GP has a noise constraint, with the lower bound set to $1 \times 10^{-4}$.
    \item \textbf{Training}:
    \begin{itemize}
        \item \textbf{Retraining}: GPs are not retrained from scratch at each BO iteration (\texttt{retrain: False}); their hyperparameters are updated based on new data.
        \item \textbf{Optimizer}: GP hyperparameters are optimized using an Adam optimizer.
        \item \textbf{Learning rate}: The learning rate for the GP hyperparameter optimization is 0.1.
        \item \textbf{Training iterations}: Each GP model is trained for 50 iterations when updated.
    \end{itemize}
\end{itemize}

\subsection{Bayesian Early Stopping (BES)}
\begin{itemize}
    \item \textbf{Activation}: The Bayesian early stopping mechanism is active during the simulation rollouts.
    \item \textbf{Threshold ($\delta$)}: A rollout is terminated as soon as the realized simulated improvement of the sampled point, $\max(0, y_\tau - f_{m,\text{best}}^{(\tau-1)})$, falls below $\delta = 1 \times 10^{-4}$ (a one-sample proxy for the criterion in \eqnref{eq:bes_ei}, not covered by the analysis of \appref{subsec:ei_theory}).
    \item \textbf{Maximum length ($L$)}: Independently of BES, every rollout is capped at $L$ simulated steps (see the simulation budget below).
\end{itemize}

\subsection{Simulation Budget and Training Schedule}
\label{subsec:sim_budget}
The released configurations specify the number of rollouts simulated per real iteration, which are split evenly across the $M$ ensemble members ($K = \lfloor \text{rollouts}/M \rfloor$ per member), and the maximum rollout length $L$:
\begin{itemize}
    \item \textbf{Default configuration} (\texttt{dro.yaml}, used for the Ackley 2D--10D and HPO tasks): 10 rollouts per iteration, $L = 4$.
    \item \textbf{Long-rollout configuration} (\texttt{dro\_long.yaml}, used for Ackley 20D): 40 rollouts per iteration with $M=10$ ($K=4$), $L = 100$.
    \item \textbf{Short-rollout configuration} (\texttt{dro\_simple.yaml}, used for LunarLander): 20 rollouts per iteration with $M=5$ ($K=4$), $L = 2$.
\end{itemize}
The acquisition used for each maximization inside a rollout is drawn uniformly at random from the pool $\{\mathrm{EI}, \mathrm{UCB}, \mathrm{PI}, \mathrm{MES}\}$. The Decision Transformer and its Adam optimizer are instantiated once per trial and warm-started across real iterations: at every iteration the model is fine-tuned for 100 epochs on the buffer simulated at that iteration (mean-squared error on the predicted action, gradient norm clipped at $1.0$), and at inference it is conditioned on the current real state and a target return of $1.0$. The return-to-go of a simulated step is the sum of the future simulated improvements along its trajectory.

\subsection{Acquisition Functions for Simulation}
\begin{itemize}
    \item \textbf{Strategy}: During the simulation phase for generating trajectories, DRO employs a strategy of rotating through different acquisition functions.
    \item \textbf{Candidate functions}: The pool of acquisition functions for rotation includes Expected Improvement (EI), Upper Confidence Bound (UCB), Probability of Improvement (PI), and Max-value Entropy Search (MES).
    \item \textbf{Hyperparameters}:
    \begin{itemize}
        \item \textbf{UCB $\kappa$ (or $\beta$)}: The $\kappa$ parameter for UCB is set to 2.0.
        \item \textbf{EI/PI $\xi$}: The trade-off parameter $\xi$ for EI and PI is 0.01.
        \item \textbf{MES samples}: For Max-value Entropy Search, the number of max-value samples is 10.
    \end{itemize}
    \item \textbf{ROI constraint during optimization}: When optimizing these acquisition functions within the simulation rollouts, the search space is constrained by the UCB $\ge$ max(LCB) criterion. The $\kappa$ value used for defining this UCB/LCB constraint is 6.0. This fixed value stands in for the iteration-dependent width $\beta_t^{1/2}$ of Theorem~\ref{thm:roi_regret} (the region uses $\mu \pm \kappa\sigma$): the theory lets $\beta_t$ grow slowly with $t$ so that the confidence bounds hold at every iteration, whereas a fixed $\kappa$ keeps the region from widening over time at the price of that guarantee; we note this gap between the analysis and the implementation explicitly.
\end{itemize}

\section{Additional Experimental Results}
\label{sec:additional_results}

\subsection{Effect of Simulated Acquisition Strategy}
\begin{figure}[htbp!] % Using [htbp!] for placement preference
    \centering
    \includegraphics[width=0.6\textwidth]{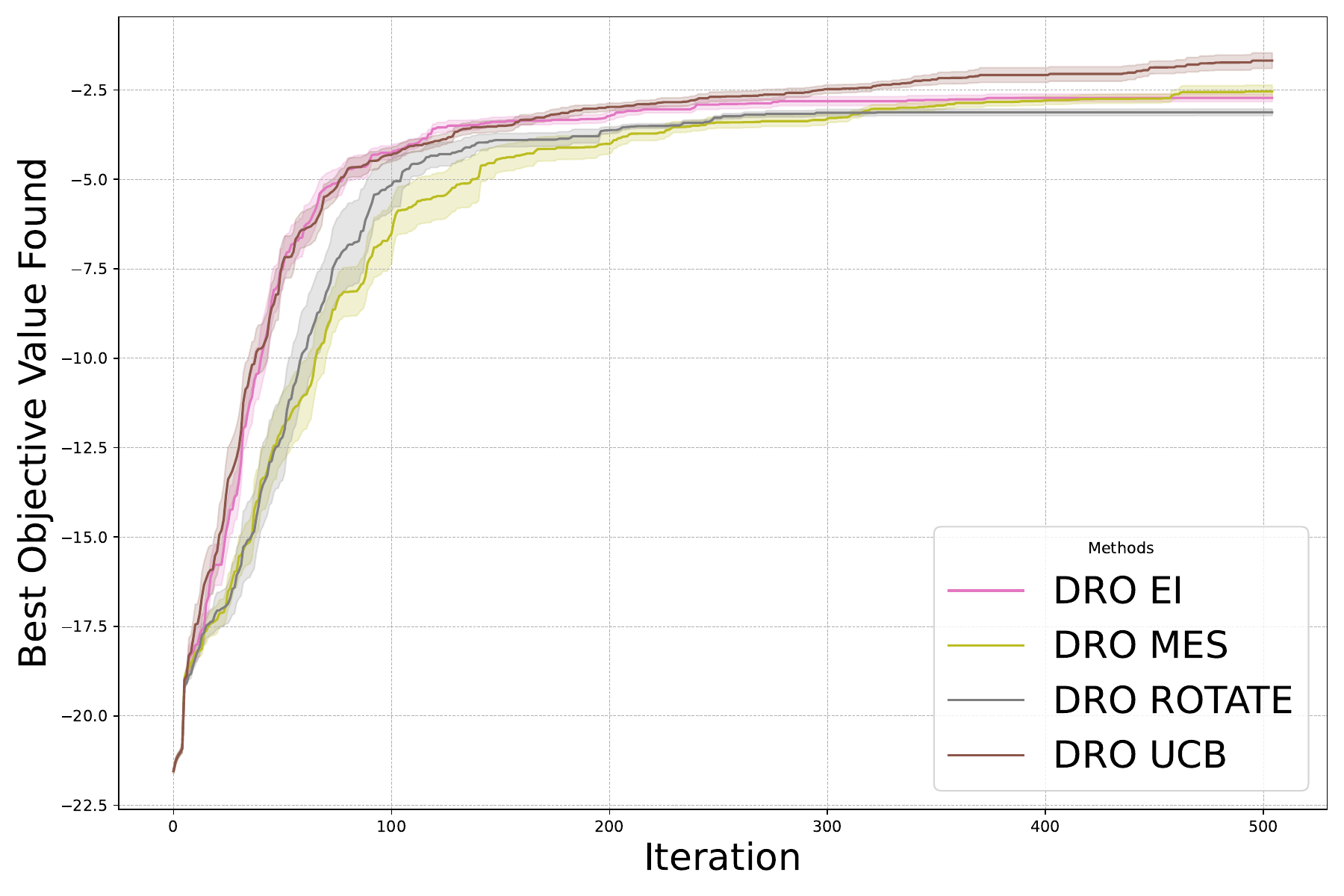} % Adjust width as needed
    \caption{Performance comparison of different acquisition function strategies used within DRO's simulation rollouts on the Ackley 10D function (Best Objective Value Found). `DRO ROTATE' refers to the strategy of cycling through EI, UCB, PI, and MES for generating simulated trajectories. Higher values are better.}
    \label{fig:ackley_10d_acq_study}
    % \vspace{-4mm} % Optional: adjust vertical space if needed
\end{figure}

To evaluate the impact of the acquisition function strategy used within the simulation rollouts, we compared DRO variants using fixed acquisition functions (Expected Improvement---logEI, Max-value Entropy Search---MES, Upper Confidence Bound---UCB) against our default strategy of rotating through EI, UCB, PI and MES (`DRO ROTATE'). \figref{fig:ackley_10d_acq_study} presents these results on the Ackley 10D function. It is well known that there is no silver bullet acquisition function across all scenarios. The `DRO ROTATE' strategy demonstrates robust performance, achieving competitive results comparable to using a single fixed acquisition function throughout the optimization. This supports its choice as a means to diversify the simulated trajectories.

\subsection{Simpler Ackley Benchmark (Reproduced logEI Setting)}
\label{subsec:simpler_ackley_logei}
To complement the main Ackley experiments (which include Gaussian observation noise with standard deviation $0.01$, a $+10$ shift of the input in every coordinate, and the full symmetric domain $[-32.768, 32.768]^D$), we reproduce the simplified evaluation protocol used in the logEI study \citep{ament2023unexpected}. This additional benchmark (dimensions 2D, 8D, 16D) differs from our primary setup in three key ways:
\begin{enumerate}
    \item \textbf{Noiseless, unshifted objective:} We remove observation noise and do not apply the input shift; the global optimum retains its canonical central location prior to domain truncation.
    \item \textbf{Smaller, asymmetric domain:} Instead of the standard hypercube, each dimension is restricted to an asymmetric interval: $[(-32.768/4,\ 32.768/2)]$, i.e.,
    \texttt{bounds = [(-32.768 / 4, 32.768 / 2) for \_ in range(dim)]},
    contrasting with the main setting's $[-32.768, 32.768]^D$.
    \item \textbf{Larger initialization set:} The number of Sobol initialization points follows the logEI protocol: $N_{\text{init}} = 2 (d + 1)$ (instead of the smaller fixed $N_{\text{init}}=5$ typically used in our primary experiments).
\end{enumerate}

All other implementation details (GP ensemble with RBF kernels, ROI-based rollouts, Decision Transformer architecture, and acquisition rotation during simulation) remain unchanged unless incompatible with the altered initialization size or noiseless setting.

\begin{figure}[htbp]
    \centering
    \includegraphics[width=\linewidth]{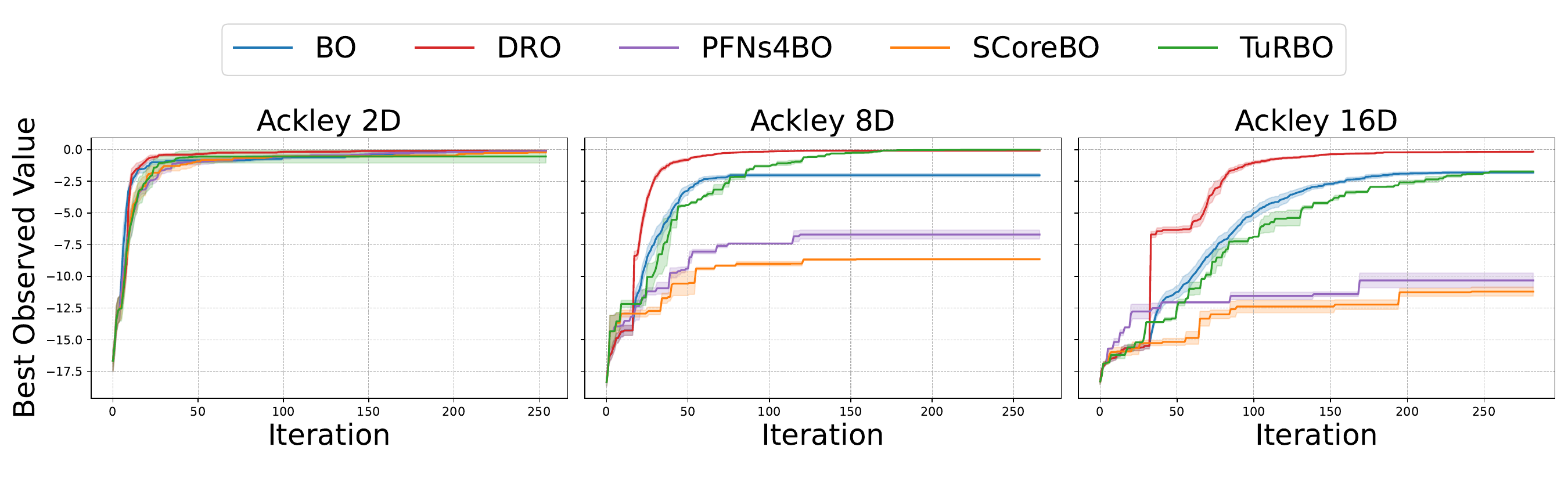}
    \caption{Performance on the simplified Ackley benchmark (2D, 8D, 16D) reproducing the logEI study setting \citep{ament2023unexpected}. Differences from the main Ackley experiments: (i) noiseless observations; (ii) no output shift; (iii) reduced asymmetric domain $[(-32.768/4, 32.768/2)]^D$; (iv) Sobol initialization with $N_{\text{init}} = 2(d+1)$. DRO performs best or near-best across all shown dimensions. Higher is better (best objective value found).}
    \label{fig:ackley_logei_reproduce}
\end{figure}

\paragraph{Results.} Figure~\ref{fig:ackley_logei_reproduce} presents the best objective value found over the budget for dimensions 2D, 8D, and 16D under this simplified protocol. DRO performs best or near-best across all three dimensions when compared to standard BO with logEI and other baselines, mirroring (and in some cases amplifying) the relative performance gains observed in the main paper's noisier, shifted, full-domain experiments. This consistency underscores DRO's robustness to evaluation protocol variations and indicates that its advantages are not artifacts of the noise model, the global optimum shift, or the larger symmetric domain.

\subsection{Execution Time Analysis}
\label{subsec:time}
We provide a brief analysis of the execution time of DRO compared to other baselines. Table~\ref{tab:time_comparison} reports the wall time in seconds per iteration, measured on the Ackley 8D and 16D runs of the simplified protocol of \appref{subsec:simpler_ackley_logei} with the default configuration, on a MacBook Pro with an M1 Pro chip.

\begin{table}[h]
\centering
\caption{Comparison of wall time (seconds/iteration) on Ackley benchmarks.}
\label{tab:time_comparison}
\begin{tabular}{l|cc}
\toprule
\textbf{Method} & \textbf{Ackley 8D} & \textbf{Ackley 16D} \\
\midrule
BO & 0.154 & 0.145 \\
DRO & 3.680 & 3.548 \\
SCoreBO & 1.411 & 1.344 \\
TuRBO & 0.283 & 0.484 \\
PFNs4BO & 0.585 & 0.590 \\
\bottomrule
\end{tabular}
\end{table}

The results show that DRO is computationally affordable, with each iteration taking only a few seconds. While it is slower than some baselines like standard BO and TuRBO, it is within an order of magnitude of other state-of-the-art methods like SCoreBO and PFNs4BO.

The per-iteration times in Table~\ref{tab:time_comparison} are for the default configuration (10 rollouts of at most $L=4$ simulated steps per iteration; \appref{subsec:sim_budget}). The Ackley 20D runs use the long-rollout configuration (40 rollouts of up to $L=100$ steps), i.e., up to two orders of magnitude more simulated steps per iteration, which explains why the 10-trial Ackley 20D experiment takes above 33 hours on a shared server with an AMD 96-core CPU, while a batch of 10 trials of the default configuration under the simplified protocol on an M1 Pro Mac consumes between 1 and 2 hours without parallelization, compared to baselines that finish within 5 minutes on the same MacBook. Early-stopping behavior and hardware differences make a perfectly consistent comparison difficult. Another point we want to raise is that for pre-trained models like PFNs4BO, the pre-training cost depends largely on the scope of pre-training data curation. Regarding the ensemble size $M$, our ablation study shows diminishing marginal gains as it increases. This suggests that practitioners can make a case-by-case decision, trading computational resources for gains in sample efficiency based on their specific needs. We do want to mention that the code we provided uses Hydra, which keeps the running time in its log output.

\subsection{Ablation Study: Decision Transformer vs. PPO Policy}
\label{subsec:ppo_ablation}
To further validate our choice of the Decision Transformer as the policy model, we conducted an ablation study comparing our standard DRO approach with a variant, which we term \textbf{DRO-PPO}. In this variant, the Decision Transformer is replaced with a Proximal Policy Optimization (PPO)-based policy, a widely used policy-gradient method in RL. This study directly tests the efficacy of our direct regret conditioning approach against a more traditional policy-gradient method within the same robust, ensemble-based simulation framework. We note that this ablation is distinct from other RL-based BO methods like EARL-BO, which learns a policy within a single, non-ensembled GP environment.

The results are presented in Tables~\ref{tab:ppo_ultimate} and \ref{tab:ppo_intermediate}. Table~\ref{tab:ppo_ultimate} shows the final best objective value found, while Table~\ref{tab:ppo_intermediate} shows the performance at iteration 100 to evaluate sample efficiency.

\begin{table}[htbp!]
\centering
\caption{Ultimate Results (Best Objective Value Found): DRO vs. DRO-PPO. Best results are in bold.}
\label{tab:ppo_ultimate}
\resizebox{\textwidth}{!}{%
\begin{tabular}{l|ccccccccc}
\toprule
\textbf{Method} & \textbf{ackley\_2d} & \textbf{ackley\_5d} & \textbf{ackley\_10d} & \textbf{ackley\_20d} & \textbf{xgboost} & \textbf{lunarlander} & \textbf{adam\_wine\_acc} & \textbf{adam\_breast\_acc} & \textbf{adam\_iris\_acc} \\
\midrule
DRO-PPO & \textbf{-0.04 $\pm$ 0.01} & \textbf{-0.87 $\pm$ 0.22} & \textbf{-2.69 $\pm$ 0.16} & -17.09 $\pm$ 0.00 & \textbf{0.940 $\pm$ 0.004} & 291.69 $\pm$ 33.09 & \textbf{0.873 $\pm$ 0.000} & 0.919 $\pm$ 0.000 & \textbf{0.992 $\pm$ 0.000} \\
DRO & -0.12 $\pm$ 0.02 & -1.69 $\pm$ 0.17 & -3.12 $\pm$ 0.10 & \textbf{-4.39 $\pm$ 0.19} & 0.938 $\pm$ 0.004 & \textbf{344.12 $\pm$ 66.55} & 0.861 $\pm$ 0.014 & \textbf{0.922 $\pm$ 0.002} & 0.982 $\pm$ 0.001 \\
\bottomrule
\end{tabular}%
}
\end{table}

\begin{table}[htbp!]
\centering
\caption{Results at Iteration 100 (Intermediate Performance): DRO vs. DRO-PPO. Best results are in bold.}
\label{tab:ppo_intermediate}
\begin{tabular}{l|ccccc}
\toprule
\textbf{Method} & \textbf{ackley\_2d} & \textbf{ackley\_5d} & \textbf{ackley\_10d} & \textbf{ackley\_20d} & \textbf{lunarlander} \\
\midrule
DRO-PPO & -0.68 $\pm$ 0.17 & -2.89 $\pm$ 0.37 & \textbf{-4.49 $\pm$ 0.21} & -17.09 $\pm$ 0.00 & 265.42 $\pm$ 33.45 \\
DRO & \textbf{-0.57 $\pm$ 0.23} & \textbf{-2.70 $\pm$ 0.27} & -5.19 $\pm$ 0.78 & \textbf{-10.37 $\pm$ 0.67} & \textbf{317.22 $\pm$ 66.76} \\
\bottomrule
\end{tabular}
\end{table}

The final-performance picture is mixed: DRO-PPO attains better final values on the low-dimensional Ackley tasks, on XGBoost and on two of the three Adam tasks, while DRO is markedly better on Ackley 20D (where DRO-PPO collapses) and nominally better on LunarLander (overlapping standard errors) and Adam Breast. At iteration 100, DRO leads on four of the five tasks. Since both variants share the ensemble, the ROI and the rollout simulation and differ only in the policy head, we read this ablation as evidence that return conditioning is the more sample-efficient and the more robust choice in the hardest settings, not as evidence of uniform superiority over policy gradients.

\subsection{Matched-Budget Comparison with Multi-Step Lookahead Baselines}
\label{subsec:lookahead_appendix}
\tabref{tab:lookahead} reports the best observed objective at iteration 100 on Ackley 5D and 10D for logEI, KG, TuRBO, DRO, and two explicit multi-step methods implemented as one-shot multi-step trees \citep{jiang2020efficient} in BoTorch: two-step EI \citep{wu2019practical} and two-step KG. All arms use the same ten seeds, the same Sobol initialization and the same noisy, shifted Ackley protocol as the main experiments; the budget is $T=100$ because one-shot trees are rarely run at the $T=500$, up-to-20D scale of \figref{fig:ackley_dim_hpo_study}. The two-step trees do not improve over their one-step counterparts on this protocol (two-step EI is no better than logEI, and two-step KG is behind one-step KG on 10D), so DRO's advantage over the lookahead family does not reduce to ``any non-myopic method would do''.

\begin{table}[htbp]
\centering
\caption{Matched-budget comparison with lookahead acquisitions on Ackley (best observed objective at iteration $100$; mean $\pm$ standard error over 10 trials, higher is better).}
\label{tab:lookahead}
\begin{tabular}{lcc}
\toprule
\textbf{Method} & \textbf{Ackley 5D} & \textbf{Ackley 10D} \\
\midrule
BO (logEI)            & $-16.56 \pm 0.42$ & $-19.29 \pm 0.10$ \\
Two-step EI           & $-16.87 \pm 0.42$ & $-19.69 \pm 0.22$ \\
KG (one-step)         & $-5.58 \pm 0.28$  & $-5.74 \pm 0.31$ \\
Two-step KG           & $-5.19 \pm 0.44$  & $-7.61 \pm 0.25$ \\
TuRBO                 & $-5.12 \pm 1.37$  & $-7.13 \pm 0.88$ \\
DRO                   & $\mathbf{-2.63 \pm 0.22}$ & $\mathbf{-5.05 \pm 0.71}$ \\
\bottomrule
\end{tabular}
\end{table}

\paragraph{Per-iteration comparison with KG.} Because a policy trained against terminal regret may sacrifice early progress, we compared the incumbents of DRO and KG at every iteration of the same runs rather than only at the reported checkpoints. On Ackley 10D, DRO is behind KG and two-step KG at iteration 50 ($-11.13 \pm 1.18$ versus $-9.33 \pm 1.28$ and $-9.13 \pm 0.65$), remains behind KG until iteration $\approx 82$, and leads for the rest of the run ($+0.69$ at iteration 100, with DRO continuing to improve to $-3.12$ by iteration 500 while KG in \figref{fig:ackley_dim_hpo_study} falls further behind). On Ackley 5D, DRO leads at every iteration ($+2.05$ at iteration 25, $+2.96$ at iteration 100, with DRO reaching $-1.69$ by iteration 500). We therefore do not claim that DRO leads at every budget; it is best or near-best at the full budget on every task we report, and where it trails KG on Ackley it is early in the run.

\subsection{Component Attribution: Region of Interest versus Learned Policy}
\label{subsec:attribution}
\tabref{tab:attribution} removes each component of DRO in turn on Ackley 10D, keeping the GP ensemble ($M=10$, RBF kernels, lengthscales in $[0.1, 10]$), the ROI rule ($\mathrm{UCB} \ge \max \mathrm{LCB}$ with $\kappa = 6$), the seeds and the initialization fixed. DRO-GLOBAL keeps the transformer but simulates rollouts over the whole domain; DRO-NoDT keeps the ensemble, the ROI and the rotating acquisition pool but maximizes the acquisition inside the ROI directly, without rollouts or a transformer; ROI-logEI selects, at every iteration, the member with the highest marginal likelihood on the observed data and maximizes logEI restricted to that member's ROI. The two no-transformer arms were run in a rebuilt environment (pinned to the versions in the released requirements), so a fresh DRO control run under identical code, environment and seeds is reported alongside the stored paper run; the fresh control finishes slightly above the stored run ($-2.77 \pm 0.13$ versus $-3.12 \pm 0.10$), consistent with minor simulation-hyperparameter differences, and every paired comparison below is made against the fresh control.

\begin{table}[htbp]
\centering
\caption{Component attribution on Ackley 10D (best observed objective, mean $\pm$ standard error over 10 trials, higher is better).}
\label{tab:attribution}
\small
\begin{tabular}{lccc}
\toprule
\textbf{Variant} & \textbf{it $=100$} & \textbf{it $=200$} & \textbf{it $=500$} \\
\midrule
logEI, global (paper run)                     & $-19.29 \pm 0.11$ & $-18.89 \pm 0.27$ & $-18.49 \pm 0.27$ \\
DRO-GLOBAL (transformer, no ROI)              & --                & --                & $-15.12 \pm 0.33$ \\
ROI-logEI (one acquisition, no transformer) & $-18.29 \pm 1.01$ & $-16.17 \pm 1.51$ & $-12.11 \pm 2.49$ \\
DRO-NoDT (ensemble + ROI + pool, no transformer) & -- & -- & $-10.01 \pm 2.56$ \\
TuRBO (paper run)                             & $-7.13 \pm 0.89$  & $-3.51 \pm 1.04$  & -- \\
DRO (paper run)                               & $-5.05 \pm 0.75$  & $-3.61 \pm 0.10$  & $-3.12 \pm 0.10$ \\
DRO control (fresh environment) & $-4.59 \pm 0.30$ & $-3.42 \pm 0.17$ & $-2.77 \pm 0.13$ \\
\bottomrule
\end{tabular}
\end{table}

The ROI helps but is far from sufficient, and the failure mode without the transformer is stark. The sorted final values of ROI-logEI over the ten trials are $-21.1$, $-21.0$, $-20.8$, $-19.9$, $-13.7$, $-5.8$, $-5.2$, $-5.0$, $-4.6$, $-4.1$: the arm is bimodal, with four or five seeds never escaping the initial plateau and the rest reaching $\approx -5$, still short of DRO. The across-trial standard deviation is $7.5$ for ROI-logEI versus $0.3$ for DRO, which succeeds on all ten seeds. With the ensemble and the ROI held identical, removing the transformer costs both performance and, more importantly, reliability; removing the ROI (DRO-GLOBAL) costs even more. The ROI is necessary, the learned policy is necessary, and neither is sufficient alone. TuRBO, which adapts a trust region without a learned policy, reaches comparable values on high-dimensional Ackley at large budgets but is less sample-efficient and trails on LunarLander and the HPO tasks (\figref{fig:ackley_dim_hpo_study}, \figref{fig:ablation_and_lunar_results}).

\paragraph{The policy is not an imitator of its pool.} Under pure imitation, DRO could not exceed the best member of its own rollout pool. Rollouts driven by EI alone train a policy reaching $-2.72 \pm 0.11$ on Ackley 10D (\figref{fig:ackley_10d_acq_study}), while logEI itself plateaus at $-18.49 \pm 0.25$ (\figref{fig:ackley_dim_hpo_study}); the return-to-go labels, not the behavior acquisition, carry the signal.

\subsection{Composition of the Rollout Acquisition Pool}
\label{subsec:pool}
\tabref{tab:pool} varies the pool of acquisition functions used to generate rollouts, on the Ackley 10D protocol of \tabref{tab:attribution}, with everything else identical to the fresh DRO control except for the arm marked $^{\dagger}$.
\begin{itemize}
    \item \textbf{Two-step behavior policy.} The pool $\{\mathrm{EI}, \mathrm{UCB}, \mathrm{PI}, \mathrm{MES}\}$ is extended with a fantasy-based two-step EI member that scores shortlisted candidates by $\mathrm{EI}(\mathbf{x}) + \mathbb{E}_y[\max_{\mathbf{x}'} \mathrm{EI}(\mathbf{x}' \mid \mathcal{D} \cup \{(\mathbf{x}, y)\})]$ with 3 posterior fantasies and 96 inner candidates, all inside the ROI, under uniform rotation. The result is statistically indistinguishable from the control at every checkpoint (final paired difference $+0.20 \pm 0.37$, paired $t$-test $p = 0.61$, Wilcoxon $p = 0.85$; numerically the two-step pool is slightly ahead). Seeding the pool with an explicitly non-myopic behavior policy does not materially change DRO: with terminal-return credit assignment in place, upgrading the behavior policy adds little, which is the direct test of the hypothesis that non-myopia cannot arise from myopic rollout data. The two-step pool runs at $4.2$~s/iteration versus $4.6$ for the control, the overhead being absorbed by shorter rollouts under early stopping.
    \item \textbf{Pure-variance exploration.} A posterior-standard-deviation acquisition is added to the rotation. At an equal share with the four standard members, DRO degrades from $-3.12 \pm 0.09$ to $-8.70 \pm 1.32$ at iteration 500 under the stored-run protocol: variance-driven rollouts rarely end with low regret, so their return labels carry little training signal and dilute the buffer. At a $\approx 10\%$ minority share (the four standard members at $22.5\%$ each), the effect is neutral (paired difference versus the control $-0.09 \pm 0.14$, $p = 0.55$). The dose-response is therefore: an equal share hurts substantially; a small minority is neutral.
\end{itemize}

\begin{table}[htbp]
\centering
\caption{Composition of the rollout acquisition pool on Ackley 10D (best observed objective, mean $\pm$ standard error over 10 paired trials, higher is better). $^{\dagger}$Run under the stored-run protocol; compare with the paper run in \tabref{tab:attribution}.}
\label{tab:pool}
\small
\begin{tabular}{lccc}
\toprule
\textbf{Rollout pool} & \textbf{it $=100$} & \textbf{it $=200$} & \textbf{it $=500$} \\
\midrule
DRO control (EI, UCB, PI, MES; fresh)                & $-4.59 \pm 0.30$ & $-3.42 \pm 0.17$ & $-2.77 \pm 0.13$ \\
$+$ two-step EI (uniform rotation)                   & $-4.40 \pm 0.37$ & $-3.32 \pm 0.14$ & $-2.58 \pm 0.28$ \\
$+$ posterior variance ($10\%$ share)                & $-5.07 \pm 0.50$ & $-3.60 \pm 0.21$ & $-2.86 \pm 0.13$ \\
$+$ posterior variance (equal share)$^{\dagger}$ & --               & --               & $-8.70 \pm 1.32$ \\
\bottomrule
\end{tabular}
\end{table}

\subsection{Kernel Diversity and Evidence-Weighted Rollout Allocation}
\label{subsec:kernel_diversity}
The RBF-only ensemble with uniformly allocated rollouts is a deliberate simplification. \tabref{tab:kernel} evaluates a variant in which the $M=10$ members cycle through RBF, Mat\'ern-5/2, Mat\'ern-3/2 and rational-quadratic kernels over the same lengthscale range, and rollouts are allocated across members by the softmax of each member's marginal log-likelihood on the real data, so that poorly fitting hypotheses are down-weighted and effectively excluded as evidence accumulates. The variant is statistically indistinguishable from the RBF-only control (paired difference $+0.27 \pm 0.23$, $p = 0.27$) and numerically ahead at every checkpoint: kernel diversity plus evidence-based rollout allocation is at worst free and possibly slightly beneficial, so the dynamic weighting is safe to adopt when misspecification of the kernel family is a concern.

\begin{table}[htbp]
\centering
\caption{Mixed-kernel ensemble with evidence-weighted rollout allocation on Ackley 10D (best observed objective, mean $\pm$ standard error over 10 paired trials, higher is better).}
\label{tab:kernel}
\small
\begin{tabular}{lccc}
\toprule
\textbf{Ensemble} & \textbf{it $=100$} & \textbf{it $=200$} & \textbf{it $=500$} \\
\midrule
RBF only, uniform allocation (control)              & $-4.59 \pm 0.30$ & $-3.42 \pm 0.17$ & $-2.77 \pm 0.13$ \\
Mixed kernels, marginal-likelihood weights          & $-4.13 \pm 0.39$ & $-3.09 \pm 0.18$ & $-2.50 \pm 0.16$ \\
\bottomrule
\end{tabular}
\end{table}

\paragraph{Protocol notes for the new arms.} All new arms use Ackley 10D on $[-32.768, 32.768]^{10}$ with observation noise $\sigma = 0.01$, the $+10$ input shift (optimum at $\mathbf{x} = 10\cdot\mathbf{1}$), $N_{\text{init}} = 5$ Sobol points, seeds 42--51, and $M = 10$ ensemble members; wall times were measured on the same Apple-silicon CPU with two threads per run. Paired comparisons of the new variants are made against the fresh control run under identical code, environment and seeds, not against the stored curves.

\end{document}